\documentclass{article}

\usepackage{arxiv}

\usepackage[utf8]{inputenc} %
\usepackage[T1]{fontenc}    %
\usepackage{hyperref}       %
\usepackage{url}            %
\usepackage{booktabs}       %
\usepackage{amsfonts}       %
\usepackage{nicefrac}       %
\usepackage{microtype}      %
\usepackage{lipsum}		%
\usepackage{graphicx}
\usepackage{natbib}
\usepackage{doi}
\usepackage{amsmath}
\usepackage{multirow}
\usepackage{color}

\title{GS-Octree: Octree-based 3D Gaussian Splatting for Robust Object-level 3D Reconstruction Under Strong Lighting}

\author{ Jiaze Li$^*$ \\
	College of Computing and Data Science\\
	Nanyang Technological University\\
    Singapore\\
	\And
	Zhengyu Wen$^*$ \\
	College of Computing and Data Science\\
	Nanyang Technological University\\
    Singapore\\
        \And
	Luo Zhang\thanks{J. Li, Z. Wen and L. Zhang contribute equally to the project.} \\
	College of Computing and Data Science\\
	Nanyang Technological University\\
    Singapore\\
        \And
	Jiangbei Hu \\
    International School of Information Science \& Engineering\\
	Dalian University of Technology \\
    China\\
        \And
	Fei Hou \\
	Institute of Software\\
        Chinese Academy of Sciences\\
    China
         \And
	Zhebin Zhang \\
    InnoPeak Technology, Inc.\\
    United States of America\\
        \And
	Ying He\thanks{Corresponding author: Y. He (yhe@ntu.edu.sg) } \\
College of Computing and Data Science\\
	Nanyang Technological University\\
    Singapore\\
}

\renewcommand{\shorttitle}{\textit{arXiv} Template}

\begin{document}
\maketitle

\newcommand{\imgleneight}{0.121}
\newcommand{\imglenseven}{0.133}
\newcommand{\imglensix}{0.155}
\newcommand{\imglenfive}{0.182}
\newcommand{\imglenfour}{0.24}

\begin{abstract}
	The 3D Gaussian Splatting technique has significantly advanced the construction of radiance fields from multi-view images, enabling real-time rendering. While point-based rasterization effectively reduces computational demands for rendering, it often struggles to accurately reconstruct the geometry of the target object, especially under strong lighting. To address this challenge, we introduce a novel approach that combines octree-based implicit surface representations with Gaussian splatting. Our method consists of four stages. Initially, it reconstructs a signed distance field (SDF) and a radiance field through volume rendering, encoding them in a low-resolution octree. The initial SDF represents the coarse geometry of the target object. Subsequently, it introduces 3D Gaussians as additional degrees of freedom, which are guided by the SDF. In the third stage, the optimized Gaussians further improve the accuracy of the SDF, allowing it to recover finer geometric details compared to the initial SDF obtained in the first stage. Finally, it adopts the refined SDF to further optimize the 3D Gaussians via splatting, eliminating those that contribute little to visual appearance. Experimental results show that our method, which leverages the distribution of 3D Gaussians with SDFs, reconstructs more accurate geometry, particularly in images with specular highlights caused by strong lighting. %
\end{abstract}

    

\section{Introduction}
\begin{figure*}[t] 
    \centering
    \setlength\tabcolsep{1pt}
    \begin{scriptsize}
    \begin{tabular}{cccccccc}
    \includegraphics[width=\imgleneight\textwidth]{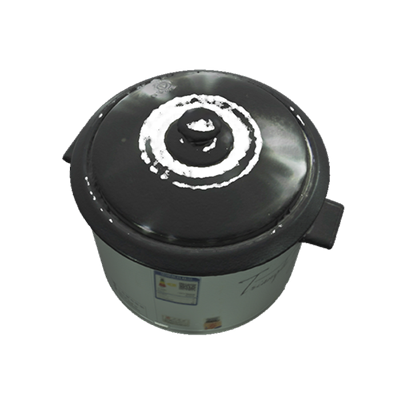}&
    \includegraphics[width=\imgleneight\textwidth]{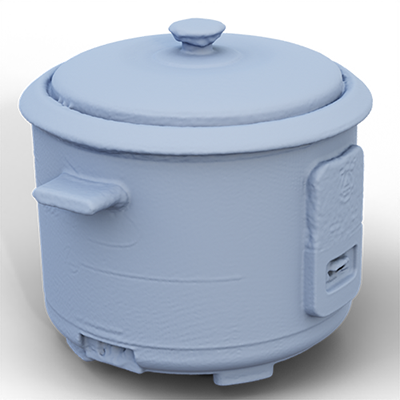}&
    \includegraphics[width=\imgleneight\textwidth]{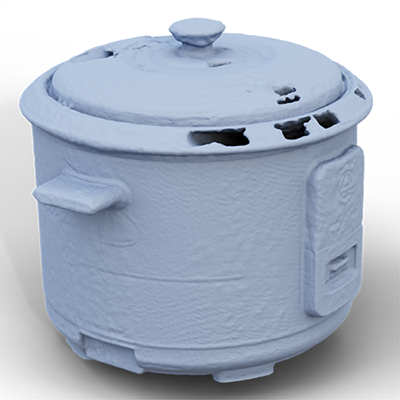}&
    \includegraphics[width=\imgleneight\textwidth]{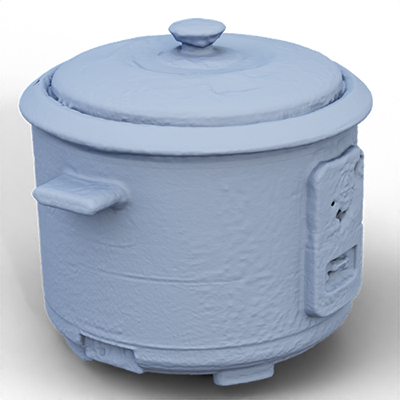}&
    \includegraphics[width=\imgleneight\textwidth]{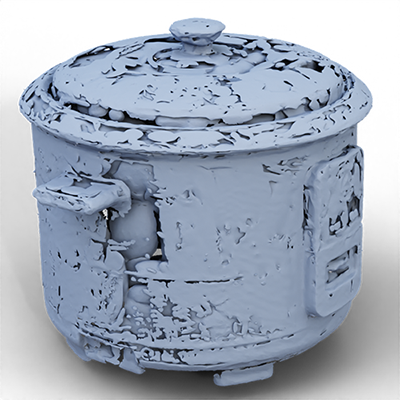}&
    \includegraphics[width=\imgleneight\textwidth]{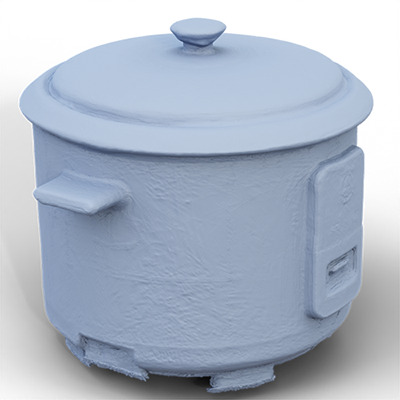}&
    \includegraphics[width=\imgleneight\textwidth]{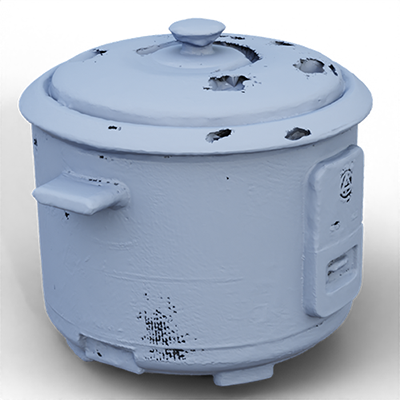}&
    \includegraphics[width=\imgleneight\textwidth]{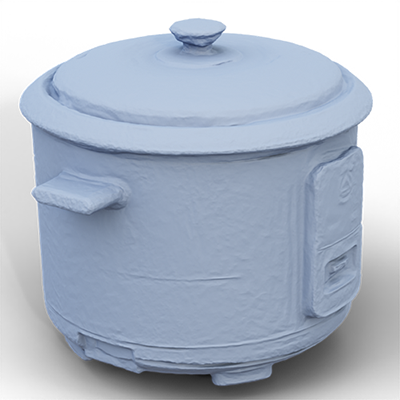}
    \\
    & CD & 9.76 & 4.14 &  14.06 & 1.48 & 6.10 & \textbf{0.27} \\
    & PSNR, FPS, $N_{\mathrm{GS}}$ & 35.09, 1.91, - & 34.44, 8.94, - & 34.38, 32.66, - & \textbf{34.88}, 62.97, 60.14 & 33.58, 91.81, 80.29 & 33.24, \textbf{189.87}, \textbf{58.33} \\
    \includegraphics[width=\imgleneight\textwidth]{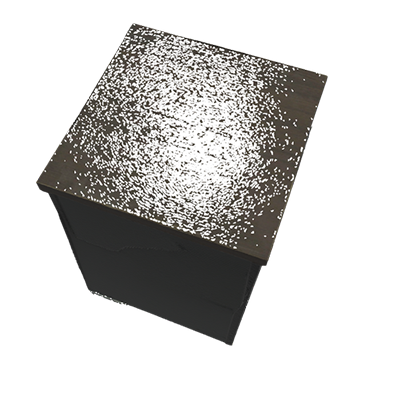}&
    \includegraphics[width=\imgleneight\textwidth]{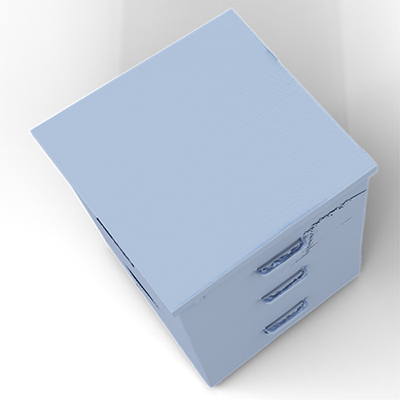}&
    \includegraphics[width=\imgleneight\textwidth]{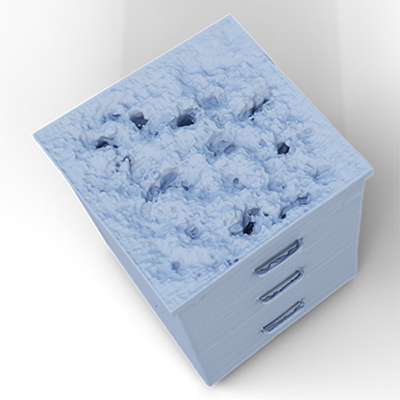}&
    \includegraphics[width=\imgleneight\textwidth]{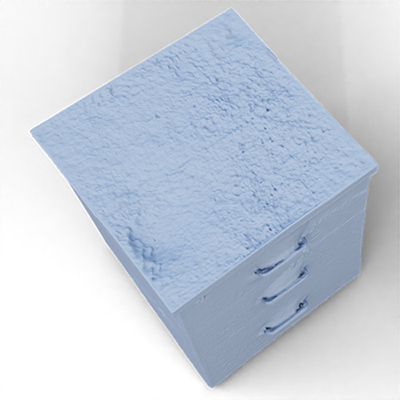}&
    \includegraphics[width=\imgleneight\textwidth]{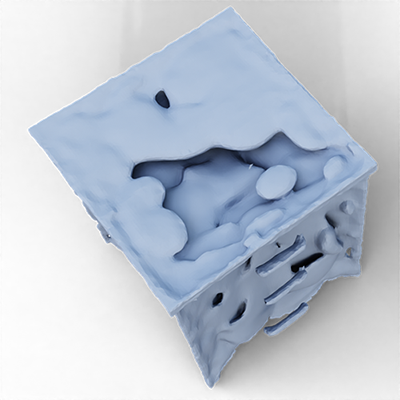}&
    \includegraphics[width=\imgleneight\textwidth]{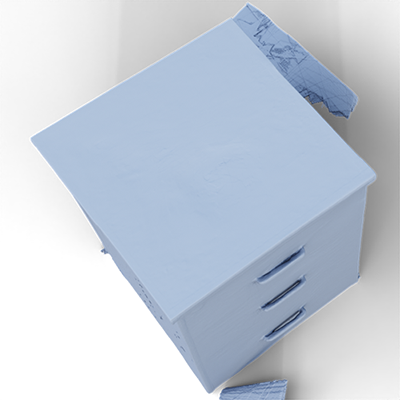}&
    \includegraphics[width=\imgleneight\textwidth]{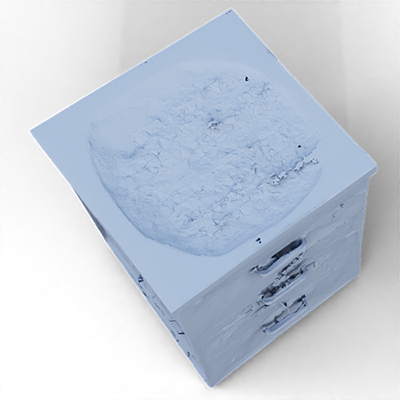}&
    \includegraphics[width=\imgleneight\textwidth]{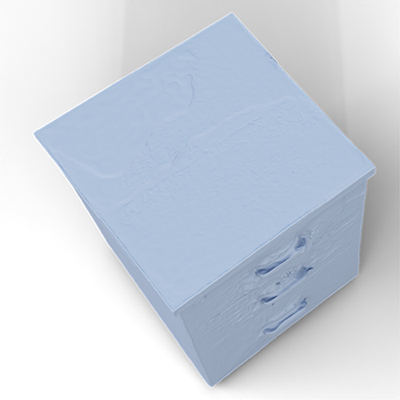}
    \\
    & CD & 6.46 & 3.96 & 56.40 & 58.05 & 10.04 & \textbf{3.87} \\ 
    & PSNR, FPS, $N_{\mathrm{GS}}$ & 37.81, 2.48, - & 38.69, 5.09, - & 33.99, 36.20, - & 38.80, 56.94, 87.46 & \textbf{39.09}, 56.21, 107.76 & 37.75, \textbf{150.84}, \textbf{57.72}\\
    GT image & GT mesh & Voxurf & NeuS2 & SuGaR & 2DGS & GOF & Ours\\
    \end{tabular}
    \vspace{-2mm}
    \begin{tabular}{ccccccc}
    \includegraphics[width=\imglenseven\textwidth]{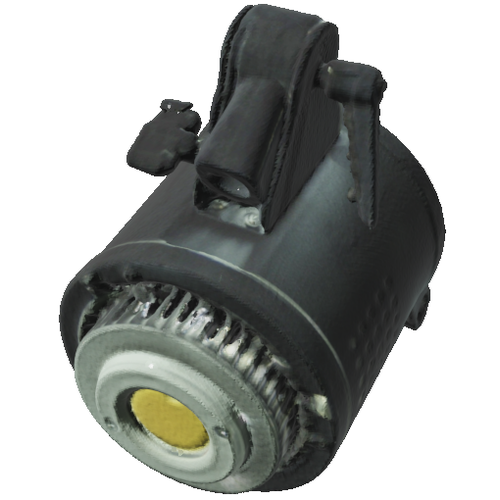}&
    \includegraphics[width=\imglenseven\textwidth]{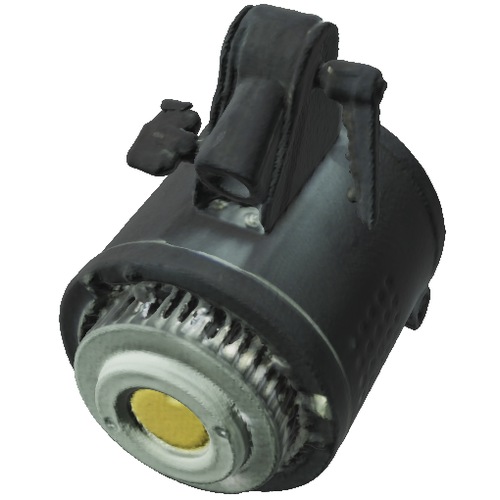}&
    \includegraphics[width=\imglenseven\textwidth]{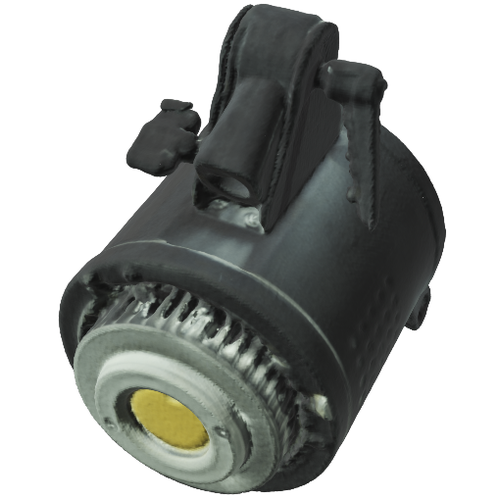}&
    \includegraphics[width=\imglenseven\textwidth]{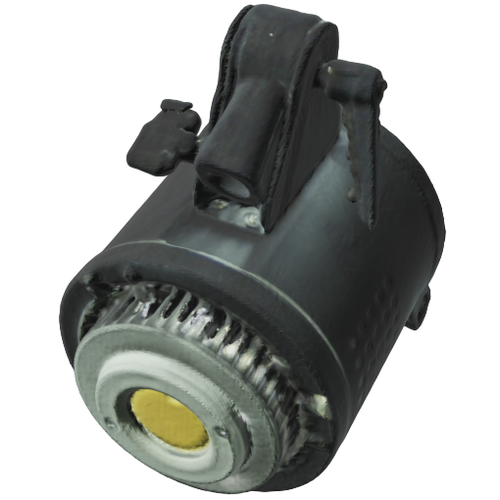}&
    \includegraphics[width=\imglenseven\textwidth]{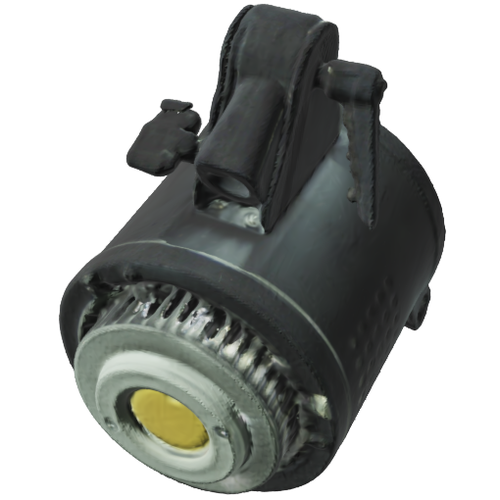}&
    \includegraphics[width=\imglenseven\textwidth]{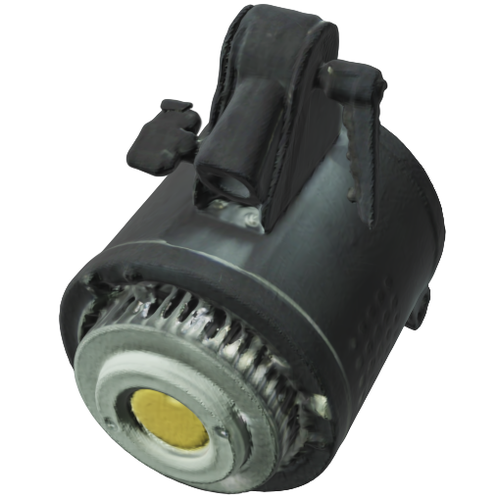}&
    \includegraphics[width=\imglenseven\textwidth]{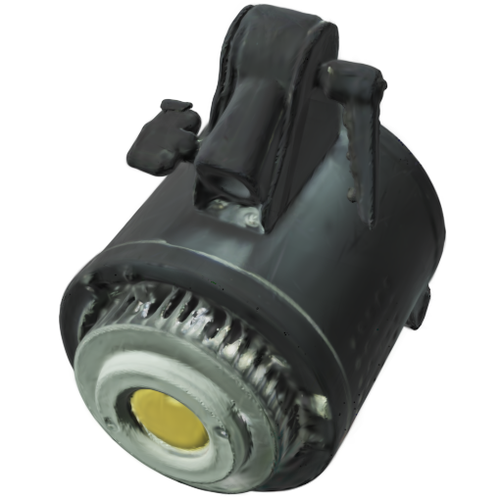}
    \\
    CD & 1.98 & 1.03 & - & 1.32 & 1.06 & \textbf{0.55} \\
    PSNR, FPS, $N_{\mathrm{GS}}$ & 35.03, 1.61, - & \textbf{37.12}, 11.15, - & 35.27, 180.07, 107.50 & 36.66, 65.18, 70.35 & 35.58, 88.73, 81.66 & 34.42, \textbf{215.33}, \textbf{54.60}\\
   
    GT image & Voxurf & NeuS2 & 3DGS & 2DGS & GOF & Ours\\
    \end{tabular}
    \end{scriptsize}
    \caption{Our method integrates octree-based implicit surface representations with Gaussian splatting, enabling real-time rendering for novel view synthesis using fewer Gaussians. This integration allows us to robustly reconstruct high-quality geometry from input images with large areas of specular highlight due to strong lighting. The values below each figure represent the Chamfer distance ($10^{-4}$), PSNR, FPS and the number of Gaussians $N_{\mathrm{GS}}$ (in thousands), with the best results highlighted in bold.} \label{fig:teaser}
\end{figure*}

Reconstructing 3D geometry and appearance from multi-view images is a critical area in 3D vision. Recently, Neural Radiance Field (NeRF) based methods~\cite{nerf,mipnerf,mipnerf360} have significantly advanced the state-of-the-art for this problem. However, these methods often face challenges with long training times and slow rendering speeds due to the complexity of optimizing neural networks. 
To enhance training speed and reduce rendering time while maintaining quality, some methods adopt explicit representations to store features~\cite{plenoxels,wu2022voxurf,dvgo,nsvf}. While these approaches offer faster rendering compared to implicit methods, they often struggle with accurate geometric reconstruction. Furthermore, the lack of compactness in these methods poses challenges in improving resolution and reconstruction accuracy. These limitations are primarily due to the constraints of density-based volume rendering, which relies on the precision of the grid and the initial structure setup. To address these issues, DVGO~\cite{dvgo} and DVGOv2~\cite{dvgov2} integrate explicit and implicit methods by employing a shallow multi-layer perception (MLP) on the voxel grid representing features, achieving a better balance between the quality and efficiency of novel view synthesis.


Recently, 3D Gaussian Splatting (3DGS)~\cite{3Dgaussians} has emerged as a promising direction for differentiable rendering. It enables efficient training and rendering by optimizing many 3D Gaussian points, yielding visually appealing results. However, due to the lack of geometric constraints, 3DGS struggles to reconstruct high-quality geometry. Subsequent works, such as~\cite{sugar, neusg, dreamgaussian, gaussiandreamer, gof, 2dgs}, attempt to extract explicit geometries from Gaussian splatting. For instance, 2DGS~\cite{2dgs} deforms 3D Gaussians into 2D oriented disks and utilises them to model surfaces, significantly improving scene-level geometry quality. However, for object-level reconstruction,  particularly when the target object contains regions with relatively few views in the input images, 2DGS often yields suboptimal results with redundant components. Consequently, the geometry quality is not comparable to pure volume-rendering-based neural methods, such as Voxurf~\cite{wu2022voxurf} and NeuS2~\cite{neus2}.
GOF~\cite{gof} induces a Gaussian opacity field from 3D Gaussians and extracts its level set through efficient marching tetrahedra, further improving the quality of geometry extracted from 3DGS~\cite{3Dgaussians}. However, its surface extraction, which relies on opacity, is sensitive to strong lighting. In regions with strong light, the opacity is often very low, resulting in holes in the reconstructed surface at those positions.

Volume rendering-based methods, such as VolSDF~\cite{volsdf}, NeuS~\cite{neus}, NeuS2~\cite{neus2} and Voxurf~\cite{wu2022voxurf}, can achieve high geometric quality in object level reconstruction, however, strong lighting makes the color of object surface change dramatically under different view directions which can disrupt the reconstruction of its opacity field, resulting in inaccurate geometry surface reconstruction. We found that the position of 3DGS points is robust to strong light, as Gaussian points can still surround the surface of the object even under strong light, which can assist the volume rendering framework to generate more accurate geometry under strong light. In this paper, we propose a novel octree-based method for reconstructing object-level implicit surfaces under strong light. We formulate the reconstruction as a coarse-to-fine optimization problem, guided by SDF and 3D Gaussians, without relying on any neural networks. Our method consists of four stages. Initially, it reconstructs an SDF and a radiance field through volume rendering, encoding the coarse geometry of the target object in an octree with low resolution. Subsequently, it introduces 3D Gaussians as additional degrees of freedom, optimized through the progressive refinement of the octree, guided by the SDF. In the third stage, the optimized Gaussians further improve the accuracy of the SDF, enabling the recovery of finer geometric details by pulling surfaces towards 3D Gaussian points, offsetting the artefacts caused by specular highlights. Finally, the refined SDF is utilised to further optimize 3D Gaussians via splatting, eliminating those that contribute little to the visual appearance. See Fig~\ref{fig:teaser} for a comparative illustration of the results generated by ours and other methods.


We summarize our main contributions as follows:

\begin{itemize}
\item 
We introduce a novel octree-based representation that seamlessly combines SDFs and 3D Gaussians, enabling real-time rendering with high-quality geometric surface reconstruction.
\item Our approach harnesses the complementary advantages of SDF and 3D Gaussian representations, enabling the reconstruction of accurate geometry even in the presence of strong lighting. Our methods also reduce the number of 3D Gaussians by more than 50\% while preserving competitive rendering quality and fast rendering speed.
\item
Our octree-based method allows for more efficient computation of high-order derivatives of the information field. This enables us to introduce singular-Hessian loss to enhance the reconstruction of geometry. 
\end{itemize}


\section{Related Work}

\subsection{Neural Implicit Surfaces}
NeRF~\cite{nerf} utilizes neural volume rendering to synthesize pixel colors from multiple points sampled on each camera ray. However, it lacks a practical constraint on the reconstructed geometry's surface, relying on threshold settings in its volume density function, which can introduce noise and potential issues with the reconstructed geometry. In contrast, methods based on neural implicit surface functions (e.g., occupancy fields and SDFs) achieve superior surface modeling and geometric reconstruction without compromising the quality of visual appearance~\cite{idr,dvr}. However, these approaches often require 2D mask supervision for training, which can be challenging. NeuS~\cite{neus}, VolSDF~\cite{volsdf} and their many follow-up works naturally combine neural volume rendering with SDFs and are able to achieve high-quality geometry for object-level reconstruction. However, it is hard to extend these methods to scene-level reconstruction tasks.

\subsection{Training Performance Improvement}
Training neural radiance fields and their implicit surface-based variants, such as NeuS~\cite{neus} and VolSDF~\cite{volsdf}, require significant computational resources. Various techniques~\cite{ingp,rosu2023permutosdf} have been proposed to reduce the training cost. 
For example, INGP~\cite{ingp} achieves a substantial reduction in the number of training parameters through the implementation of multi-resolution hash coding coupled with highly optimized CUDA kernels. NeuS2~\cite{neus2}, PermutoSDF~\cite{rosu2023permutosdf}, and Neuralangelo~\cite{Neuralangelo} employ multi-resolution hash encodings and shallow MLPs proposed in INGP to replace a large neural network for accelerating NeuS training. DVGO~\cite{dvgo} and Voxurf~\cite{wu2022voxurf} achieve training acceleration by incorporating both voxelization and smaller MLPs. PermutoSDF~\cite{rosu2023permutosdf} employed a triangular pyramid grid for reducing calculation overhead during interpolation. TensorRF\cite{chen2022tensorf} and Strivec~\cite{gao2023strivec} leverage tensor decomposition for tensor grid modeling. However, most methods are limited by the network structure and cannot achieve high-speed rendering.

\subsection{Real-time Rendering}
Various strategies ~\cite{fastnerf,yu2021plenoctrees,fourierplenoctree} have been proposed for significantly increasing rendering speeds. For example, Plenoctree~\cite{yu2021plenoctrees} and RT-Octree~\cite{rtoctree} expedite the process by converting NeRF models into an explicit octree, while Plenoxels~\cite{plenoxels} adopts voxel-based representation for fast rendering. Additionally, BakedSDF~\cite{yariv2023bakedsdf} pre-calculates and stores rendering parameters through the baking method, enabling a smaller MLP to swiftly read these parameters during rendering.

\subsection{3D Gaussian Splatting}
The recently proposed 3D Gaussian Splatting~\cite{3Dgaussians} takes a completely different approach from volume rendering-based neural radiance fields. It innovatively integrates point-based rendering and differentiable rendering, achieving real-time rendering through efficient parallel computing. Moreover, the explicit modelling of 3D scenes with Gaussians provides a shortcut for controlling the dynamics of a scene, particularly crucial in complex situations with diverse geometries and changing lighting conditions.

The capability for control and modification, coupled with an efficient rendering process, positions 3DGS as a revolutionary technique in shaping multi-view 3D reconstruction~\cite{sugar, neusg, scaffold-gs, GauU-Scene, lyu20243dgsr, gof, 2dgs}. SuGaR~\cite{sugar} optimizes the position of 3D Gaussians by aligning them with the surfaces and using an SDF to guide the position and orientation of 3D Gaussians. Meanwhile, NeuSG~\cite{neusg} leverages NeuS~\cite{neus} with 3D Gaussian guidance to enhance the optimization process of NeuS-generated SDF. GOF~\cite{gof} establishes a Gaussian opacity field from 3D Gaussians and provides an approximation of surface normals. 2DGS~\cite{2dgs} models the surfaces by using flat 3D Gaussians, deforming them into 2D oriented Gaussian disks. 3DGSR~\cite{lyu20243dgsr} trains an SDF and proposes a differentiable SDF-to-opacity transformation function, to generate corresponding opacities of 3D Gaussians.

Although 3DGS is capable of reconstructing scene-level geometries, current 3DGS-based reconstruction methods~\cite{gof, 2dgs} are not robust to specular highlights in images or need postprocessing to remove floating surfaces. Our method addresses these challenges by jointly optimizing 3DGS with octree-encoded SDFs. Leveraging the network-free feature of our method, we can utilize effective geometric regularizers with high-order derivatives, such as the singular Hessian loss, to improve the quality of the geometry. We refer readers to recent surveys~\cite{3dgssurvey, DBLP:journals/corr/abs-2403-11134,DBLP:journals/corr/abs-2401-03890} for a comprehensive overview of 3DGS techniques.

\section{Preliminaries}
\subsection{Volume Rendering} 
\begin{figure*}[t] 
\centering    
\includegraphics[width=1.0\linewidth]
{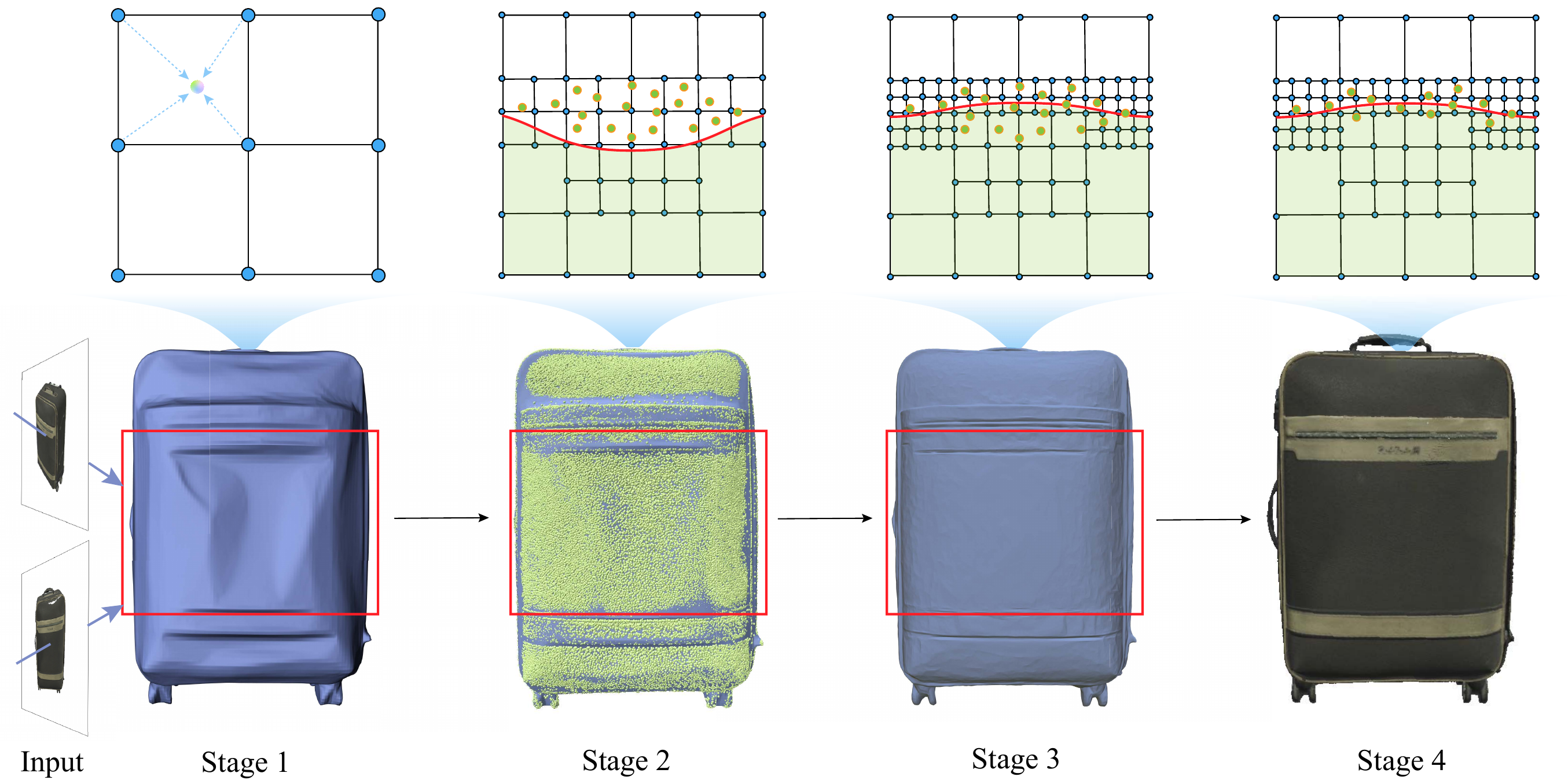}
    
\caption{\textbf{Algorithmic pipeline.} Our framework consists mainly of four stages. Utilizing the octree, we progressively optimize the signed distance values $s_{ij}$ for geometry and the SH coefficients $\mathbf{a}_{ij}$ for radiance for each octree node in a coarse-to-fine manner, alternating between volume rendering and point splatting. 
} 
\label{fig:pipeline}
\end{figure*}

The volume rendering technique~\cite{nerf} synthesizes pixel colors for the output images by utilizing 3D fields of radiances $\mathbf{c}$ and densities $\sigma$.
Specifically, color is computed by integrating along ray $\mathbf{r}$  shooting from the camera position $ \mathbf{o}$ with sampling points  $\{\mathbf{o}+t_i\mathbf{d}\}_{i=1}^N$ as
\begin{equation}
\label{eq:nerf_color}
\mathbf{C}(\mathbf{r}) = \sum_i^N T_i(1-\exp(-\sigma_i\delta_i))\mathbf{c}_i,
\end{equation}
where $T_i=\exp(-\sum_{j=1}^{i-1} \sigma_i\delta_i)$ represents the transparency, and $\delta_i$ is the distance between sampling points. NeRF~\cite{nerf} employs a neural network to learn radiance fields from the given multi-view images. 
However, its inefficiency in training and inference has spurred the recent development of network-free methods, such as 3DGS~\cite{3Dgaussians}.

Neural implicit surface-based methods, such as NeuS~\cite{neus} and VolSDF~\cite{volsdf}, jointly learn an SDF $\mathcal{S}$ and a radiance field $\mathbf{c}$ from the input images. To enable volume rendering, they convert the SDF $\mathcal{S}(\mathbf{x})$ into volume density as follows:
\begin{equation}
\label{eq:sdf f2}
\sigma(\mathbf{x}) =
  \begin{cases}
    \frac{1}{2\beta}\exp(\frac{\mathcal{S}(\mathbf{x})}{\beta})       & \quad \text{if } \mathcal{S}(\mathbf{x}) \text{ $\leq$ 0},\\
    \frac{1}{\beta}-\frac{1}{2\beta}\exp(-\frac{\mathcal{S}(\mathbf{x})}{\beta})   & \quad \text{if } \mathcal{S}(\mathbf{x}) \text{ $>$  0},
  \end{cases}
\end{equation}
where $ \beta$ is a learnable parameter to adjust the smoothness of the density function $ \sigma $ near the object boundary. In our framework, we introduce an octree-based structure to encode both SDFs and radiances, enabling real-time volume rendering.

\subsection{Point-Based Rendering}
3DGS~\cite{3Dgaussians} utilizes an ensemble of 3D Gaussians to represent the radiance fields, offering greater flexibility and efficiency than volumetric radiance fields. Specifically, a Gaussian at position $\mathbf{x}$ is defined as
\begin{equation}
\label{eq: gs f2}
G(\mathbf{x}) = e^{-\frac{1}{2}{\mathbf{x}^T}\Sigma^{-1}\mathbf{x}},
\end{equation}
where $\Sigma$ is the covariance matrix controlling the size and orientation of the Gaussian. During rendering, the 3D Gaussians $G(\mathbf{x})$ are first transformed into 2D Gaussians $G'(\mathbf{x})$ on the projected image plane. After that, a tile-based rasterizer is applied to split the image plane into tiles and sort these 2D Gaussians efficiently. Finally, $\alpha$-blending is performed on these 2D Gaussians and the color of each pixel is obtained as follows:
\begin{equation}
\label{eq: gs f1}
C(\mathbf{x'}) = \sum_{i\in N}c_i \sigma_i \prod^{i-1}_{j=1} (1-\sigma_j),  \sigma_i = \alpha_i G'_i(\mathbf{x'}),
\end{equation}
where $\mathbf{x'}$ is the pixel positions on the image plane and $\alpha_i$ is the opacity of the $i$-th Gaussian. The parameters of Gaussian functions can be directly optimized using a gradient-based solver without neural networks, significantly improving the training and rendering efficiency. 


\begin{figure}[b] \centering
    \includegraphics[width=0.24\linewidth]{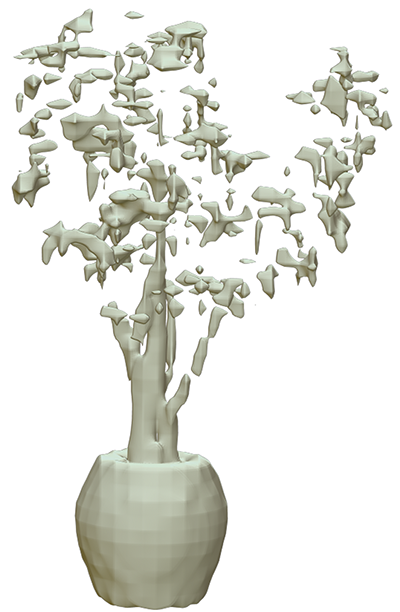}
    \includegraphics[width=0.24\linewidth]{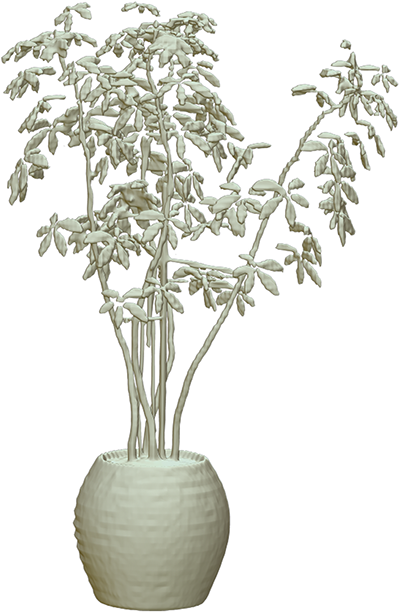}
    \includegraphics[width=0.24\linewidth]{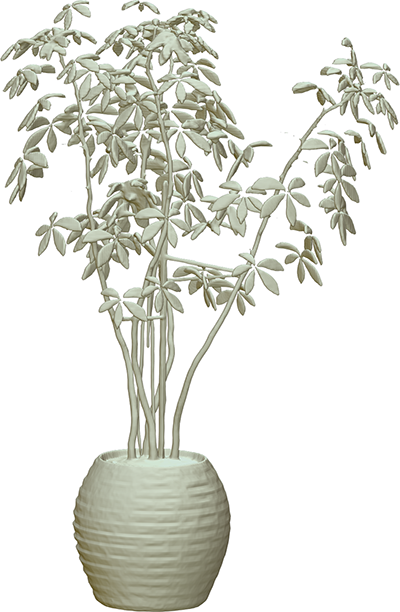}
    \includegraphics[width=0.24\linewidth]{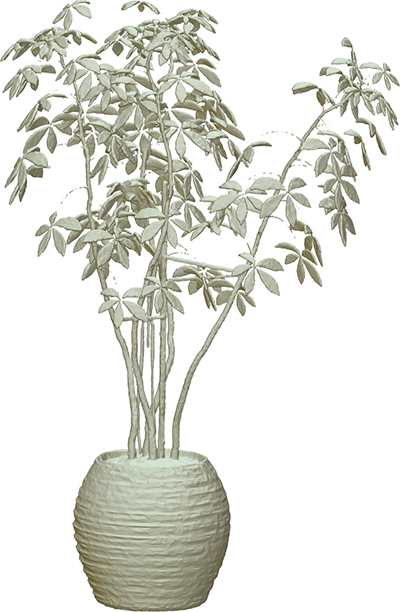}
    \\
    \makebox[0.24\linewidth]{\footnotesize $l=6$}
    \makebox[0.24\linewidth]{\footnotesize $l=7$}
    \makebox[0.24\linewidth]{\footnotesize $l=8$}
    \makebox[0.24\linewidth]{\footnotesize $l=9$}
    \caption{Qualitative results from octrees spanning level 6 to 9. As the resolution of the octree increases, it provides more degrees of freedom, effectively improving the quality of the reconstructed geometry. } 
    \label{fig:subdive}
\end{figure}

\section{Method}

\begin{figure*}[t]
    \centering
    \makebox[0.12\linewidth]{GT}
    \makebox[0.42\linewidth]{w/o Gaussian guidance}
    \makebox[0.42\linewidth]{w/ Gaussian guidance}
    \\
    \includegraphics[width=1.0\linewidth, trim=0 1.5cm 0 1.8cm, clip]{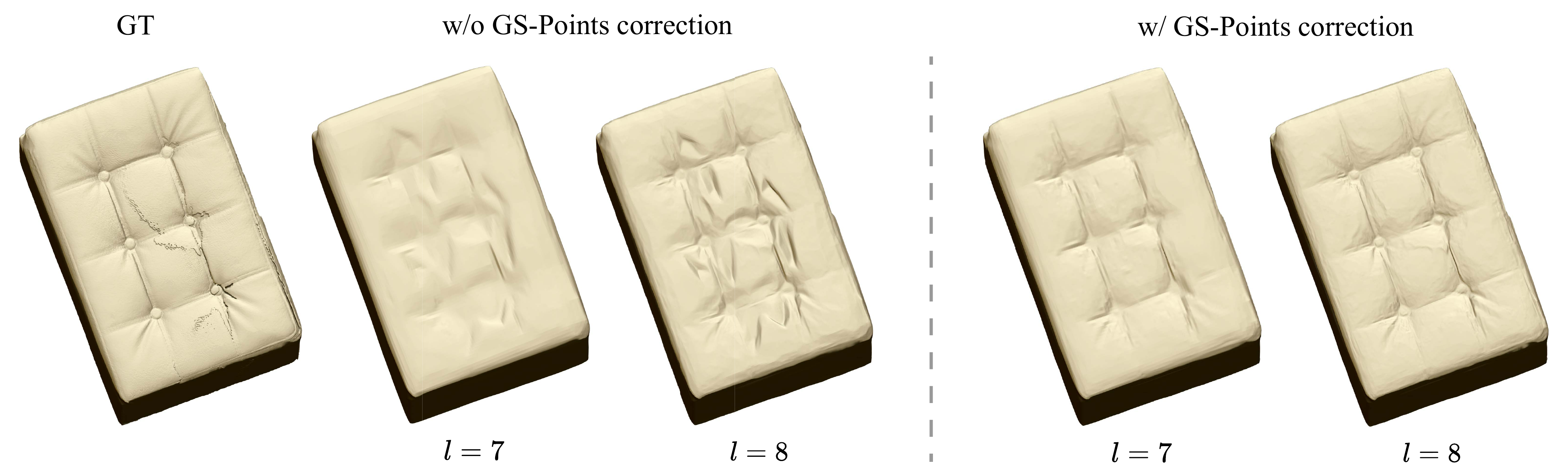}
    \\
    \makebox[0.18\linewidth]{}
    \makebox[0.19\linewidth]{$l=7$}
    \makebox[0.19\linewidth]{$l=8$}
    \makebox[0.04\linewidth]{}
    \makebox[0.17\linewidth]{$l=7$}
    \makebox[0.17\linewidth]{$l=8$}
    \caption{Gaussian-guided geometric optimization. Left: Pure octree-based SDFs are prone to inaccurate geometry, and simply refining the octree does not address the issue.  Right: We propose to leverage a Gaussian point cloud generated from GS to enhance reconstruction accuracy.} \label{fig:gsp_corr}
\end{figure*}

\textbf{Overview.} As illustrated in Figure~\ref{fig:pipeline}, our method consists of four stages. 
    In Stage 1, we learn an implicit surface under volume rendering and encode it by a low-resolution octree, representing the rough geometry of the target object. In Stage 2, we initialize a point cloud $G_p$ from the SDF obtained in Stage 1 and optimize the position of each point under point splatting. In Stage 3, we leverage the optimized $G_p$ to improve the geometry of the SDF via progressively refining the octree. At the end of this stage, we obtain an implicit surface with high-quality geometry. Finally, in Stage 4, we further optimize the number and positions of Gaussians under the guidance of SDF, eliminating the points that contribute little to the rendering. The final images are then produced in real-time by applying point splatting to the finalized 3D Gaussians.

\begin{figure}[htbp] \centering
    \begin{minipage}{0.27\textwidth}
        \centering
        \includegraphics[width=0.8\textwidth]{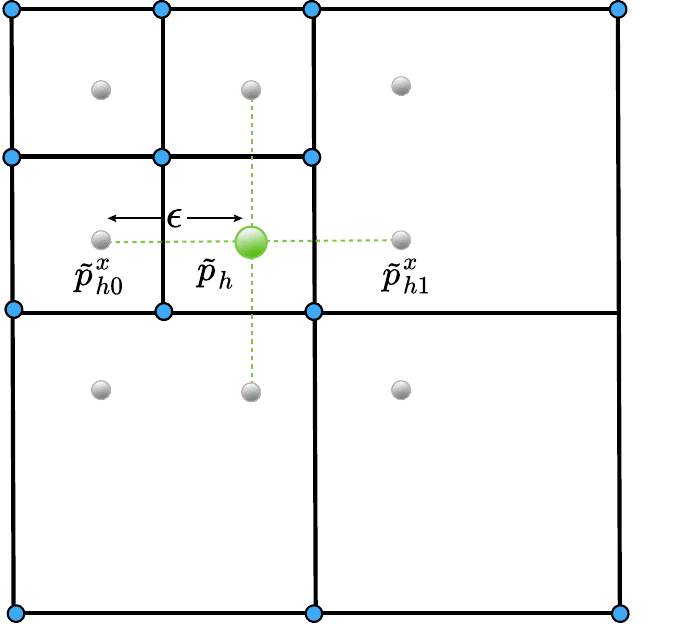}
        \caption{We calculate the Hessian matrix using a numerical method based on octree grids. } 
        \label{fig:hess}
    \end{minipage}\hfill
    \begin{minipage}{0.7\textwidth}
        \centering
        \includegraphics[width=0.24\linewidth]{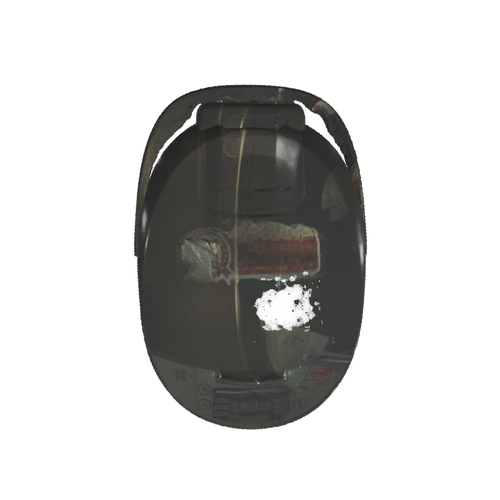}
        \includegraphics[width=0.24\linewidth]{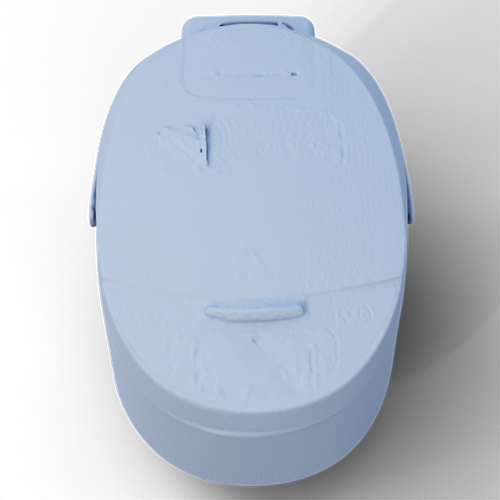}
        \includegraphics[width=0.24\linewidth]{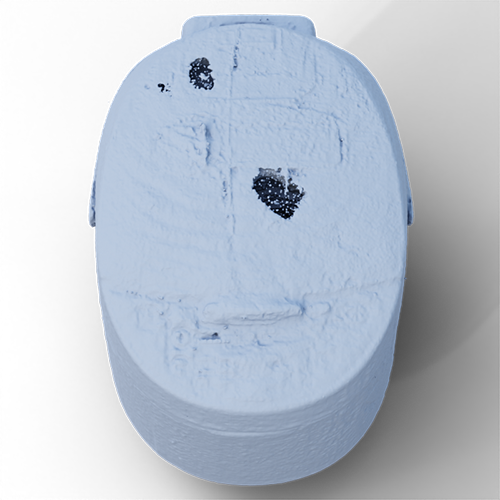}
        \includegraphics[width=0.24\linewidth]{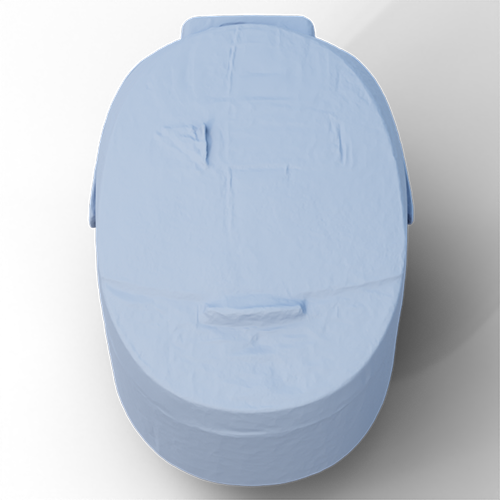}
        \\
        \makebox[0.24\linewidth]{GT image}
        \makebox[0.24\linewidth]{GT mesh}
        \makebox[0.24\linewidth]{w/o Hessian}
        \makebox[0.24\linewidth]{w/ Hessian}
        \caption{By incorporating the singular-Hessian loss term, we significantly reduce the artifacts generated from $G_p$, most of which are not on the surface and exhibit noise.}
         \label{fig:nohe}
    \end{minipage}
\end{figure}

\subsection{Octree-based 3D Reconstruction}

As a hierarchical data structure, octree is well-suited for managing complex spatial information~\cite{octrees}. It starts with a root node representing a cubical region, which iteratively subdivides into octants according to the data contained in each node.
We define the final subdivided grid as leaf nodes ${\ell_i}, {i\in I^{l}}$, where $l$ is the subdivision level of this node, and $I^l$ is the total number of nodes at level $l$. Each leaf node comprises eight grid points $\{\mathbf{p}_{ij}\}_{j=1}^8$.

We store the SDF values $s_{ij}=\mathcal{S}(\mathbf{p}_{ij})$ at the octree grid points $\mathbf{{p}_{ij}}$ to implicitly represent geometric shapes. We also store the SH coefficients $\mathbf{a}_{ij}=\{a_l^m\}_{ij}$ for radiance.
Under the supervision of the input images, we optimize the parameters $\Theta=\{\theta_{ij}=(s_{ij}, \mathbf{a}_{ij})\}_{i,j}$ using the gradient descent method. 
We use Tri-linear interpolation is to compute property values for randomly selected 256 points within octree leaf nodes. We then calculate the minimum SDF value $s_i$ for each lead nodes, guiding further octree subdivision (see Fig.~\ref{fig:subdive}). With the SDFs encoded in octree, we can directly extract the zero-level set by applying the Marching Cubes algorithm to each octree node~\cite{lorensen1998marching}. For reconstructing the radiance field, we convert the SDF values into density using Eq.\eqref{eq:sdf f2} and then obtain color information for different viewpoints through SH interpolation. Subsequently, we use volume rendering to synthesize imagesfrom novel views. 

Many network-based methods learn an SDF by regularizing it using the Eikonal loss $\mathcal{L}_{\text{eik}}$:
\begin{equation} 
\label{eq:eiko}
\mathcal{L}_\text{eik} = \int \left(\Vert \nabla \mathcal{S}(\mathbf{x}) \Vert - 1\right)^2 d\mathbf{x}
\end{equation} 
This strategy works well for SDFs encoded by a neural network. However, as pointed out by Pumarola et al.~\cite{viscoty}, when optimizing SDFs via an explicit data structure, such as octrees, the Eikonal loss often leads to the distance value $\mathcal{S}$ falling into undesirable minima due to the lack of the inductive bias from a neural network. To address this problem, we adopt the singular-Hessian loss proposed by Wang et al.~\cite{hessian}, which can not only prevent the optimization from getting into unexpected local minima but also simultaneously smooth the surface, reducing noise during the optimization. Unlike traditional general-purpose smoothing items such as the Laplacian, the singular-Hessian loss, which is tailored to distance fields, has the ability to preserve geometric details to some extent while smoothing the surface. The singular-Hessian loss is defined as follows:
\begin{equation} 
\label{eq:hess}
\begin{split}
&\mathcal{L}_\text{hess} = \frac{1}{N_s} \sum_{\tilde{p}_s^i\in P_s} \Vert \text{Det}(\mathbf{H}_S(\tilde{p}_s^i))\Vert_1\quad \text{for}~ i=1, 2,\cdots, N_{s},
\\
    &\mathbf{H}_S(p)
    =
    \left[
    \begin{array}{ccc}
        D_{xx}(p)& D_{xy}(p) & D_{xz}(p) \\
        D_{yx}(p)& D_{yy}(p) & D_{yz}(p) \\
        D_{zx}(p)& D_{zy}(p) & D_{zz}(p) 
    \end{array}
    \right],
\end{split}
\end{equation}
where $P_s$ is the set of randomly sampled points around the zero level-set in the leaf nodes with max depth $l_c$ and $l_c - 1$, and $N_s$ is the number of samples.

Given that the octree structure is explicit, computing second-order derivatives is straightforward. Similar to~\cite{Neuralangelo}, we utilize the finite difference method for this calculation. As shown in Fig.~\ref{fig:hess}, for a sampled point $\tilde{p}_h$, we sample two additional points along each coordinate axis surrounding $\tilde{p}_h$ with a step size of $\epsilon$. Taking the second-order derivative on $x$-axis as an example, it is computed as follows:
\begin{equation} 
 D_{xx}(\tilde{p}_h) = \frac{\mathcal{S}(\tilde{p}_{h0}^x) - 2\mathcal{S}(\tilde{p}_{h}) + \mathcal{S}(\tilde{p}_{h1}^x)}{\epsilon^2},
\end{equation}
    In addition, we employ the relaxing Eikonal term $\mathcal{L}_\text{re-eik}$ proposed by Wang et al.~\cite{hessian} instead of the regular Eikonal $\mathcal{L}_{\text{eik}}$, which allows other loss terms to play a more significant role and provides $\mathcal{S}$ with sufficient expressiveness:
\begin{equation}
\label{eq:re}
\mathcal{L}_\text{re-eik} = \frac{1}{N_s} \sum_{\tilde{p}_s^i\in P_s}
    \text{ReLU}\left(\sigma_{\text{min}} - \Vert \nabla S(\tilde{p}_s^i)\Vert\right) +
    \text{ReLU}\left(\Vert \nabla S(\tilde{p}_s^i)\Vert - \sigma_{\text{max}}\right) ,
\end{equation} 
where ReLU is the operator of $\text{max}(0, \cdot)$, and the thresholds $\sigma_{min}$ and $\sigma_{max}$ are set to 0.8 and 1.2, respectively. 

\subsection{Gaussian-guided Geometric Optimization}
Within the volume rendering framework, eliminating artifacts such as holes, gaps, and depressions obtained from a coarse octree is not straightforward. Simply subdividing the octree, as illustrated in Fig.~\ref{fig:gsp_corr} (left), may not effectively address these issues. We observe that the Gaussians $G_p$ progressively encircle the object surfaces throughout the GS optimization process. Motivated by this observation, we propose utilizing the point cloud $G_p$ generated from GS to improve geometry reconstruction. 
Upon obtaining the SDF represented by the coarse octree following stage 1, we employ the Marching Cube method~\cite{lorensen1998marching} to extract the zero isosurface, thereby generating a rough mesh $M_o$. We then use the positions of vertices on $M_o$ to initialize that of $G_p$ and optimized $G_p$ in 3DGS pipeline for 7k iterations. 

\textbf{GS Loss.} Although the optimized Gaussian centroids $G_p$ mostly surround the object's surface, few of them distribute on the surface. Thus, we cannot reconstruct the implicit surface directly from $G_p$. We define the following GS-points loss to make the reconstructed implicit surface just enclose the Gaussian centroids $G_p$,
\begin{equation} 
\label{eq:Gaussian_points}
\mathcal{L}_\text{gs} = \frac{1}{N_{g}}\sum_{i=1}^{N_g}\text{ReLU}(\mathcal{S}(\mathbf{p}_g^i) - \sigma_{g}),\quad \mathbf{p}^i_g\in G_p,
\end{equation}
where $N_g$ is the total number of Gaussian centroids, $\sigma_{g}$ is a relaxation term since there exists noise in $G_{p}$, we set $\sigma_{g} = -0.01$ in our experiments. In addition, the inclusion of the singular-Hessian term as Eq.(\ref{eq:hess}) can also facilitate the reduction of undesired surface variations in the geometry optimized from $G_p$, even though the majority of points in $G_p$ are not on the surface and exhibit noise, as depicted in Fig.~\ref{fig:nohe}. 

\textbf{Optimization.} Given the discrete and sparse nature of the octree grid,  we introduce a Laplacian loss term on SDF and a Total Variation loss on SH coefficients during the initialization stage.
Consequently, the loss  $\mathcal{L}_\text{init}$ in Stage 1 (Fig.~\ref{fig:pipeline}) can be summarized as:
 \begin{equation} 
 \label{eq:stage1}
 \begin{split}
&\mathcal{L}_\text{init} = \mathcal{L}_\text{col} + \lambda_{h}\mathcal{L}_\text{hess}+ \lambda_{e}\mathcal{L}_\text{re-eik}+\lambda_{l}\mathcal{L}_\text{lap}+\lambda_{t}\mathcal{L}_\text{tv},
\\
&\mathcal{L}_{\text{col}} = \frac{1}{N}\sum_{n=1}^N\left\|\mathbf{C}_n - \hat{\mathbf{C}}_n \right\|^2_2,
\\
&\mathcal{L}_{\text{lap}} = \frac{1}{N_s}\sum_{\tilde{p}_s^i\in P_s}\left\|D_{xx}^i+D_{yy}^i+D_{zz}^i \right\|^2_2,
\\
&\mathcal{L}_{\text{tv}} = \frac{1}{\left|\mathcal{V}\right|}\sum_{\mathbf{p}\in\mathcal{V}} \sqrt{(D_\mathbf{p}^x)^2+(D_\mathbf{p}^y)^2+(D_\mathbf{p}^z)^2},
\end{split}
\end{equation}
where $N$ is the number of pixels. $\mathcal{V}$ is the set of grid points. $\lambda_h$ and $\lambda_l$ are set to $10^{-9}$ and $10^{-10}$, respectively. $\lambda_e$ is set to $10^{-6}$. $\lambda_t$ is set in the range of $(10^{-5}, 10^{-3}]$, the brighter the scene, the larger $\lambda_{t}$ is recommended.
During the initialization phase, we optimize the parameters $\Theta$ and increase the octree depth from $l=6$ to $l=7$. 
Entering Stage 2, we use the vertices of $M_o$ as the initial Gaussian centroids $G_p$ and optimize it in 3DGS pipeline, as shown in Fig.~\ref{fig:pipeline}.
In Stage 3, we leverage the optimized $G_p$ to facilitate further optimization of the geometry.%
We conduct the Stage 3 with the loss $\mathcal{L}_\text{reco}$ as: 
 \begin{equation} 
\label{eq:stage3}
\mathcal{L}_\text{reco} = \mathcal{L}_\text{col} + \lambda_{h}\mathcal{L}_\text{hess}+ \lambda_{e}\mathcal{L}_\text{re-eik}+\lambda_{g}\mathcal{L}_\text{gs}+\lambda_{t}\mathcal{L}_\text{tv},
\end{equation}
where $\lambda_h$ is reduced from $10^{-9}$ to $10^{-13}$ and $\lambda_g$ decreases from $10^{-1}$ to $10^{-3}$ with the optimization step. TV term will be turned off after $l>8$. 

\subsection{SDF-guided 3D GS Optimization}
\label{sec:SDF-GS}
\begin{figure*}[t] 
    \centering
    \setlength\tabcolsep{1pt}
    \begin{footnotesize}
    \begin{tabular}{cccc}
    \includegraphics[width=\imglenfour\linewidth]{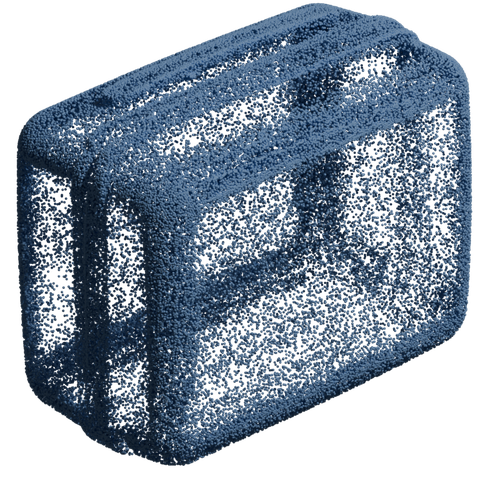}&
    \includegraphics[width=\imglenfour\linewidth]{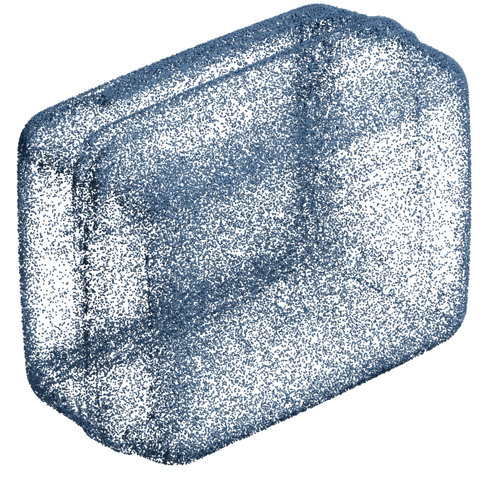}&
    \includegraphics[width=\imglenfour\linewidth]{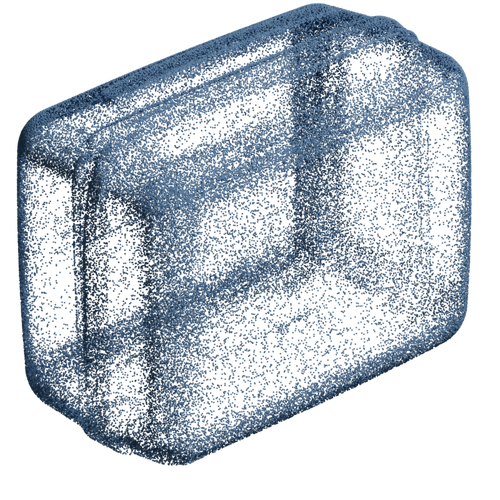}&
    \includegraphics[width=\imglenfour\linewidth]{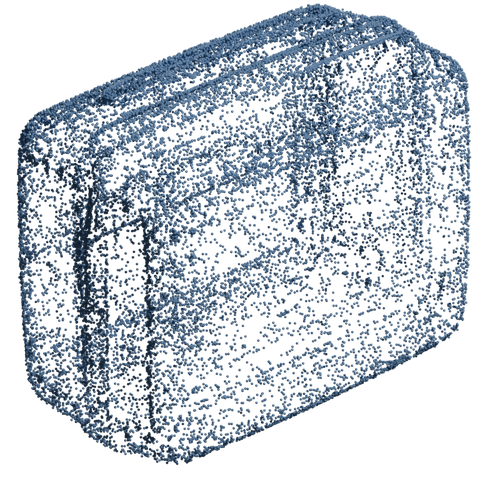}
    \\
    $N_{\mathrm{GS}}:$112.58 & $N_{\mathrm{GS}}:$98.20 & $N_{\mathrm{GS}}:$95.57 & $N_{\mathrm{GS}}:$\textbf{31.99}\\
    3DGS & 2DGS & GOF & Ours
    \end{tabular}
    \begin{tabular}{ccccc}
    \includegraphics[width=\imglenfive\linewidth]{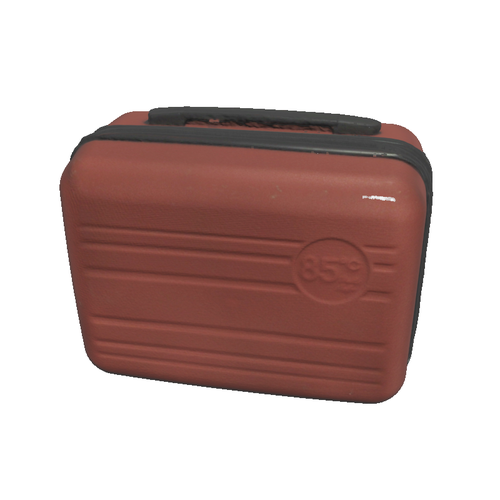}&
    \includegraphics[width=\imglenfive\linewidth]{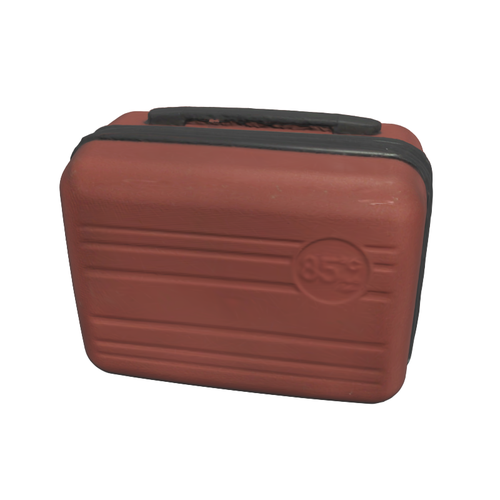}&
    \includegraphics[width=\imglenfive\linewidth]{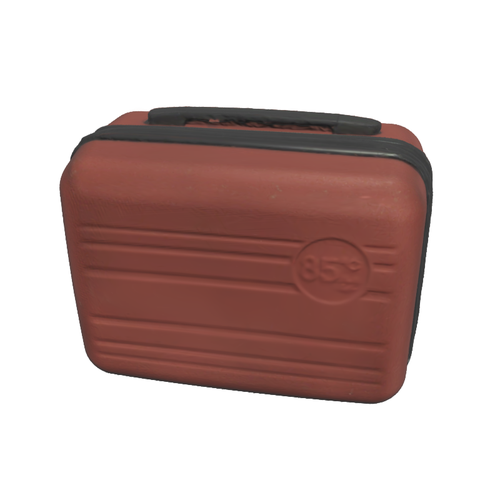}&
    \includegraphics[width=\imglenfive\linewidth]{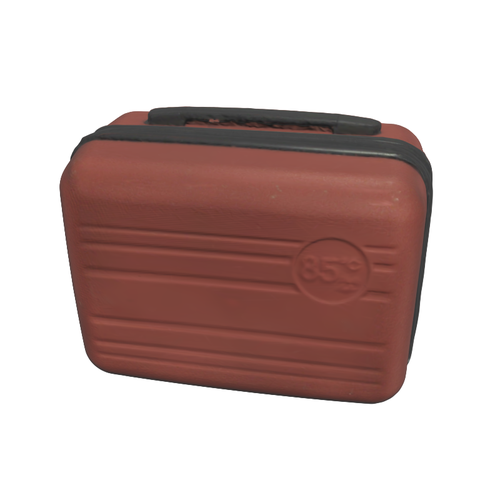}&
    \includegraphics[width=\imglenfive\linewidth]{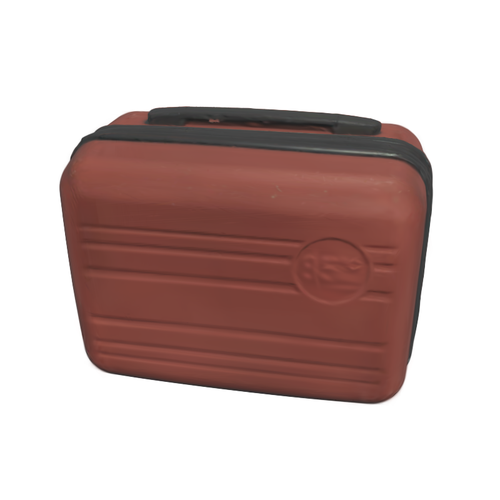}
    \\
    PSNR, FPS & 32.22, \textbf{219.98} & \textbf{33.42}, 69.86 & 32.23, 60.24 & 31.51, 193.84 \\
    GT image & 3DGS & 2DGS & GOF & Ours
    \end{tabular}
    \end{footnotesize}
    
    \caption{The visualization of 3DGS point clouds and the rendering results. The SDF successfully guides 3D Gaussians to concentrate near the object's surface and represent the object with fewer Gaussians. It also helps to resolve the Gaussian ``floaters'' which bring artefacts in rendering.}  
    \label{fig:SDF-GS}
\end{figure*}

\begin{table*}[b]\centering
    \renewcommand{\arraystretch}{0.8}
    \caption{Quantitative evaluation on the OO3D-SL dataset.}
    \label{tab:oo3d}
    \begin{tabular}{c|l|ccccccccc}
        \toprule
         &   &  Light & Ricecook  &  Sofa  & Suitcase &  Table   & Toy  & Avg \\
        \midrule
        \multirow{7}{*}{\textbf{PSNR}~($\uparrow$)}
        & Voxurf & 34.95 & 35.59 & \textbf{37.43} & 34.85 & 37.38 & 36.72 & 36.15 \\ 
        & NeuS2 & 35.44 & 35.83 & 36.36 & 35.71 & 38.26 & 36.57 & 36.36\\
        & 3DGS & 35.70 & 35.09 & 36.50 & 34.87 & 38.63 & 36.87 & 36.28\\
        & SuGaR & 35.05 & 33.12 & 35.07 & 35.44 & 33.86 & 32.91 & 34.24 \\
        & GOF & 35.55 & 35.31 & 36.49 & 34.81 & 38.66 & 36.85 & 36.28 \\
        & 2DGS & \textbf{36.28} & \textbf{36.09} & 35.58 & \textbf{35.90} & \textbf{39.38} & \textbf{37.58} & \textbf{36.80} \\
        & Ours & 34.39 & 34.02 & 35.23 & 33.79 & 37.58 & 36.18 & 35.18 \\
        \midrule
        \multirow{6}{*}{\textbf{CD}~($\downarrow$)}
        & Voxurf & 1.36  & 15.15 & 1.64  & \textbf{0.60}  & 2.00  & \textbf{0.94}  & 3.62\\ 
        & NeuS2  & 1.24  & 1.83  & 2.42  & 0.76  & \textbf{1.37}  & 1.70 & 1.55\\ 
        & SuGaR  & 13.61 & 26.58 & 13.45 & 16.04 & 21.34 & 1.55 & 15.43\\ 
        & GOF    & 1.28  & 7.19  & 2.65  & 1.14  & 4.93  & 1.24  & 3.07\\ 
        & 2DGS   & 3.89  & 1.56  & 2.96  & 3.86  & 17.16 & 1.68  & 5.18\\  
        & Ours   & \textbf{0.97}  & \textbf{0.55}  & \textbf{0.18}  & 0.86  & 1.58  & 1.30  & \textbf{0.91}\\ 
        \midrule
        \multirow{7}{*}{\textbf{FPS}~($\uparrow$)}
        & Voxurf & 2.16 & 2.06 & 2.05 & 2.08 & 2.29 & 1.68 & 2.05 \\ 
        & NeuS2 & 9.19 & 8.60 & 9.79 & 8.42 & 6.37 & 13.00 & 9.23\\
        & 3DGS & 142.83 & \textbf{217.97} & 200.13 & \textbf{183.90} & \textbf{165.14} & \textbf{328.51} & \textbf{206.41}\\
        & SuGaR & 31.79 & 31.71 & 31.88 & 32.30 & 36.33 & 33.43 & 32.89 \\
        & GOF & 107.54 & 91.09 & 76.37 & 63.68 & 58.24 & 173.26 & 95.03 \\
        & 2DGS & 68.47 & 62.24 & 65.02 & 63.14 & 62.32 & 77.46 & 66.44 \\
        & Ours & \textbf{205.55} & 178.29 & \textbf{209.90} & 170.30 & 148.75 & 298.66 & 201.91 \\
        \bottomrule
    \end{tabular}
\end{table*}

After we have got a fine-grained SDF from octree, we can utilize it to guide the optimization of 3D Gaussians further. From the observations in SuGaR~\cite{sugar} and NeuSG~\cite{neusg}, the position of 3D Gaussians should be close to the surface of objects. Moreover, the shape of these 3D Gaussians should have a thin-plate shape, representing the colour of the nearby surface. Therefore, we applied special regularization terms to utilize the observations. 

\textbf{Opacity Loss.} VolSDF~\cite{volsdf} adopts an S-shaped density function to establish a correlation between opacity and signed distances for positions in space. For points near the surface, the density value should be around 0.5. For points outside the object and away from it, the opacity should be close to 0. For points buried deep inside the object, the opacity should be 1. This perspective also applies to the distribution of 3D Gaussians. Intuitively, the 3D Gaussians far away from the surface should be transparent, while the ones closer to the surface should be opaque. Moreover, there should be only a few Gaussians buried deeply inside the object since these interior Gaussians do not contribute as much as those near the surface. Therefore, we propose the following function to estimate opacity from a given signed distance value:
\begin{equation} 
\label{eq:opacity_SDF}
\mathcal{\alpha}_\text{SDF} = 
\begin{cases}
    \frac{4e^{-k\mathcal{S}(\mathbf{p}_i)}}{(1+e^{-k\mathcal{S}(\mathbf{p}_i)})^2} & \text{if } |\mathcal{S}(\mathbf{p}_i)| < \eta, \\
    0 & \text{if } |\mathcal{S}(\mathbf{p}_i)| \geq \eta,
\end{cases}
\end{equation}
where $\mathcal{S}(\mathbf{p}_i)$ is the SDF value at position $\mathbf{p}_i$, and $k$ is a scaling factor, $\eta$ is the threshold value of SDF. Such an estimated opacity has a value range of $[0, 1]$, and takes 1 when the 3D Gaussians are on the surface. We truncated the estimated opacity values to 0 when Gaussian positions were too far away from the surface, and we set the threshold $\eta = 0.05$ in our experiments. According to this estimation, we proposed the following opacity loss term to guide the optimization of 3D Gaussians:
\begin{equation}
\label{eq:opacity_reg}
\mathcal{L}_\text{op} = \lVert \alpha - \alpha_\text{SDF} \rVert_2.
\end{equation}

\textbf{Scale Loss.} In 3D Gaussian representations, the shape of 3D Gaussian is defined by the position $\mathbf{p} \in \mathbb{R}^{3}$, rotation quaternion $\mathbf{Q} \in \mathbb{R}^{4}$ and the scaling factors $\mathbf{S} \in \mathbb{R}^{3}$. In 3DGS~\cite{3Dgaussians}, many 3D Gaussians are located within the object surface, thus contributing little to the rendering. Therefore, to encourage the 3D Gaussians to have a thin-plate shape and close to the surface, we apply the following regularization terms proposed by~\cite{neusg}: 
\begin{equation}
\label{eq:scale_reg}
\mathcal{L}_\text{scale} = \lVert \min(s_1, s_2, s_3) \rVert, 
\end{equation}

where $\mathbf{S} = \{s_1, s_2, s_3\}$ indicate the 3-dimensional scale factors. 
We try to minimize the smallest scale factor by Eq. (\ref{eq:scale_reg}), thus forcing the shape of the 3D Gaussians to be more flat. With the supervision of these 2 additional losses, we reduce the number of 3D Gaussians needed to represent an object, while preserving the quality of rendered images, as shown in Fig. (\ref{fig:SDF-GS}). In this final stage (see Stage 4 in Fig. (\ref{fig:pipeline})), the loss function we use is the following:
\begin{equation}
\label{eq:loss_third}
\begin{split}
    \mathcal{L}_\text{SDF-GS} = &\mathcal{L}_\text{1} + (1-\lambda_\text{ssim})\mathcal{L}_\text{D-SSIM} + \lambda_\text{op}\mathcal{L}_\text{op} \\ &+  \lambda_\text{scale}\mathcal{L}_\text{scale}.
\end{split}
\end{equation}
Here we set the default parameters $\lambda_\text{ssim} = 0.1$, $\lambda_\text{op} = 3$, and $\lambda_\text{scale} = 0.1$.

\section{Experiments}

\begin{figure*}[t] 
    \centering
    \setlength\tabcolsep{1pt}
    \begin{footnotesize}
    \begin{tabular}{cccccccc}
    \includegraphics[width=\imgleneight\textwidth]{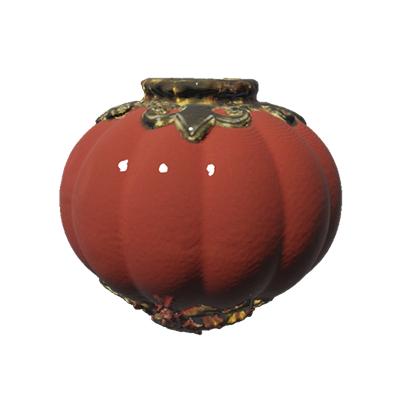}&
    \includegraphics[width=\imgleneight\textwidth]{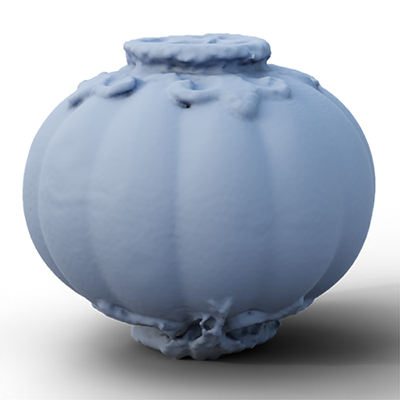}&
    \includegraphics[width=\imgleneight\textwidth]{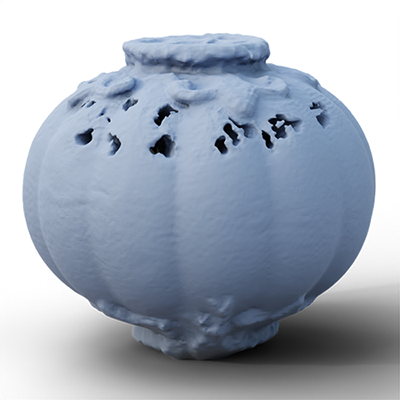}&
    \includegraphics[width=\imgleneight\textwidth]{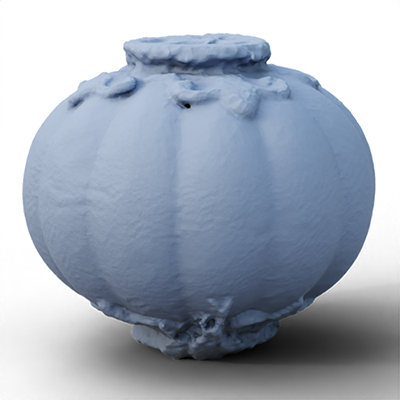}&
    \includegraphics[width=\imgleneight\textwidth]{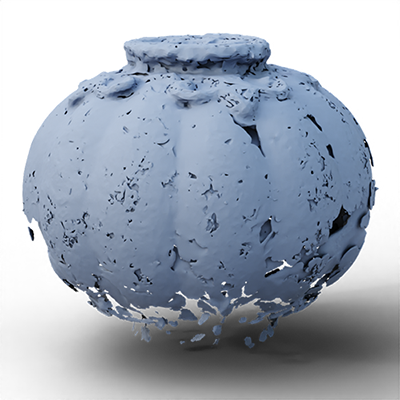}&
    \includegraphics[width=\imgleneight\textwidth]{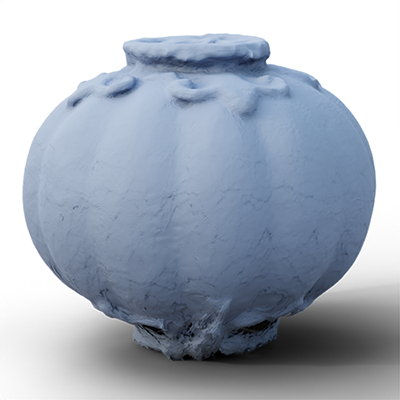}&
    \includegraphics[width=\imgleneight\textwidth]{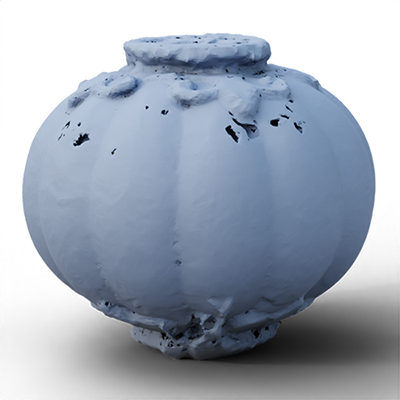}&
    \includegraphics[width=\imgleneight\textwidth]{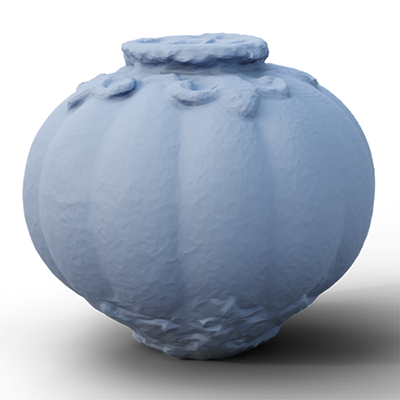}
    \\
    \includegraphics[width=\imgleneight\textwidth]{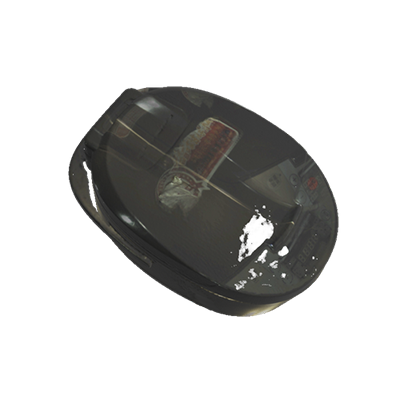}&
    \includegraphics[width=\imgleneight\textwidth]{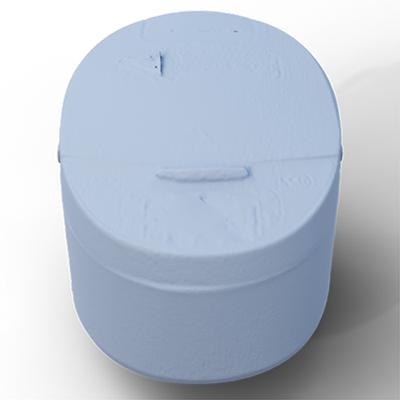}&
    \includegraphics[width=\imgleneight\textwidth]{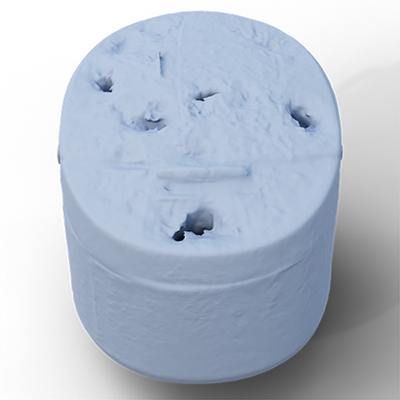}&
    \includegraphics[width=\imgleneight\textwidth]{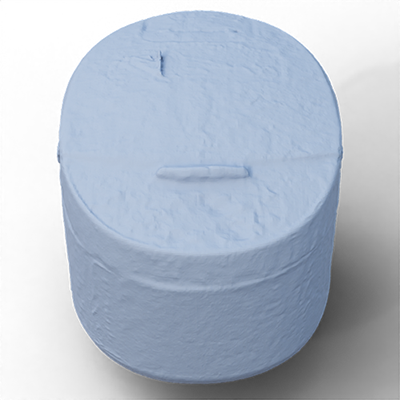}&
    \includegraphics[width=\imgleneight\textwidth]{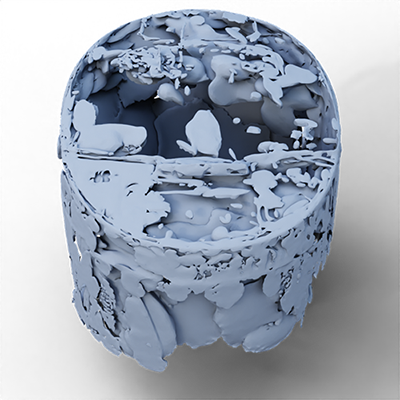}&
    \includegraphics[width=\imgleneight\textwidth]{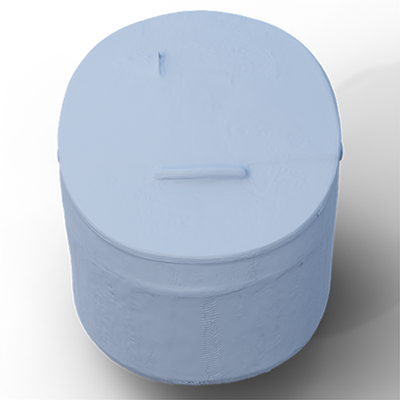}&
    \includegraphics[width=\imgleneight\textwidth]{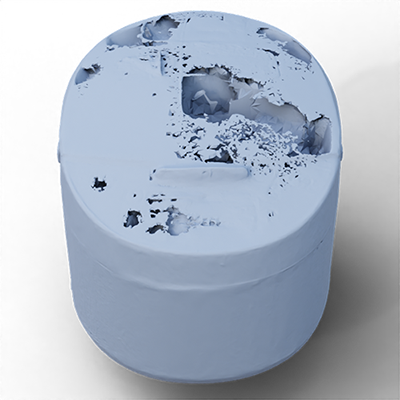}&
    \includegraphics[width=\imgleneight\textwidth]{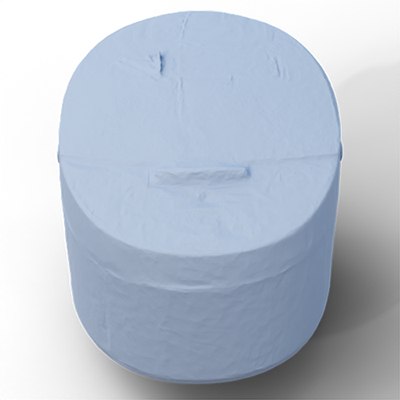}
    \\
    \includegraphics[width=\imgleneight\textwidth]{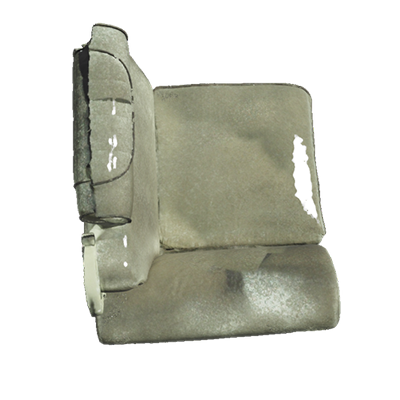}&
    \includegraphics[width=\imgleneight\textwidth]{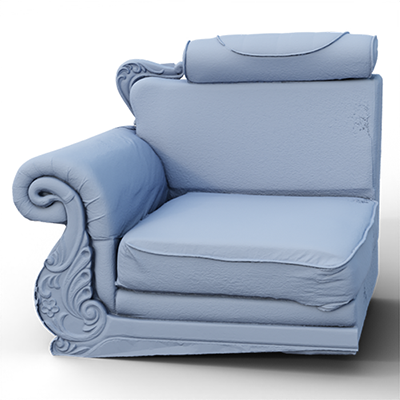}&
    \includegraphics[width=\imgleneight\textwidth]{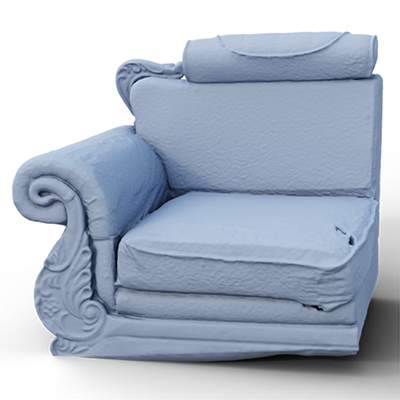}&
    \includegraphics[width=\imgleneight\textwidth]{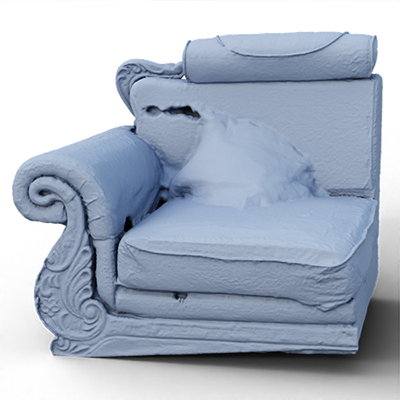}&
    \includegraphics[width=\imgleneight\textwidth]{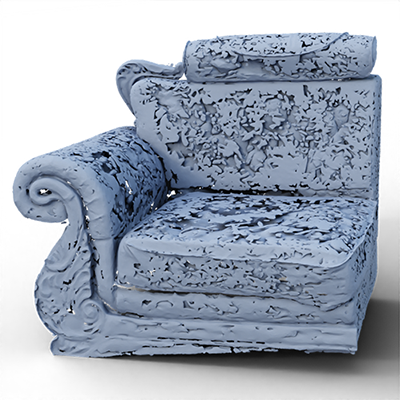}&
    \includegraphics[width=\imgleneight\textwidth]{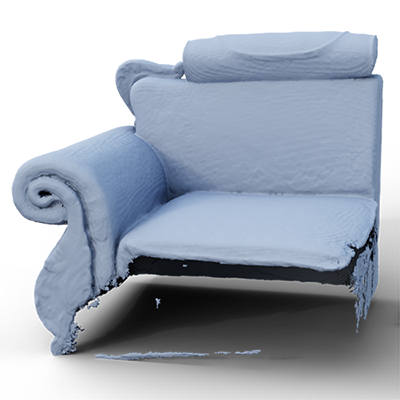}&
    \includegraphics[width=\imgleneight\textwidth]{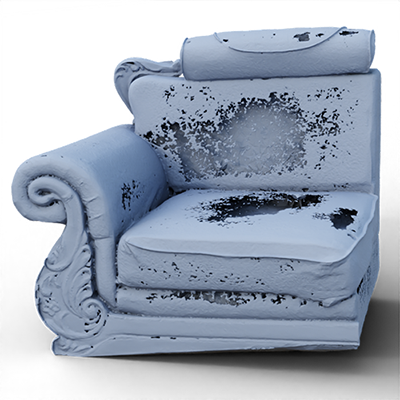}&
    \includegraphics[width=\imgleneight\textwidth]{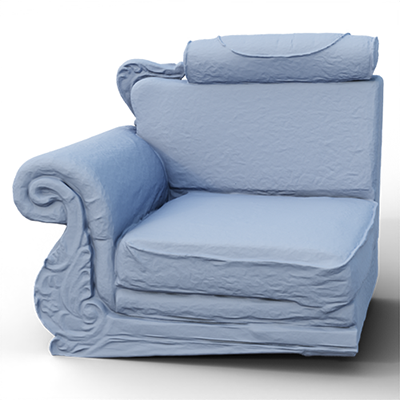}
    \\
    \includegraphics[width=\imgleneight\textwidth]{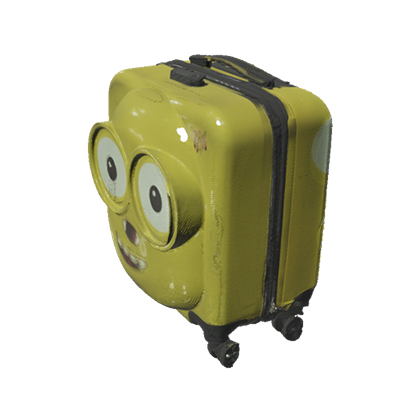}&
    \includegraphics[width=\imgleneight\textwidth]{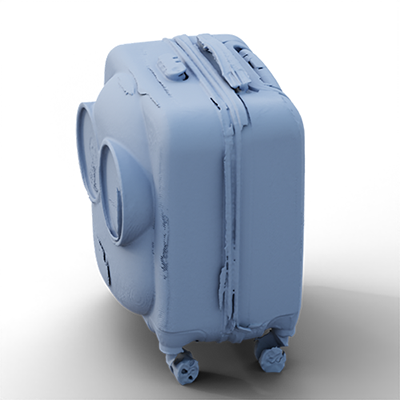}&
    \includegraphics[width=\imgleneight\textwidth]{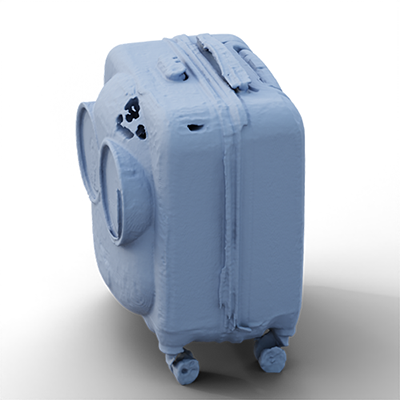}&
    \includegraphics[width=\imgleneight\textwidth]{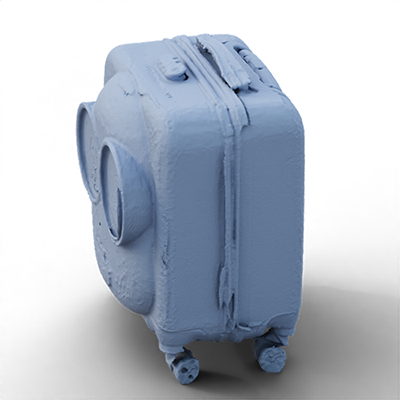}&
    \includegraphics[width=\imgleneight\textwidth]{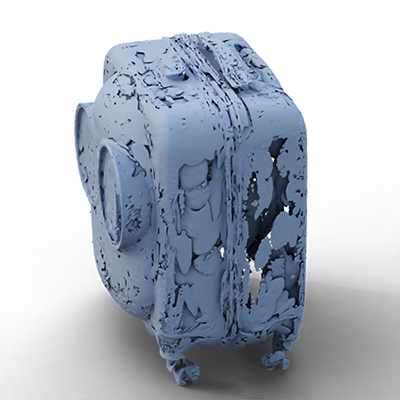}&
    \includegraphics[width=\imgleneight\textwidth]{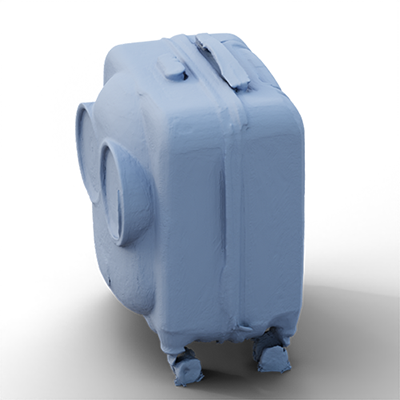}&
    \includegraphics[width=\imgleneight\textwidth]{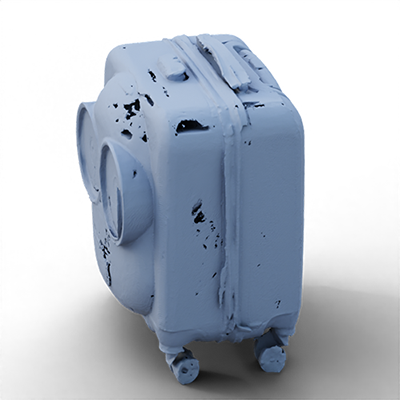}&
    \includegraphics[width=\imgleneight\textwidth]{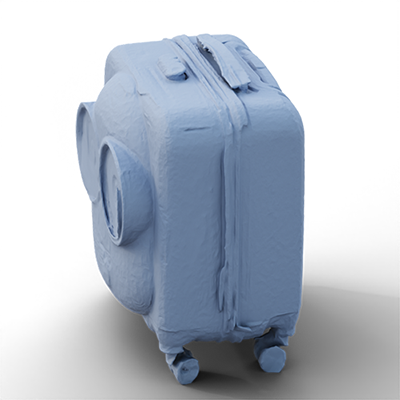}
    \\
    \includegraphics[width=\imgleneight\textwidth]{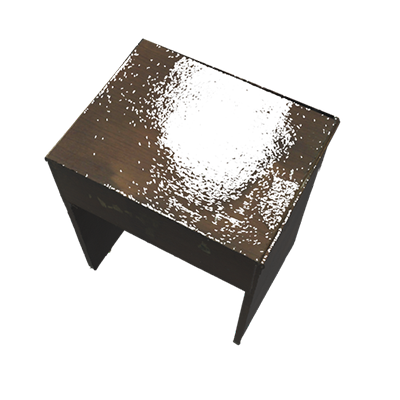}&
    \includegraphics[width=\imgleneight\textwidth]{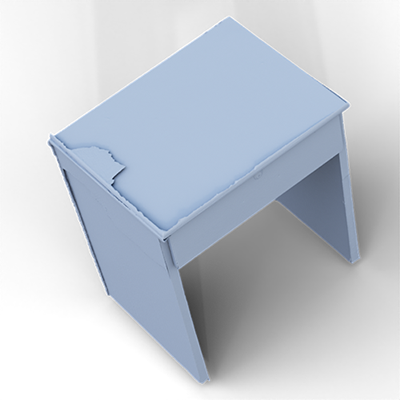}&
    \includegraphics[width=\imgleneight\textwidth]{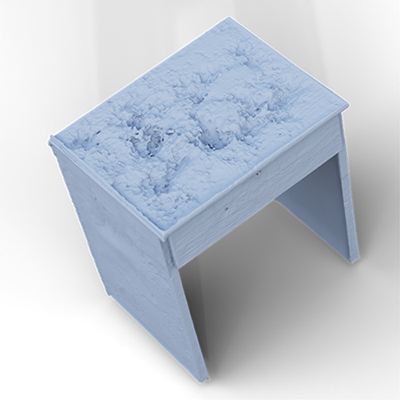}&
    \includegraphics[width=\imgleneight\textwidth]{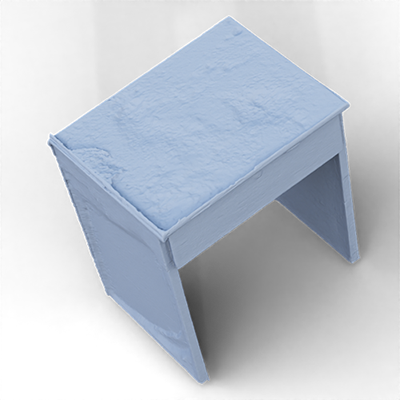}&
    \includegraphics[width=\imgleneight\textwidth]{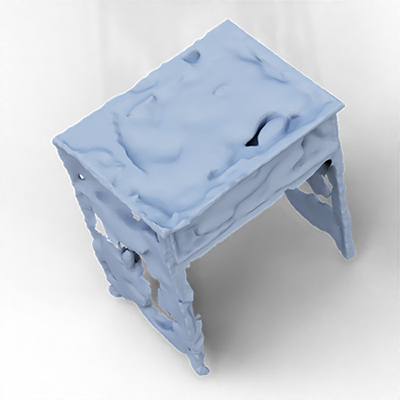}&
    \includegraphics[width=\imgleneight\textwidth]{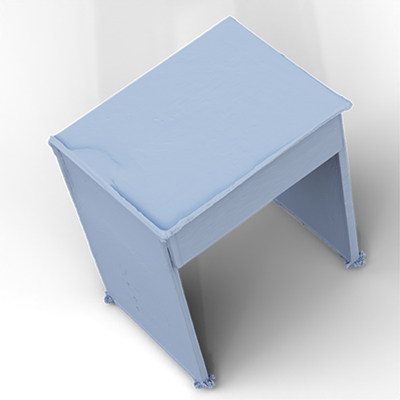}&
    \includegraphics[width=\imgleneight\textwidth]{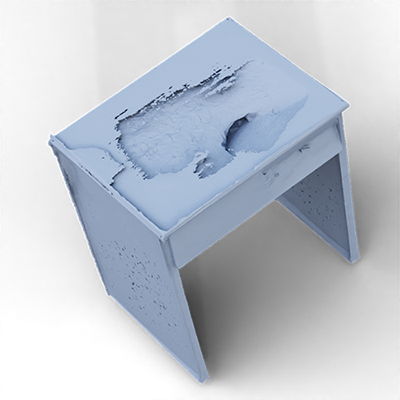}&
    \includegraphics[width=\imgleneight\textwidth]{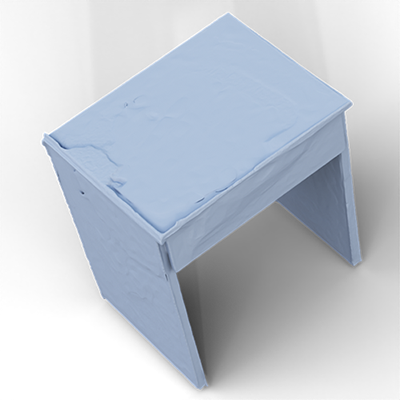}
    \\
    \includegraphics[width=\imgleneight\textwidth]{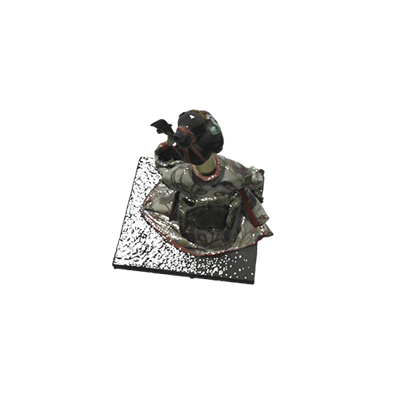}&
    \includegraphics[width=\imgleneight\textwidth]{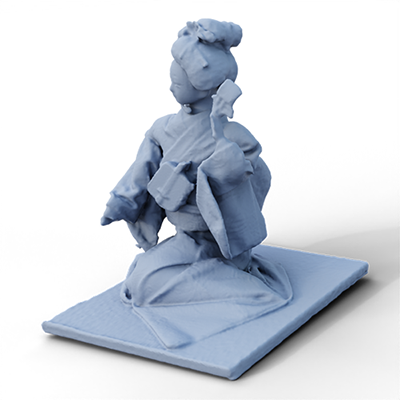}&
    \includegraphics[width=\imgleneight\textwidth]{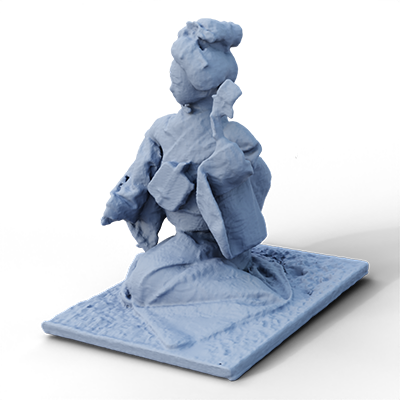}&
    \includegraphics[width=\imgleneight\textwidth]{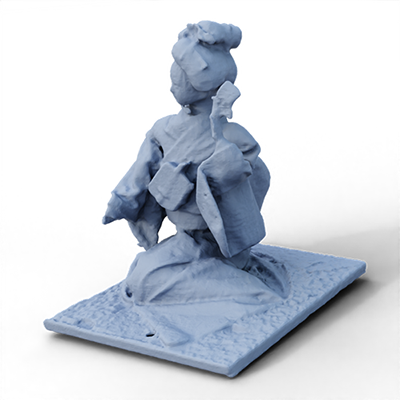}&
    \includegraphics[width=\imgleneight\textwidth]{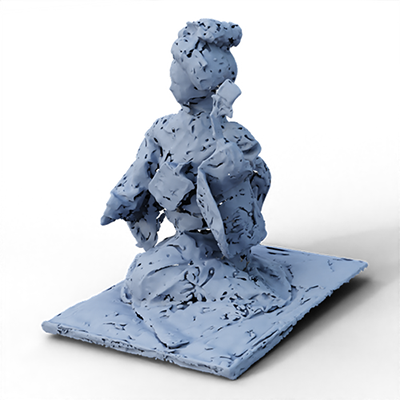}&
    \includegraphics[width=\imgleneight\textwidth]{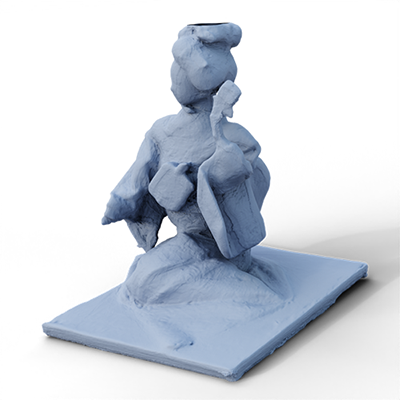}&
    \includegraphics[width=\imgleneight\textwidth]{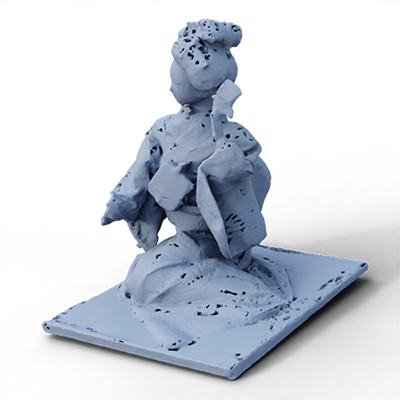}&
    \includegraphics[width=\imgleneight\textwidth]{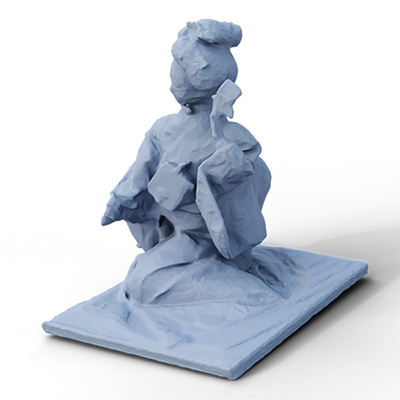}
    \\
    GT image & GT mesh & Voxurf & NeuS2 & SuGaR & 2DGS & GOF & Ours\\
    \end{tabular}
    \end{footnotesize}
      \caption{Visual comparison of geometric reconstruction results on the OO3D-SL dataset.} \label{fig:mesh_render}
\end{figure*}

\begin{figure*}[t] 
    \centering
    \setlength\tabcolsep{1pt}
    \begin{footnotesize}
    \begin{tabular}{cccccccc}
    \includegraphics[width=\imgleneight\textwidth]{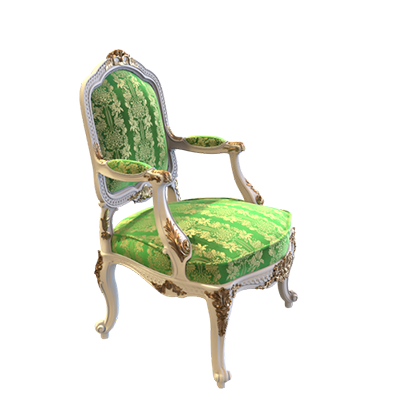}&
    \includegraphics[width=\imgleneight\textwidth]{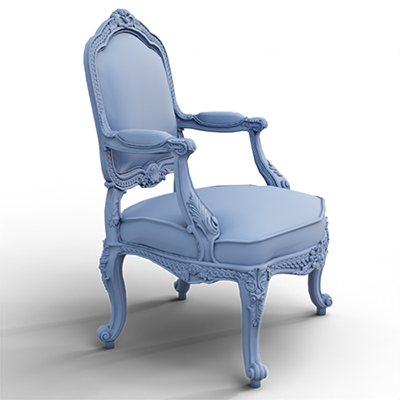}&
    \includegraphics[width=\imgleneight\textwidth]{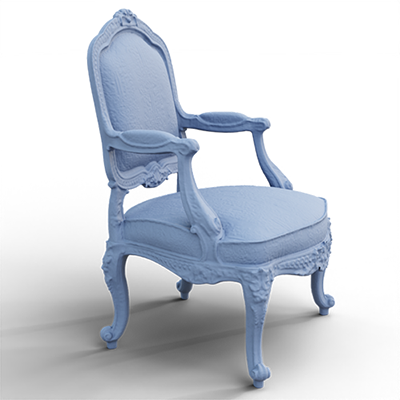}&
    \includegraphics[width=\imgleneight\textwidth]{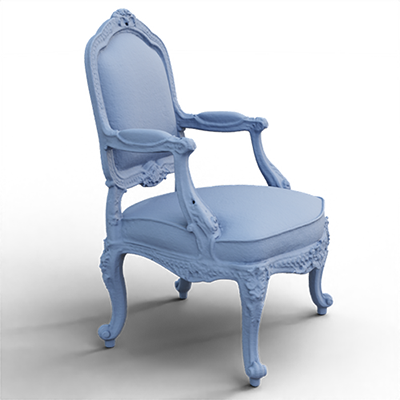}&
    \includegraphics[width=\imgleneight\textwidth]{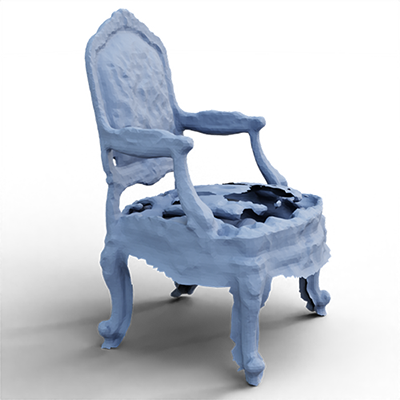}&
    \includegraphics[width=\imgleneight\textwidth]{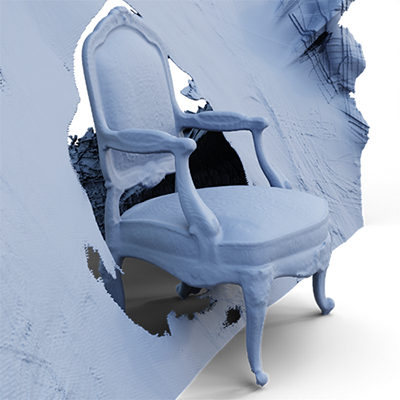}&
    \includegraphics[width=\imgleneight\textwidth]{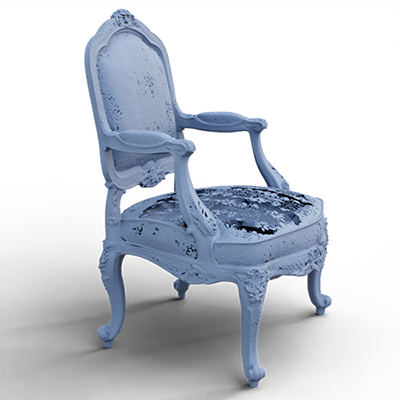}&
    \includegraphics[width=\imgleneight\textwidth]{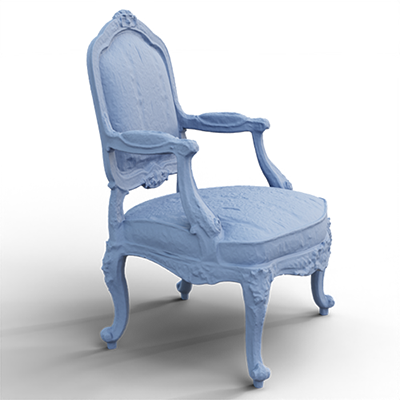}
    \\
    \includegraphics[width=\imgleneight\textwidth]{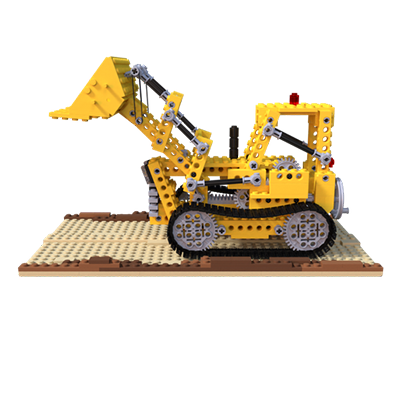}&
    \includegraphics[width=\imgleneight\textwidth]{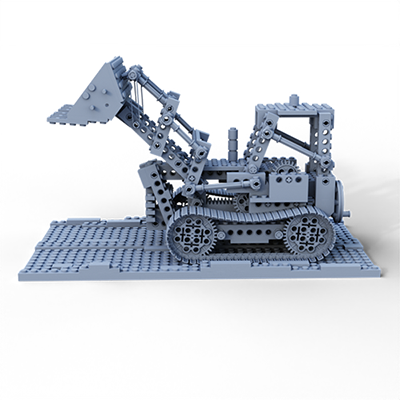}&
    \includegraphics[width=\imgleneight\textwidth]{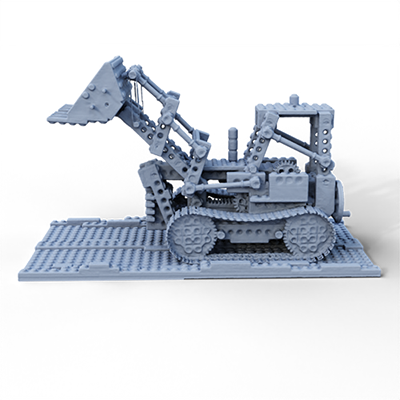}&
    \includegraphics[width=\imgleneight\textwidth]{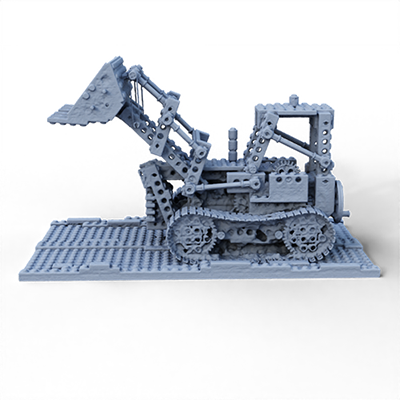}&
    \includegraphics[width=\imgleneight\textwidth]{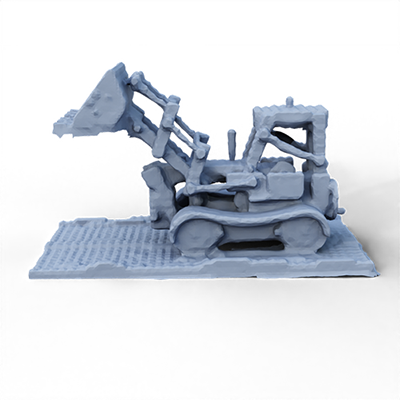}&
    \includegraphics[width=\imgleneight\textwidth]{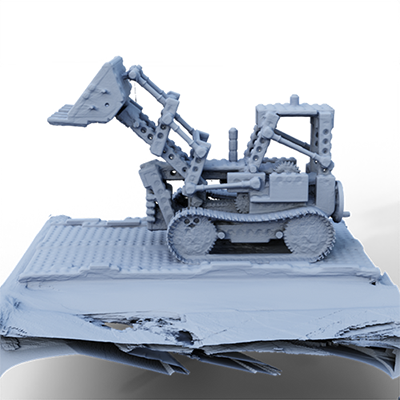}&
    \includegraphics[width=\imgleneight\textwidth]{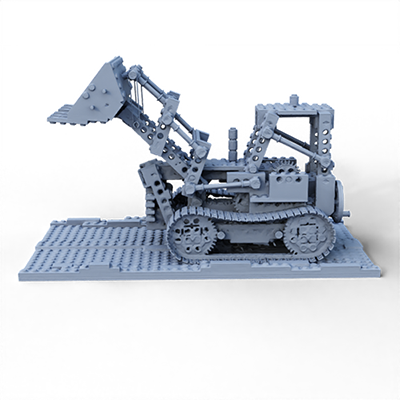}&
    \includegraphics[width=\imgleneight\textwidth]{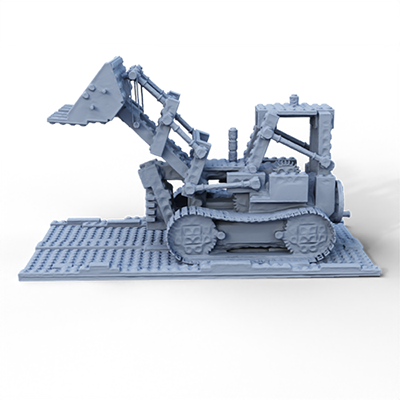}
    \\
    \includegraphics[width=\imgleneight\textwidth]{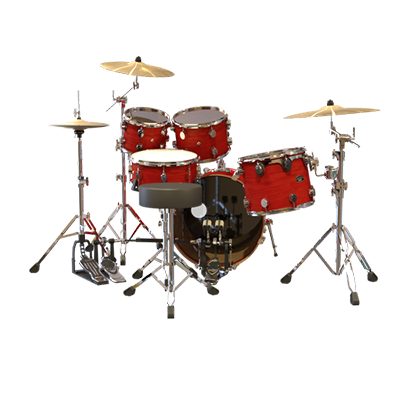}&
    \includegraphics[width=\imgleneight\textwidth]{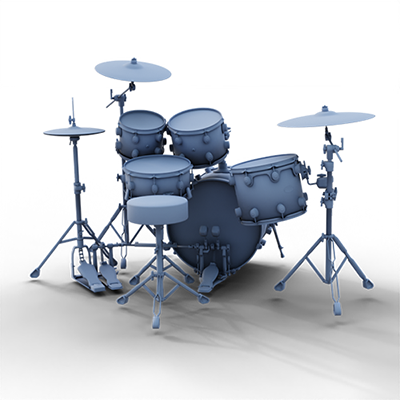}&
    \includegraphics[width=\imgleneight\textwidth]{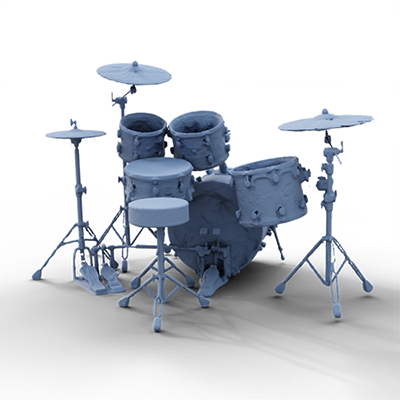}&
    \includegraphics[width=\imgleneight\textwidth]{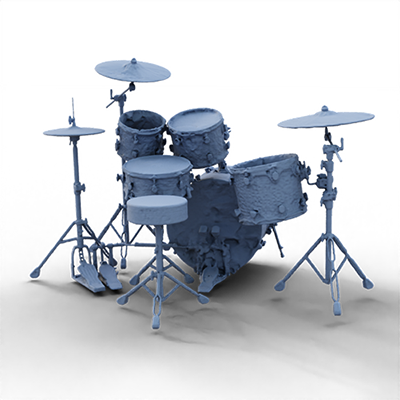}&
    \includegraphics[width=\imgleneight\textwidth]{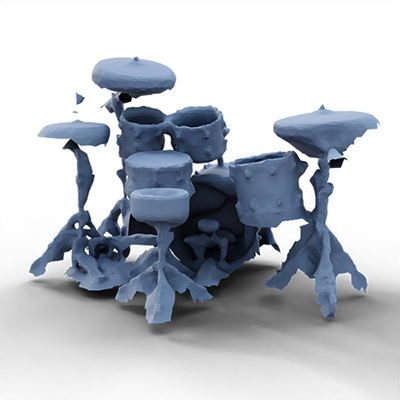}&
    \includegraphics[width=\imgleneight\textwidth]{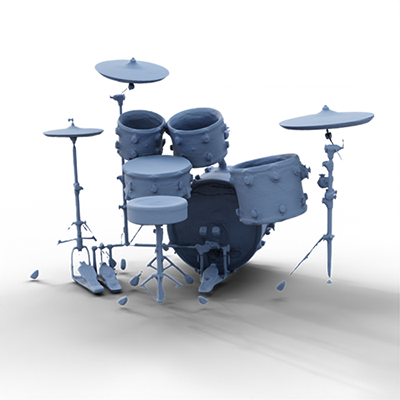}&
    \includegraphics[width=\imgleneight\textwidth]{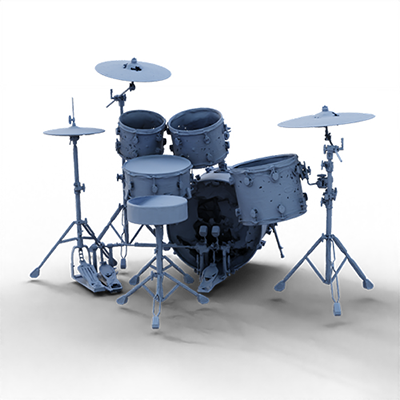}&
    \includegraphics[width=\imgleneight\textwidth]{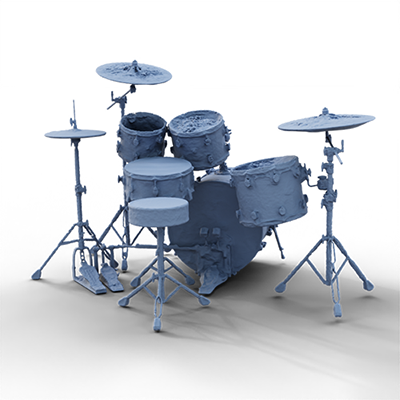}
    \\
    \includegraphics[width=\imgleneight\textwidth]{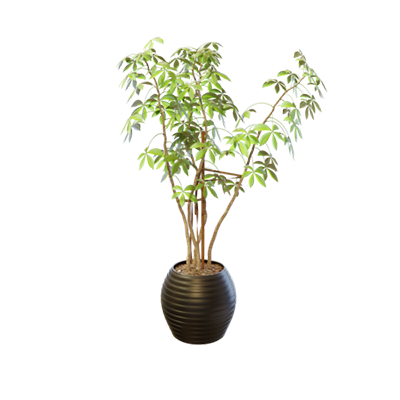}&
    \includegraphics[width=\imgleneight\textwidth]{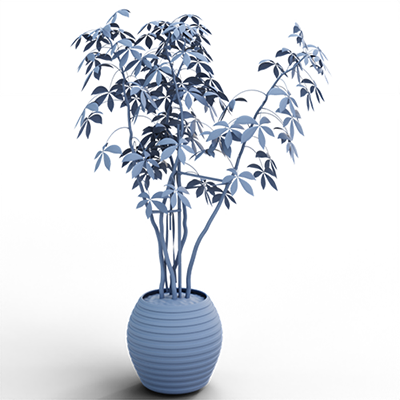}&
    \includegraphics[width=\imgleneight\textwidth]{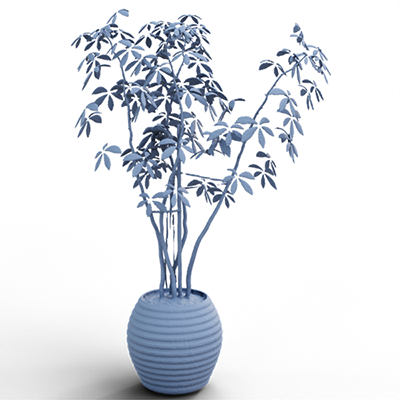}&
    \includegraphics[width=\imgleneight\textwidth]{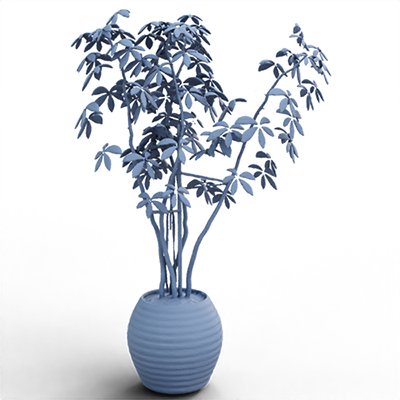}&
    \includegraphics[width=\imgleneight\textwidth]{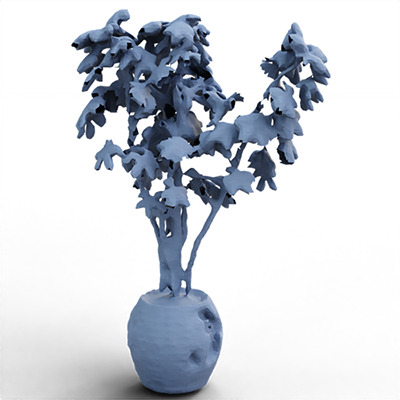}&
    \includegraphics[width=\imgleneight\textwidth]{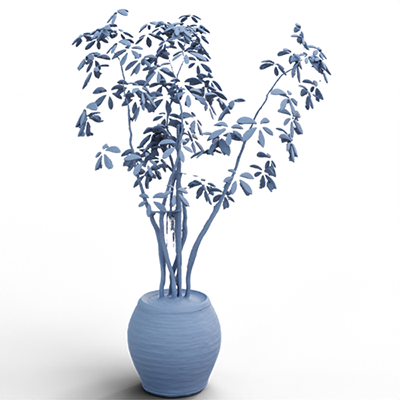}&
    \includegraphics[width=\imgleneight\textwidth]{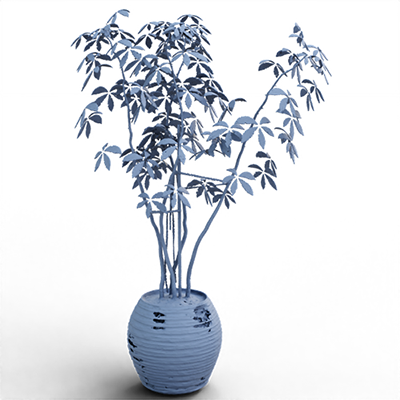}&
    \includegraphics[width=\imgleneight\textwidth]{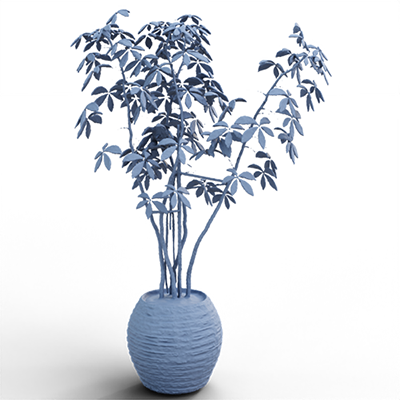}
    \\
    GT image & GT mesh & Voxurf & NeuS2 & SuGaR & 2DGS & GOF & Ours\\
    \end{tabular}
    
    \end{footnotesize}
      \caption{Visual comparison of geometric reconstruction results on the NeRF-Synthetic dataset.}  \label{fig:nerf}
\end{figure*}

\begin{figure*}[t] 
\fontsize{6pt}{6pt}\selectfont
    \centering
    \setlength\tabcolsep{1pt}
    \begin{scriptsize}
    \begin{tabular}{ccccccc}
    \includegraphics[width=\imglenseven\textwidth]{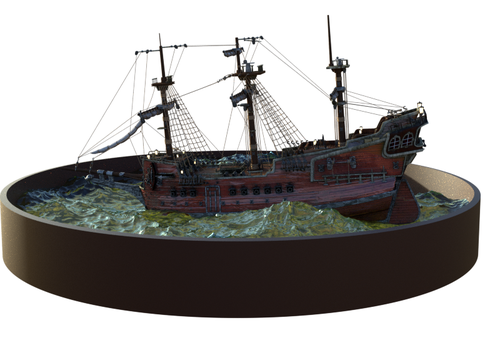}&
    \includegraphics[width=\imglenseven\textwidth]{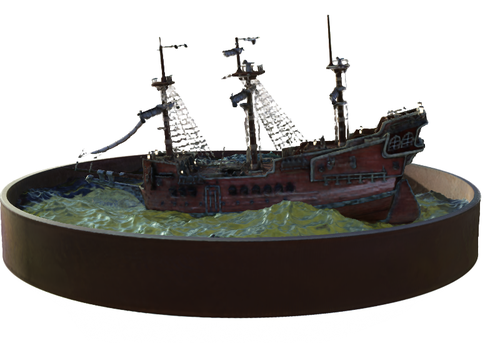}&
    \includegraphics[width=\imglenseven\textwidth]{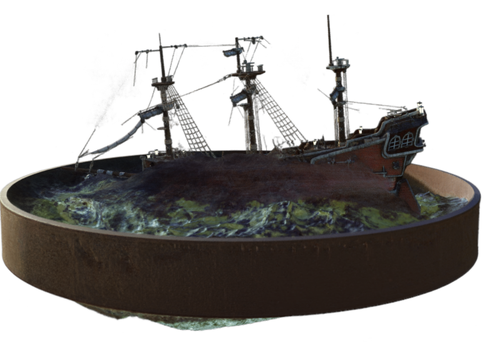}&
    \includegraphics[width=\imglenseven\textwidth]{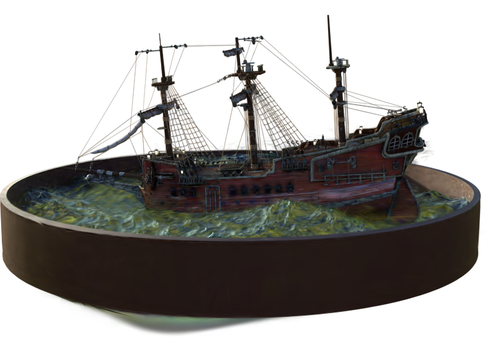}&
    \includegraphics[width=\imglenseven\textwidth]{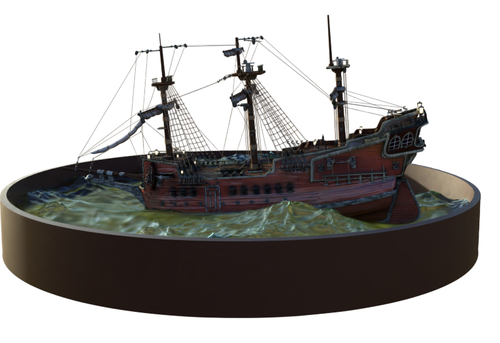}&
    \includegraphics[width=\imglenseven\textwidth]{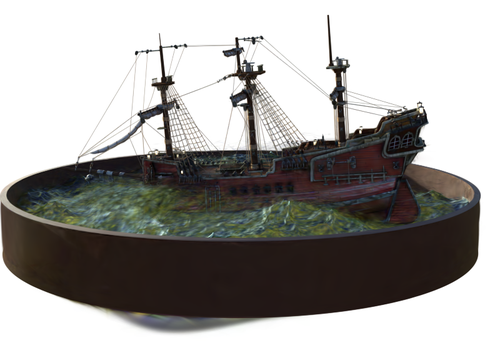}&
    \includegraphics[width=\imglenseven\textwidth]{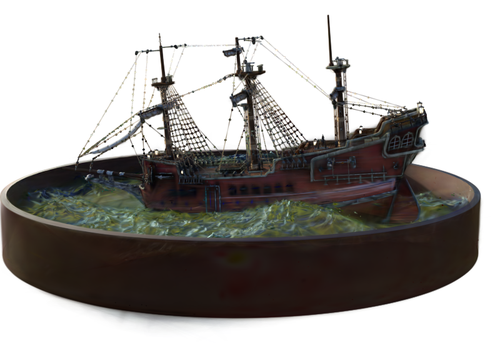}
    \\ %
    PSNR, FPS, $N_{\mathrm{GS}}$ & 28.73, 1.37, - & 28.59, 2.99, - & 31.01, \textbf{342.42}, 327.58 & \textbf{33.07}, 42.19, \textbf{150.22} & 30.68, 50.44, 271.90 & 29.05, 235.95, 244.10 \\
    \includegraphics[width=\imglenseven\textwidth]{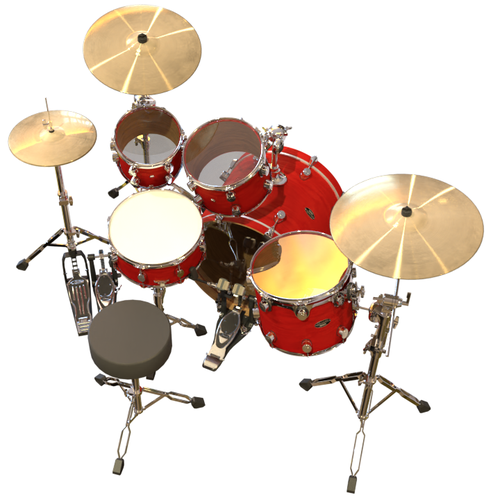}&
    \includegraphics[width=\imglenseven\textwidth]{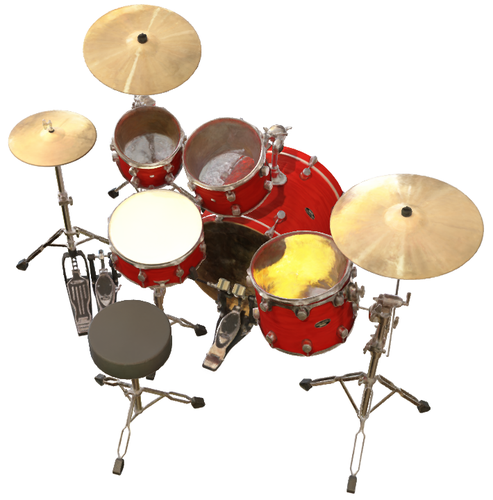}&
    \includegraphics[width=\imglenseven\textwidth]{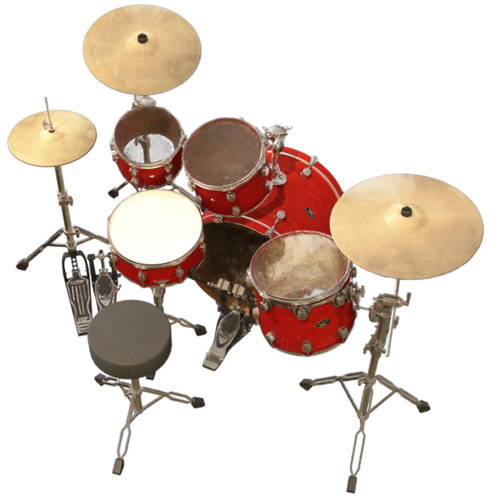}&
    \includegraphics[width=\imglenseven\textwidth]{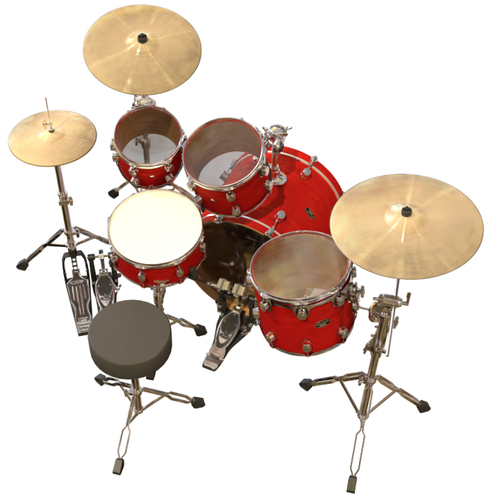}&
    \includegraphics[width=\imglenseven\textwidth]{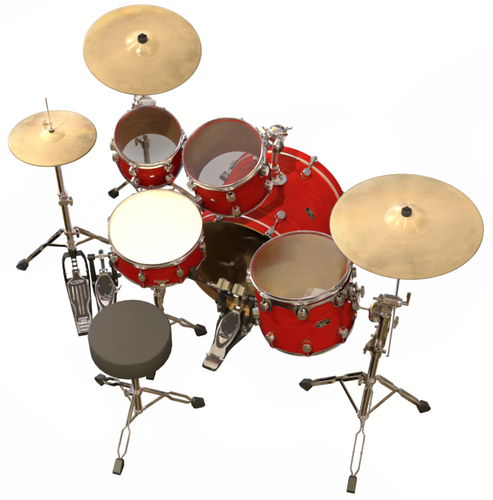}&
    \includegraphics[width=\imglenseven\textwidth]{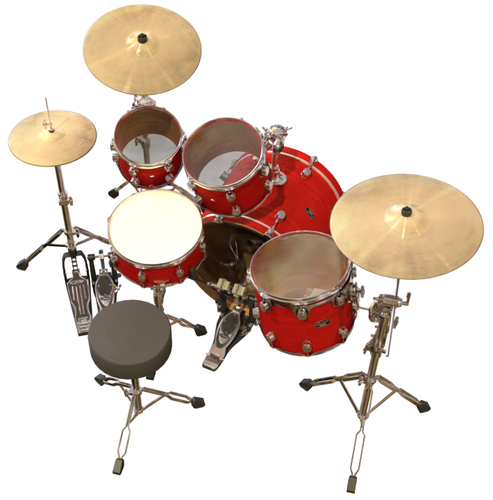}&
    \includegraphics[width=\imglenseven\textwidth]{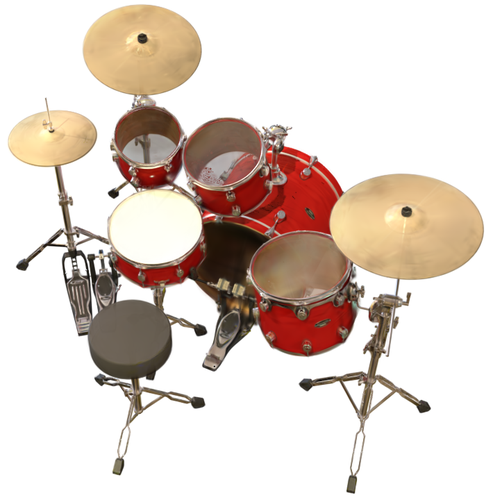}
    \\ %
    PSNR, FPS, $N_{\mathrm{GS}}$ & 25.63, 1.51, - & 25.18, 4.37, - & 26.22, \textbf{281.01}, 346.98 & \textbf{28.30}, 44.67, \textbf{155.94} & 26.22, 91.71, 200.72 & 25.52, 232.22, 207.28 \\
    \includegraphics[width=\imglenseven\textwidth]{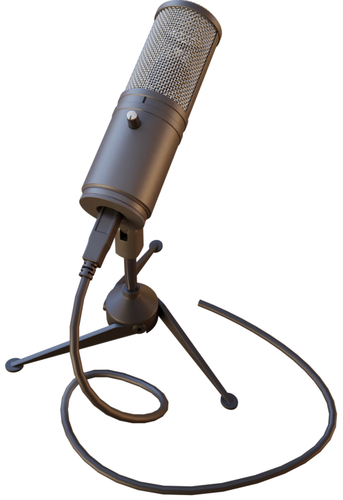}&
    \includegraphics[width=\imglenseven\textwidth]{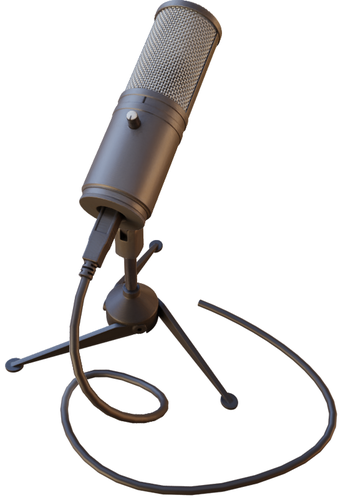}&
    \includegraphics[width=\imglenseven\textwidth]{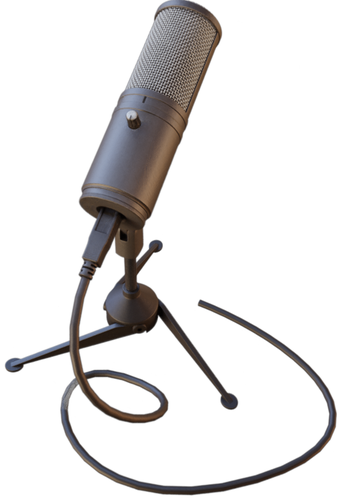}&
    \includegraphics[width=\imglenseven\textwidth]{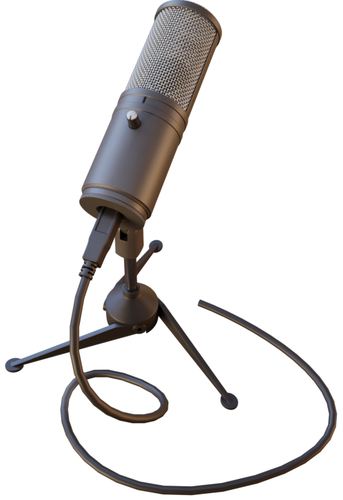}&
    \includegraphics[width=\imglenseven\textwidth]{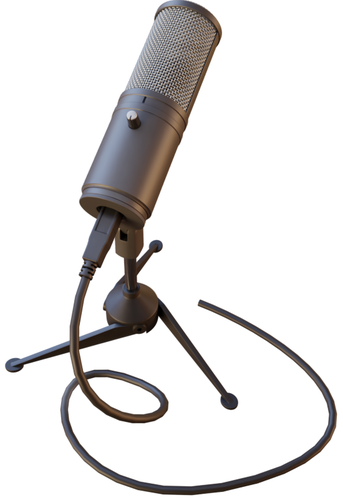}&
    \includegraphics[width=\imglenseven\textwidth]{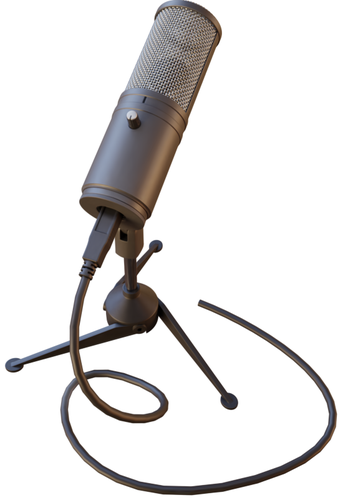}&
    \includegraphics[width=\imglenseven\textwidth]{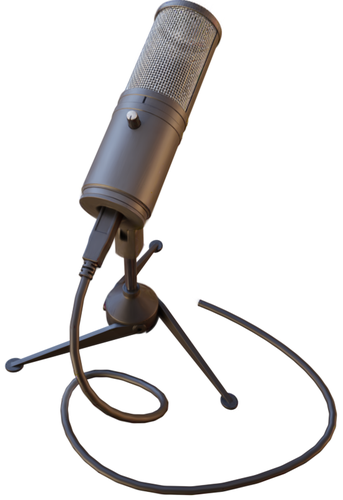}
    \\ %
    PSNR, FPS, $N_{\mathrm{GS}}$ & 34.20, 1.82, - & 34.25, 9.17, - & 35.42, \textbf{290.81}, 312.17 & \textbf{37.74}, 45.27, 142.75 & 35.65, 59.18, 217.07 & 32.57, 219.71, \textbf{110.43}\\
    
    \midrule
    
    \includegraphics[width=\imglenseven\textwidth]{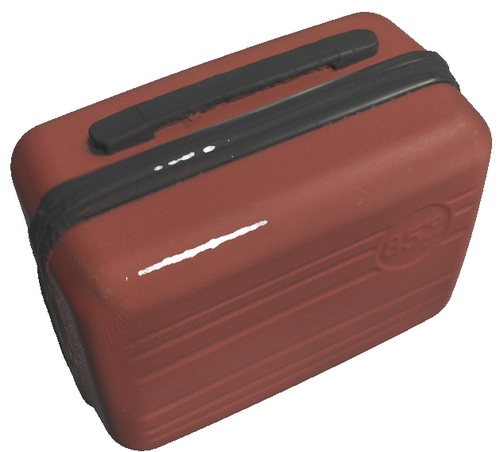}&
    \includegraphics[width=\imglenseven\textwidth]{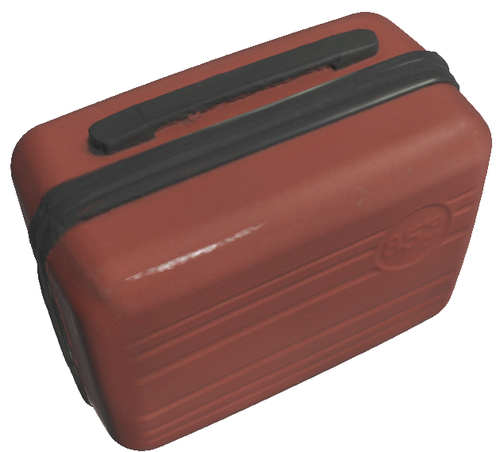}&
    \includegraphics[width=\imglenseven\textwidth]{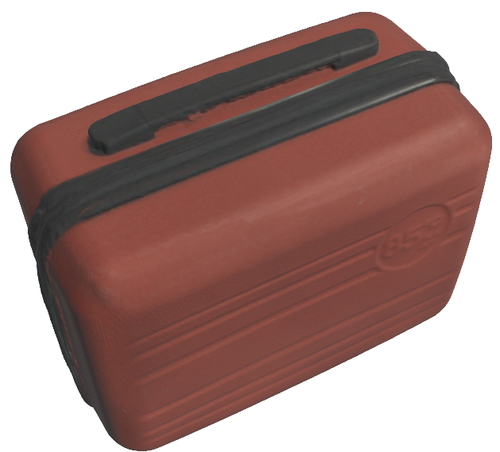}&
    \includegraphics[width=\imglenseven\textwidth]{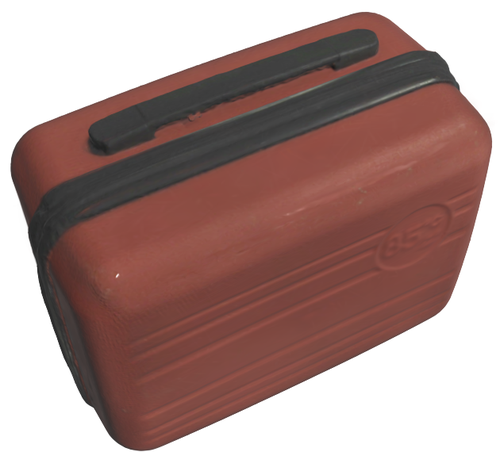}&
    \includegraphics[width=\imglenseven\textwidth]{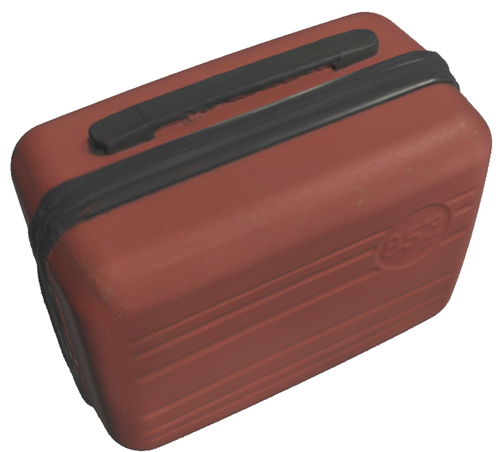}&
    \includegraphics[width=\imglenseven\textwidth]{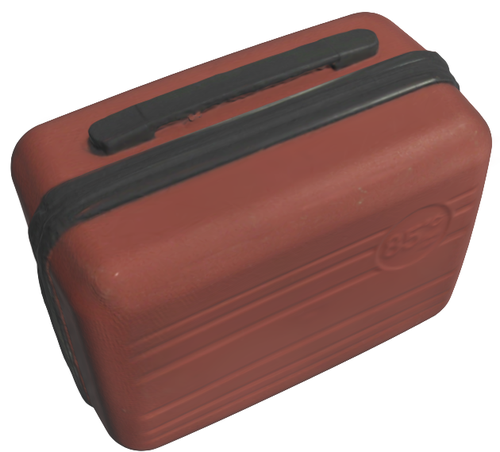}&
    \includegraphics[width=\imglenseven\textwidth]{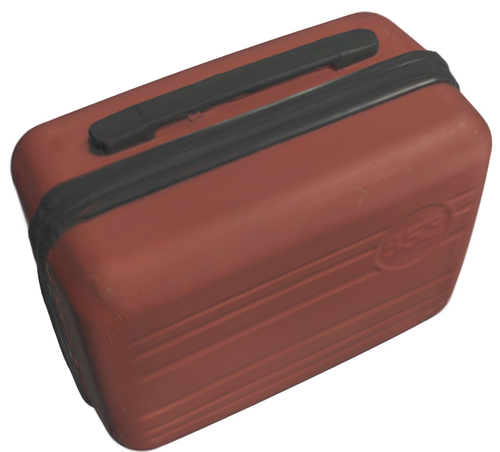}
    \\
    PSNR, FPS, $N_{\mathrm{GS}}$ & 32.79, 2.42, - & \textbf{33.53}, 9.21, - & 32.22, \textbf{219.98}, 112.58 & 33.42, 69.86, 98.20 & 32.23, 60.24, 95.57 & 31.44, 193.84, \textbf{33.11} \\
    \includegraphics[width=\imglenseven\textwidth]{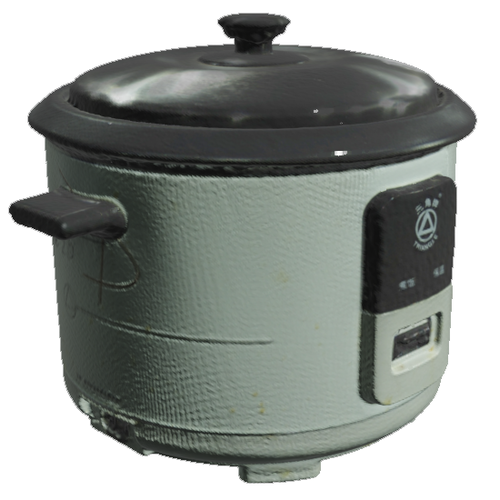}&
    \includegraphics[width=\imglenseven\textwidth]{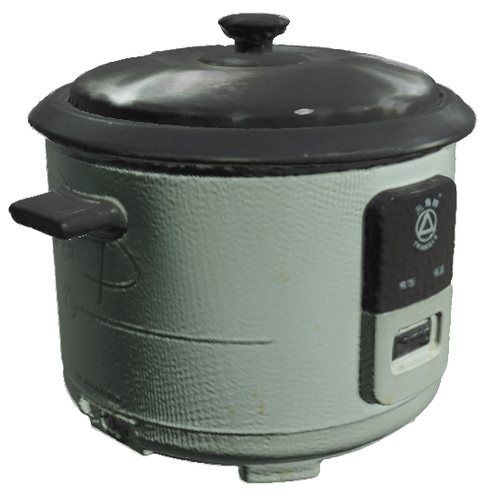}&
    \includegraphics[width=\imglenseven\textwidth]{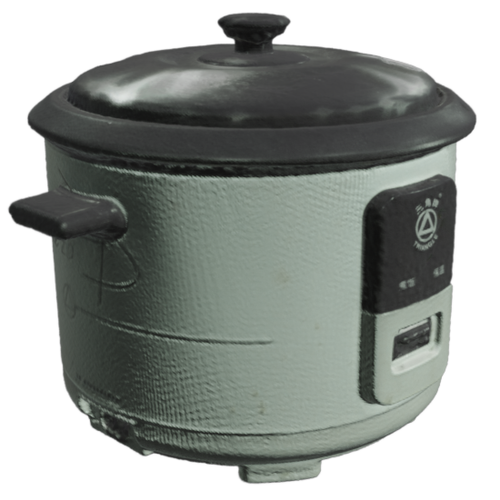}&
    \includegraphics[width=\imglenseven\textwidth]{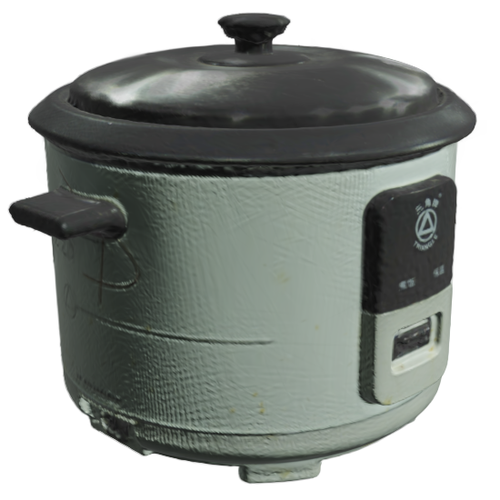}&
    \includegraphics[width=\imglenseven\textwidth]{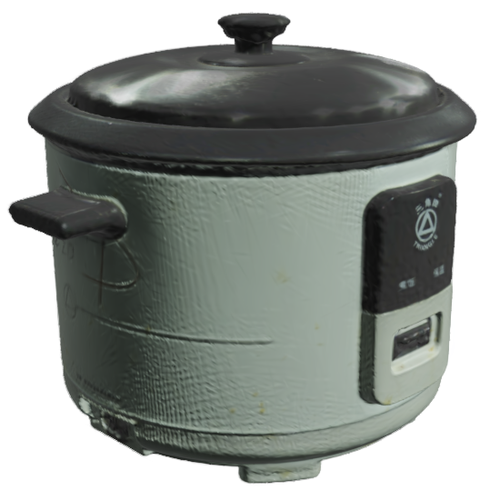}&
    \includegraphics[width=\imglenseven\textwidth]{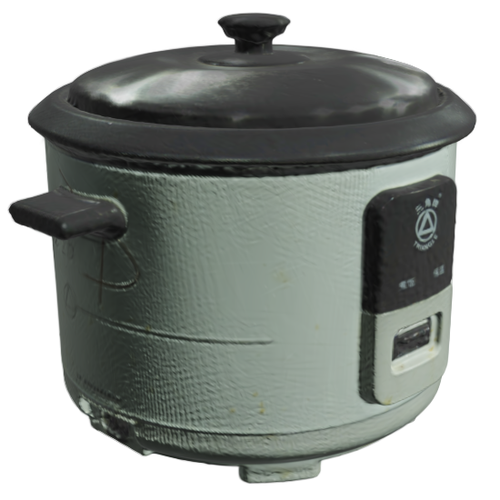}&
    \includegraphics[width=\imglenseven\textwidth]{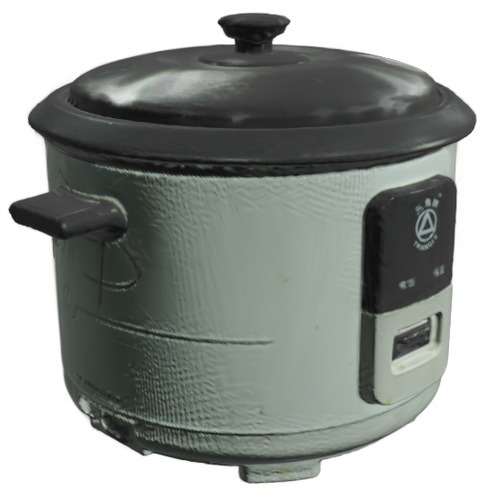}
    \\
    PSNR, FPS, $N_{\mathrm{GS}}$ & \textbf{35.09}, 1.91, - & 34.44, 8.94, - & 33.36, \textbf{197.33}, 109.39 & 34.88, 62.97, 60.14 & 33.58, 91.81, 80.29 & 33.25, 189.87, \textbf{59.84}\\
    \includegraphics[width=\imglenseven\textwidth]{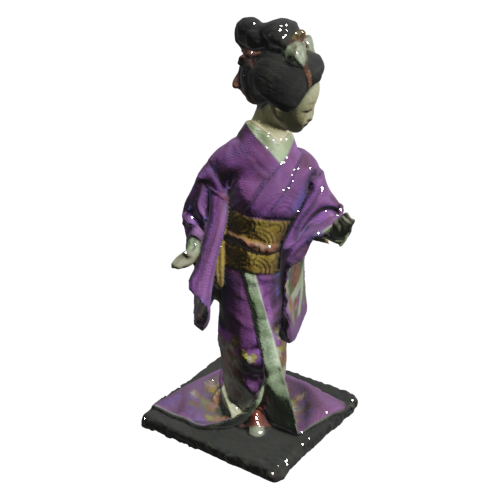}&
    \includegraphics[width=\imglenseven\textwidth]{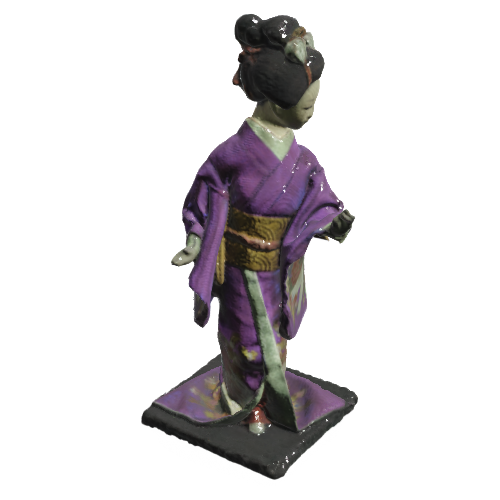}&
    \includegraphics[width=\imglenseven\textwidth]{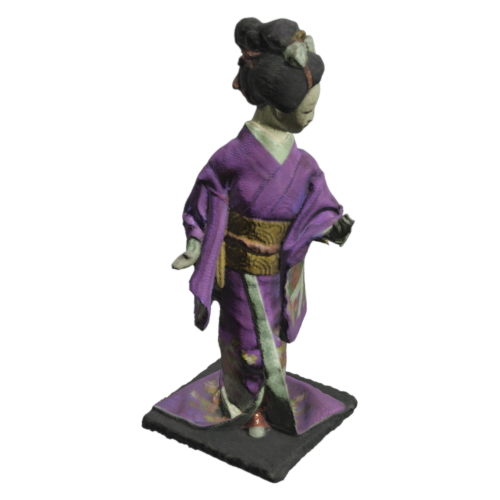}&
    \includegraphics[width=\imglenseven\textwidth]{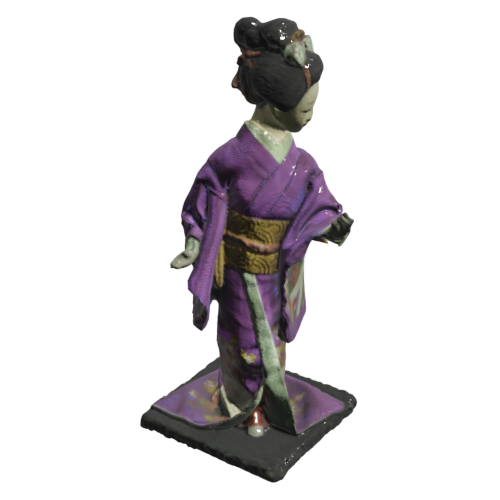}&
    \includegraphics[width=\imglenseven\textwidth]{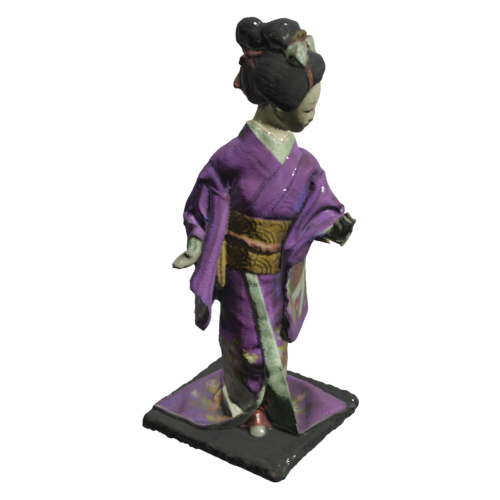}&
    \includegraphics[width=\imglenseven\textwidth]{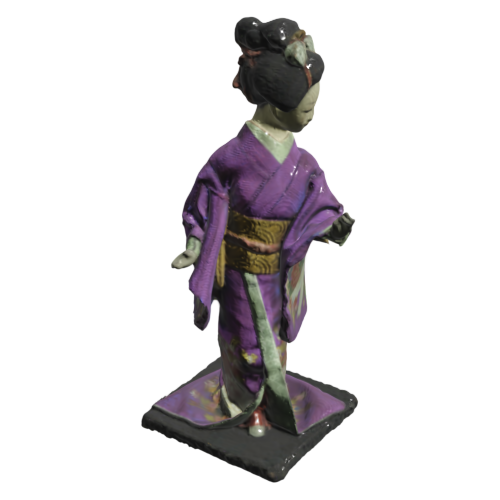}&
    \includegraphics[width=\imglenseven\textwidth]{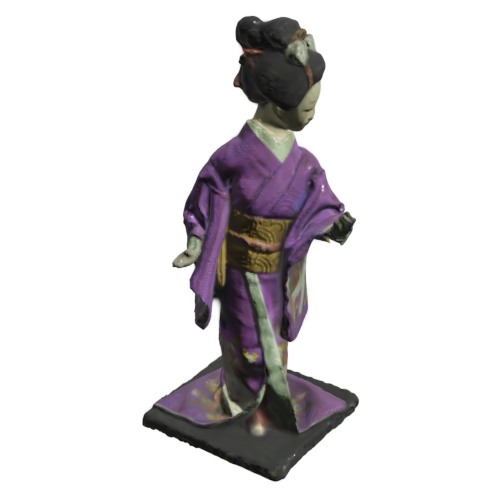}
    \\
    PSNR, FPS, $N_{\mathrm{GS}}$ & 37.24, 1.65, - & 36.67, 15.10, - & 37.43, \textbf{494.46}, 64.18 & \textbf{38.49}, 77.66, 66.56 & 37.41, 294.60, 45.73 & 36.79, 376.29, \textbf{38.29}\\
    GT image & Voxurf & NeuS2 & 3DGS & 2DGS & GOF & Ours\\
    \end{tabular}
    
    \end{scriptsize}
      \caption{Visual comparison of image rendering results on the OO3D-SL dataset and the NeRF-synthetic dataset. Our method achieves high rendering speed while maintaining comparable image rendering quality to other methods. Moreover, our method reduces the number of redundant Gaussians, making the model more compact. The 3-tuple below each figure represents the PSNR, FPS and the number of Gaussians (in thousands), with the best results highlighted in bold.}  \label{fig:synthesis}
\end{figure*}

\subsection{Setup}
\textbf{Implementation.} We realize our method before stage 4 by creating a specialized PyTorch CUDA extension library, which greatly improves the speed of reconstruction. We conducted our experiments on an NVIDIA A100 GPU with 40 GB memory and the octree construction can be completed in an hour. For the 3D Gaussian optimization, we optimize the model by 30k iterations in about 15 minutes.

\textbf{Dataset.}  We conducted the experiments on two datasets, the OmniObject3D~\cite{omniobject3d} dataset and the NeRF-Synthetic dataset. We selected 24 real objects (6 categories, each with 4 objects) that are affected by strong light and result in large areas of specular highlights on the surface, and we named it OO3D-SL dataset. For each object, there are 90 training images and 10 test images. Moreover, we also compare the results on the NeRF-Synthetic dataset~\cite{nerf}. This dataset contains a total of 8 objects. Each scene contains 100 images for training with a resolution of 800$\times$800 generated from different angles and 200 test images. For the evaluation metrics, we evaluated the Peak-to-Noise Ratio (PSNR), Frames Per Second (FPS) and Chamfer Distance (CD in $10^{-4}$). For some reconstructed models, due to occlusion or self-occlusion, there are parts of the 3D objects that remain unseen in any of the input images, such as the base of a rice cooker. We remove these parts before computing CD to measure geometric quality. In our experiments, we selected Voxurf~\cite{wu2022voxurf}, NeuS2~\cite{neus2}, 3DGS~\cite{3Dgaussians}, SuGaR~\cite{sugar}, GOF~\cite{gof} and 2DGS~\cite{2dgs} as the baseline methods for comparison.

\subsection{Results}
As shown in Tab.~\ref{tab:oo3d} and Fig.~\ref{fig:mesh_render}, the surface reconstructed by SuGaR~\cite{sugar} is discontinuous with holes, which is not good at reconstructing objects. Voxurf~\cite{wu2022voxurf} and GOF~\cite{gof} are sensitive to strong light and there are many holes in the surface. The quality of the geometric reconstruction from NeuS2~\cite{neus2} is close to ours, however, there are still a small number of objects affected by the strong light causing holes and wrong surface. 2DGS~\cite{2dgs} is almost unaffected by strong light, however, the object surface lacks detail and is poor for reconstruction of areas with fewer viewing directions. Moreover, it sometimes reconstructs the background that is not present in the images, and these redundancies are connected to the object which cannot be easily removed (see the examples of Chair and Lego in Fig.~\ref{fig:nerf}). Of all the comparison methods, our method obtains the highest geometric quality in the OO3D-SL dataset, and the reconstructed geometry is also comparable to these state-of-the-art methods in the non-strong light dataset such as NeRF-Synthetic (see Fig.~\ref{fig:nerf}). More visualization results of geometry can be seen in the appendix.

When it comes to novel view synthesis results, our methods also shows competitive quality compared with recent methods. We analyse the synthesis quality with PSNR and rendering FPS. As shown in Tab.~\ref{tab:oo3d} and Fig.~\ref{fig:synthesis}, our results achieve PSNR comparable to the SDF-based methods, such as Voxurf~\cite{wu2022voxurf} and NeuS2~\cite{neus2}. Due to the incompleteness of reconstructed meshes, SuGaR~\cite{sugar} performs slightly worse than other 3DGS-based methods, such as 2DGS~\cite{2dgs} and GOF~\cite{gof}. 
Although our PSNR performance is slightly lower than theirs, we keep the efficient rendering speed inherited from 3DGS~\cite{3Dgaussians}. Only our method reaches comparable FPS to 3DGS, as shown in Tab.~\ref{tab:oo3d}. Other 3DGS-based methods~\cite{sugar, 2dgs, gof} either have more computational costs or use their own rendering pipeline, and that reduces their rendering speed. For the non-3DGS-based methods~\cite{wu2022voxurf, neus2}, their rendering speed is slow because of the volume rendering pipeline from NeRF~\cite{nerf} and extra network inference time.
Moreover, for all 3DGS-based methods~\cite{3Dgaussians, sugar, 2dgs, gof}, we also report the number of 3D Gaussians in thousands, $N_{\mathrm{GS}}$, as shown in Fig.~\ref{fig:synthesis}. Our method uses a much smaller number of 3D Gaussians for most objects, especially in the OO3D-SL dataset, by effectively pruning redundant 3D Gaussians with our SDF supervision.

We present our qualitative comparison result of different methods in Tab.~\ref{tab:qualitative}. 
Our method is mainly developed for object-level targets. These methods~\cite{sugar, 2dgs, gof} can reconstruct satisfying geometries for scene-level targets, but they are not good at object-level reconstruction. Meanwhile, our method successfully reconstructs the object under strong lighting and preserves the details of the object. 
For rendering quality, our method slightly falls back to the recent 3DGS-based method but is still competitive with non-3DGS-based methods. 
Finally, only our method keeps a similar rendering speed to 3DGS~\cite{3Dgaussians}, thanks to the efficient rasterization module from 3DGS. Other methods either have extra computational costs from the MLP network or use different rendering pipelines, thus being unable to efficiently render images. Also, our method is capable of real-time rendering (60+ FPS) with a much higher resolution, e.g., 2048$\times$2048.

\begin{table}[htbp] \centering
    \begin{minipage}{0.55\textwidth} %
          \centering
          \caption{Qualitative comparison of different methods.}
          \scalebox{0.76}{
            \begin{tabular}{c|cccc}
            \toprule
             & Main  & Geometry  & Rendering  & Rendering  \\
            &  Target &  Quality &  Quality &  Speed \\
        
            \midrule
            Voxurf~\cite{wu2022voxurf}  & Object    & Middle    & High      & Low   \\
            NeuS2~\cite{neus2}          & Object    & High      & High    & Low   \\
            3DGS~\cite{3Dgaussians}     & Scene     & Low       & High      & High  \\
            SuGaR~\cite{sugar}          & Scene     & Middle    & Middle    & Middle\\
            2DGS~\cite{2dgs}            & Scene     & Middle    & High      & Middle\\
            GOF~\cite{gof}              & Scene     & Middle    & High      & Middle\\
            \midrule
            Ours                        & Object    & High      & Middle    & High  \\
            \bottomrule
          \end{tabular}}
          \label{tab:qualitative}
    \end{minipage}\hfill
    \begin{minipage}{0.4\textwidth}
       \centering
  \caption{Ablation on the loss terms.}
  \footnotesize
  \begin{tabular}{c|cc}
    \toprule
    Method & PSNR & $N_{\mathrm{GS}}$\\
    \midrule
    3DGS & 36.28 & +0.00\% \\
    w/ $\mathcal{L}_\text{scale}$ \& w/o $\mathcal{L}_\text{op}$ & 36.28 & +0.79\% \\
    w/ $\mathcal{L}_\text{op}$ \& w/o $\mathcal{L}_\text{scale}$ & 35.15 & -50.17\% \\
    w/ $\mathcal{L}_\text{op}$ \& w/ $\mathcal{L}_\text{scale}$ (Ours) & 35.18 & -52.33\% \\
    \bottomrule
  \end{tabular}
  \label{tab:ablation_weight}
    \end{minipage}
\end{table}

\subsection{Ablation Studies}
We conduct ablation studies on the opacity loss and the scale loss mentioned in Sec.~\ref{sec:SDF-GS}. We evaluate the performance of the baseline 3DGS~\cite{3Dgaussians} and our method, with opacity loss and scale loss activated or not. We conduct the experiment on OO3D-SL~\cite{omniobject3d} dataset and report PSNR and number of 3D Gaussians $N_{\mathrm{GS}}$. As shown in Tab.~\ref{tab:ablation_weight}, the opacity loss is an essential condition for 3D Gaussian pruning. It significantly reduces $N_{\mathrm{GS}}$, by more than 50\%. The scale loss does not contribute when locations of 3D Gaussians are not near the surface. In other words, it only works when opacity loss is activated and the 3D Gaussians are pulled towards the surface. 

\section{Conclusion}
In this study, we optimize an octree-based Gaussian splatting representation for reconstructing object-level implicit surface and radiation fields under strong lighting. The method confirms the possibility that Gaussians can guide geometric optimization, and good geometry can further optimize Gaussian points. 
Our method is robust to strong lighting and can reconstruct detailed object-level geometries, while retaining the advantages of real-time high-quality rendering of Gaussian splatting. %

\bibliographystyle{unsrtnat}
\bibliography{reference}  %

\begin{figure*}[h] 
    \centering
    \setlength\tabcolsep{1pt}
    \begin{footnotesize}
    \begin{tabular}{cccccccc}
    \includegraphics[width=\imgleneight\textwidth]{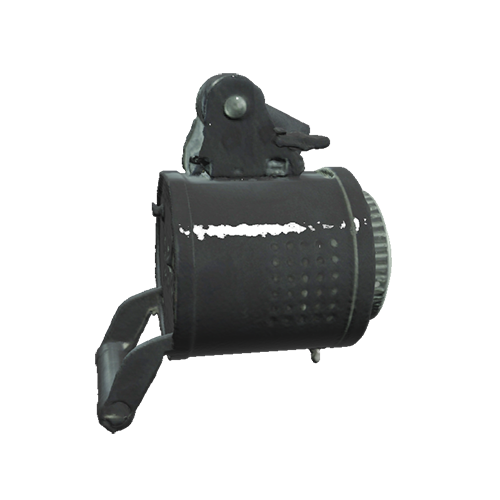}&
    \includegraphics[width=\imgleneight\textwidth]{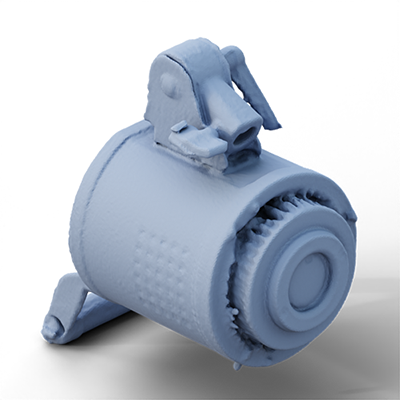}&
    \includegraphics[width=\imgleneight\textwidth]{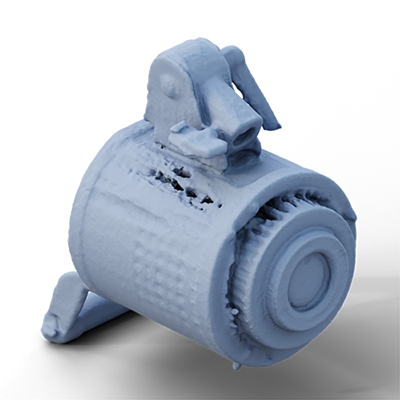}&
    \includegraphics[width=\imgleneight\textwidth]{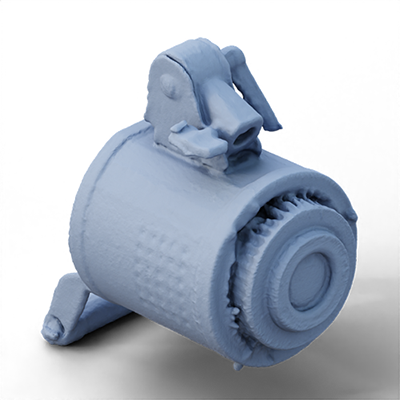}&
    \includegraphics[width=\imgleneight\textwidth]{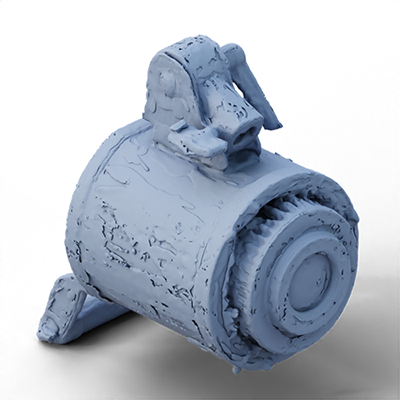}&
    \includegraphics[width=\imgleneight\textwidth]{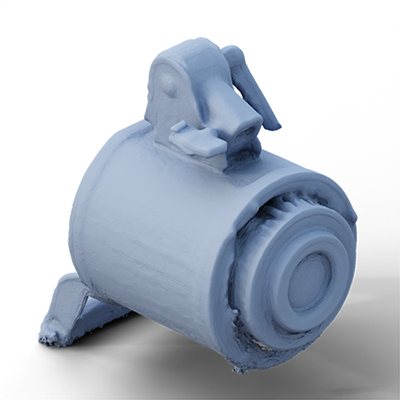}&
    \includegraphics[width=\imgleneight\textwidth]{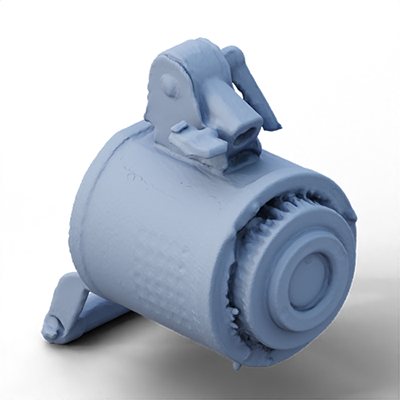}&
    \includegraphics[width=\imgleneight\textwidth]{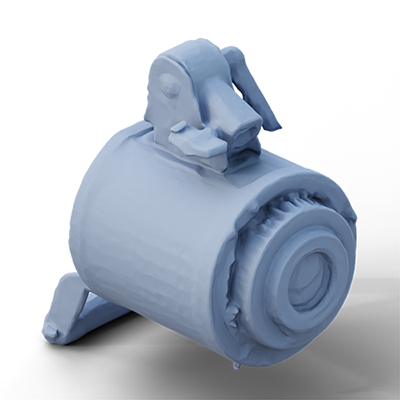}
    \\
    \includegraphics[width=\imgleneight\textwidth]{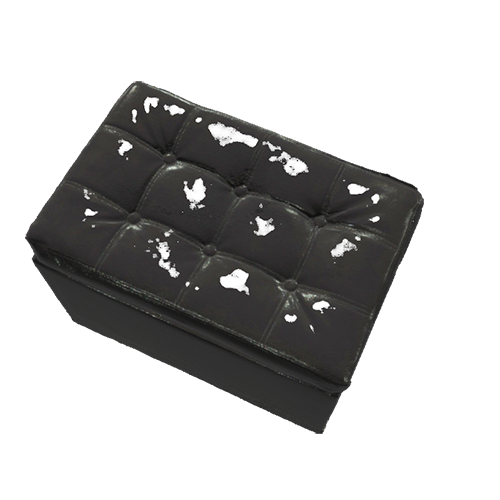}&
    \includegraphics[width=\imgleneight\textwidth]{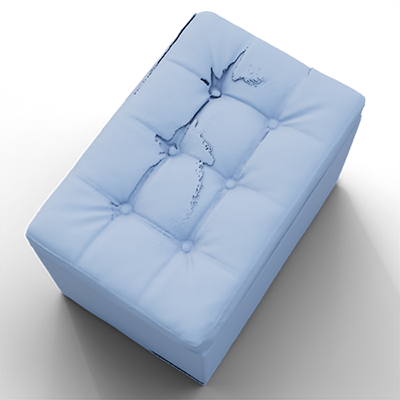}&
    \includegraphics[width=\imgleneight\textwidth]{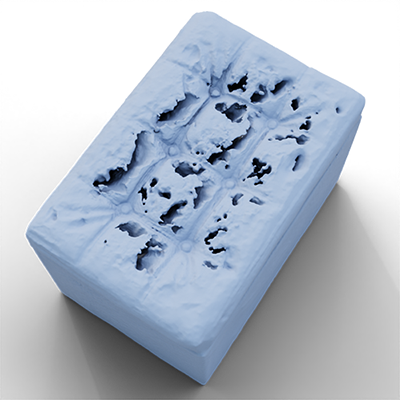}&
    \includegraphics[width=\imgleneight\textwidth]{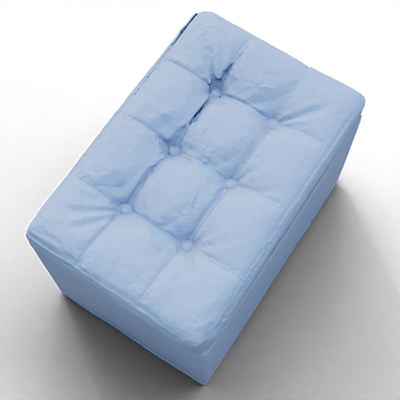}&
    \includegraphics[width=\imgleneight\textwidth]{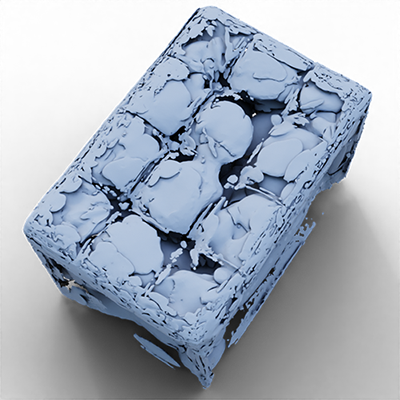}&
    \includegraphics[width=\imgleneight\textwidth]{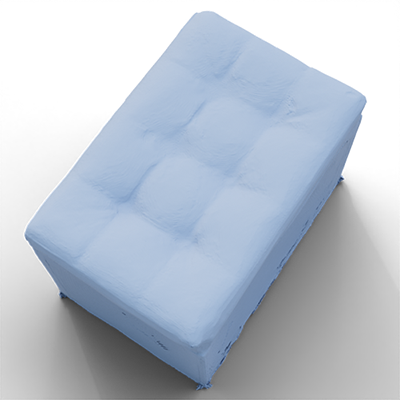}&
    \includegraphics[width=\imgleneight\textwidth]{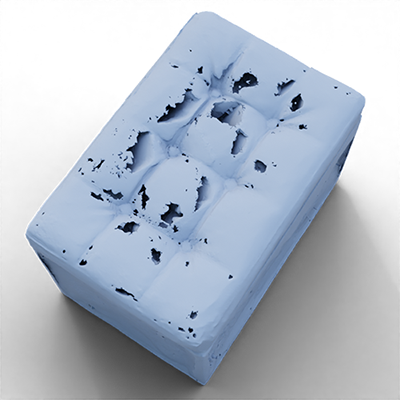}&
    \includegraphics[width=\imgleneight\textwidth]{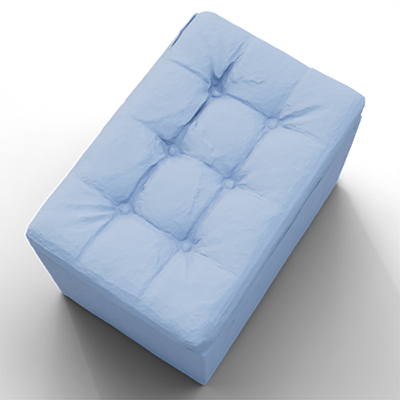}
    \\
    \includegraphics[width=\imgleneight\textwidth]{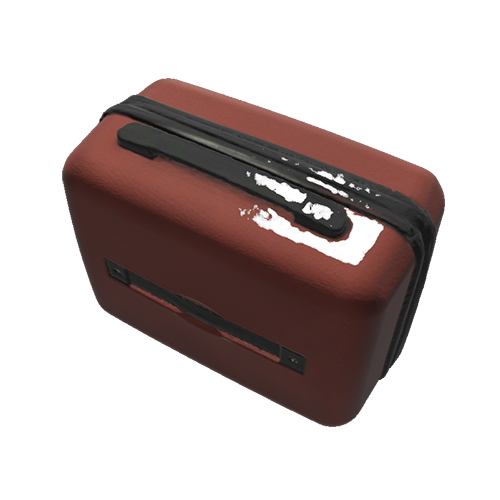}&
    \includegraphics[width=\imgleneight\textwidth]{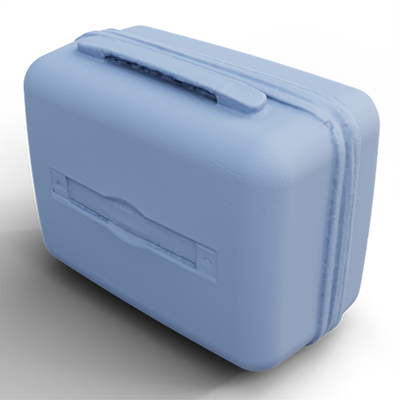}&
    \includegraphics[width=\imgleneight\textwidth]{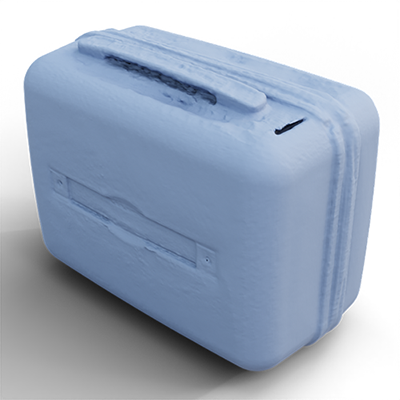}&
    \includegraphics[width=\imgleneight\textwidth]{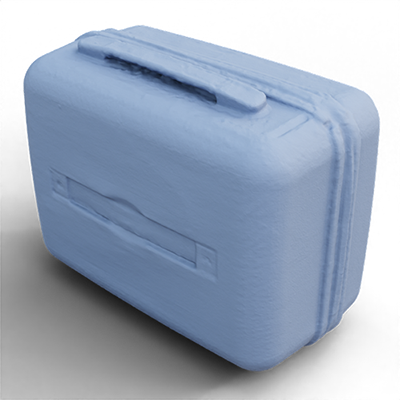}&
    \includegraphics[width=\imgleneight\textwidth]{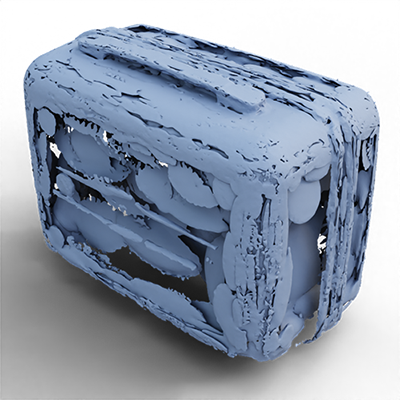}&
    \includegraphics[width=\imgleneight\textwidth]{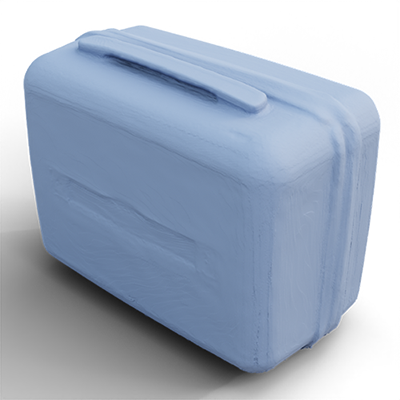}&
    \includegraphics[width=\imgleneight\textwidth]{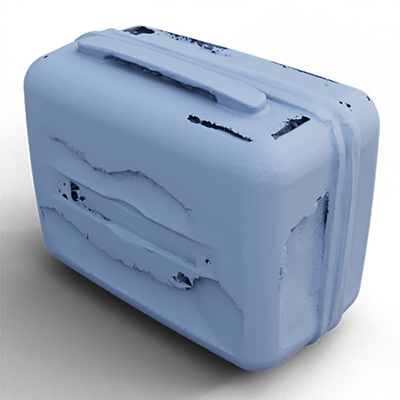}&
    \includegraphics[width=\imgleneight\textwidth]{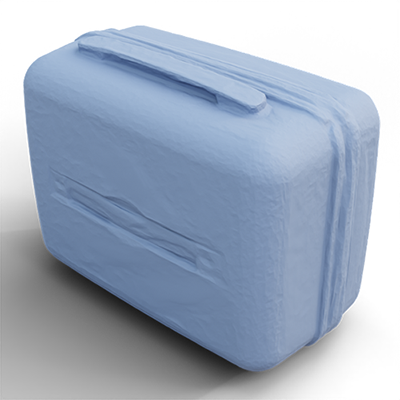}
    \\
    \includegraphics[width=\imgleneight\textwidth]{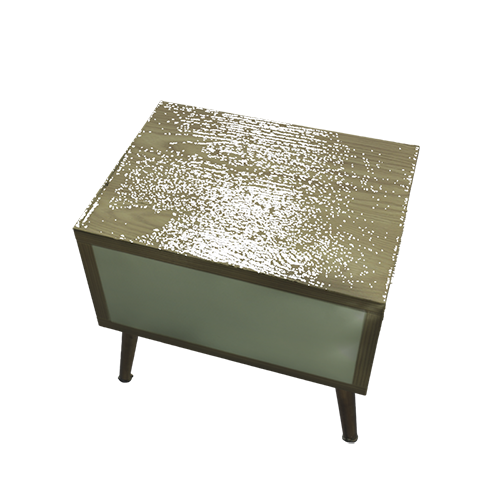}&
    \includegraphics[width=\imgleneight\textwidth]{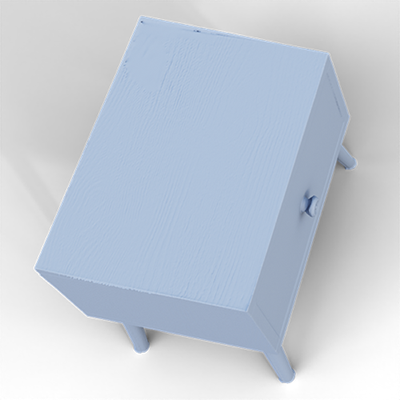}&
    \includegraphics[width=\imgleneight\textwidth]{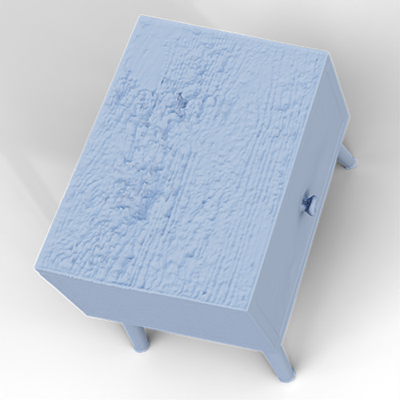}&
    \includegraphics[width=\imgleneight\textwidth]{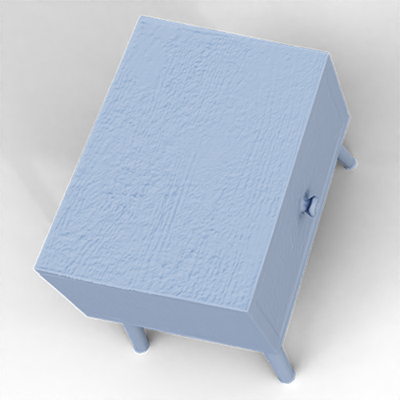}&
    \includegraphics[width=\imgleneight\textwidth]{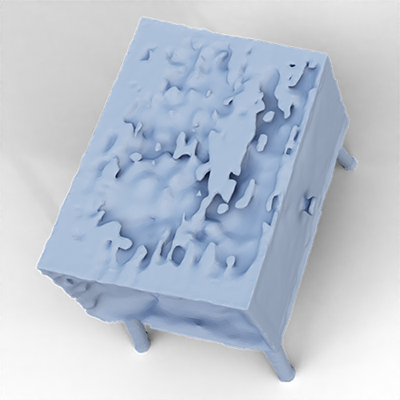}&
    \includegraphics[width=\imgleneight\textwidth]{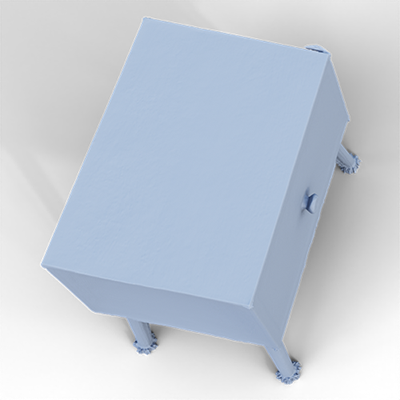}&
    \includegraphics[width=\imgleneight\textwidth]{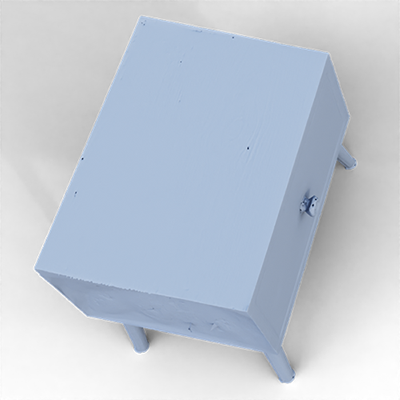}&
    \includegraphics[width=\imgleneight\textwidth]{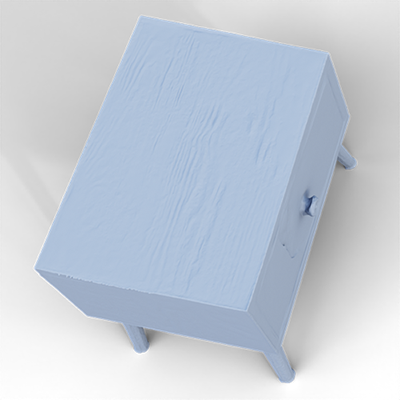}
    \\
    \includegraphics[width=\imgleneight\textwidth]{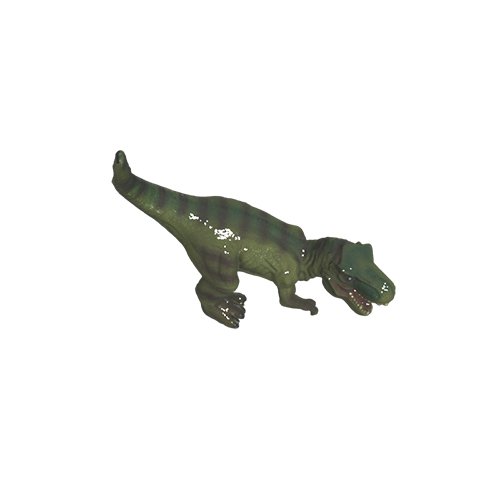}&
    \includegraphics[width=\imgleneight\textwidth]{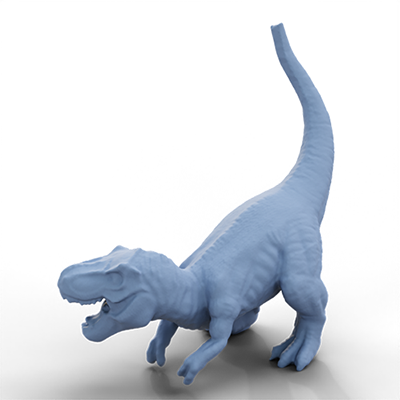}&
    \includegraphics[width=\imgleneight\textwidth]{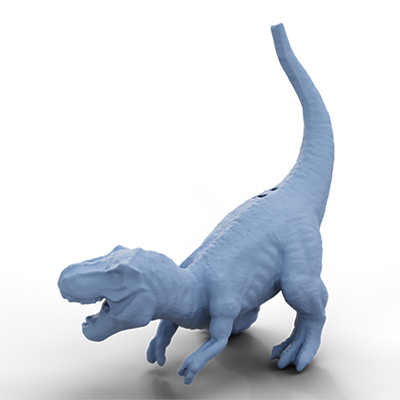}&
    \includegraphics[width=\imgleneight\textwidth]{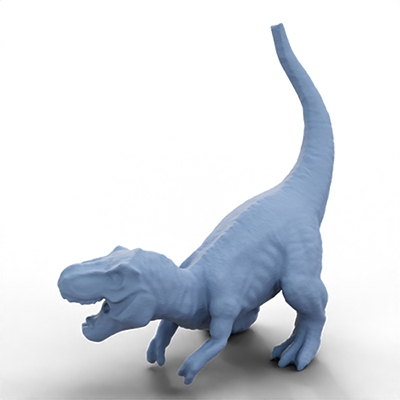}&
    \includegraphics[width=\imgleneight\textwidth]{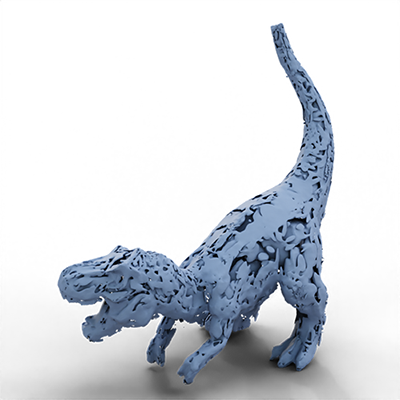}&
    \includegraphics[width=\imgleneight\textwidth]{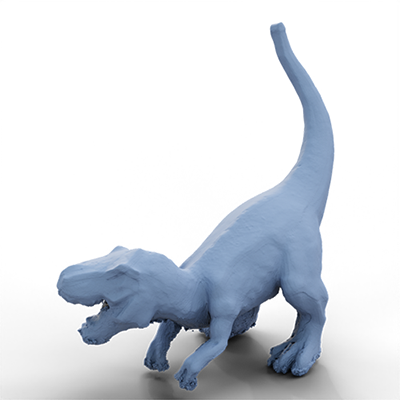}&
    \includegraphics[width=\imgleneight\textwidth]{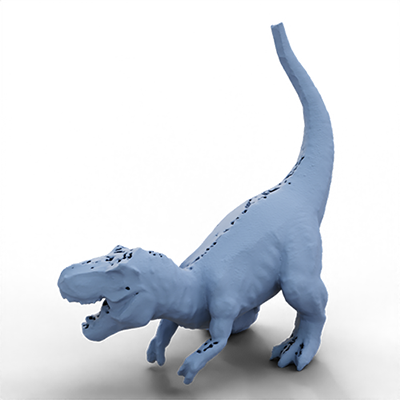}&
    \includegraphics[width=\imgleneight\textwidth]{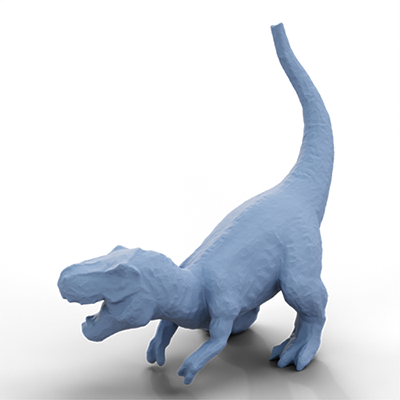}
    \\
    GT image & GT mesh & Voxurf & NeuS2 & SuGaR & 2DGS & GOF & Ours\\
    \end{tabular}
    
    \end{footnotesize}
      \caption{More comparison of geometric reconstruction results on the OO3D-SL dataset.} \label{fig:OO3D-more-3}
\end{figure*}
\begin{figure*}[t] 
    \centering
    \setlength\tabcolsep{1pt}
    \begin{footnotesize}
    \scalebox{0.97}{
    \begin{tabular}{cccccccc}
    \includegraphics[width=\imgleneight\textwidth]{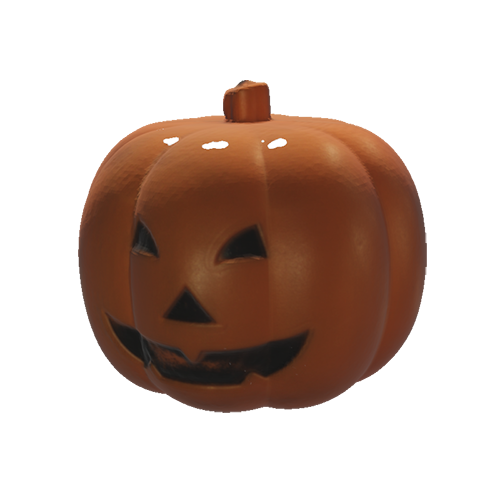}&
    \includegraphics[width=\imgleneight\textwidth]{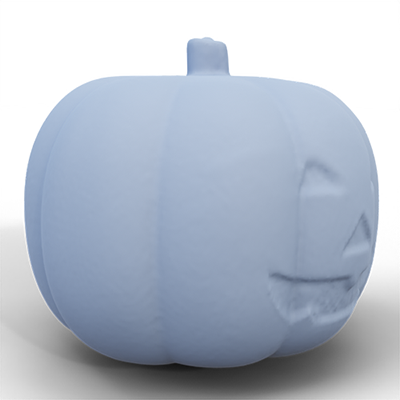}&
    \includegraphics[width=\imgleneight\textwidth]{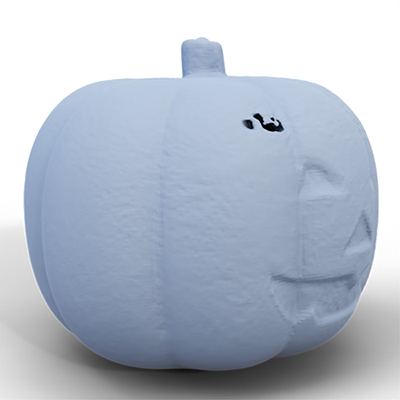}&
    \includegraphics[width=\imgleneight\textwidth]{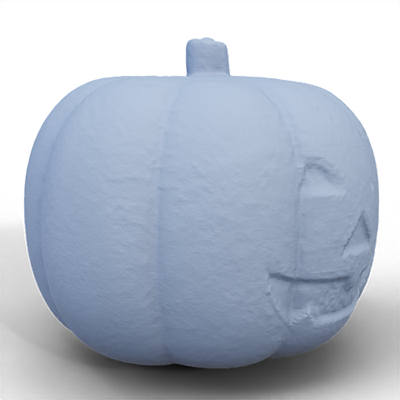}&
    \includegraphics[width=\imgleneight\textwidth]{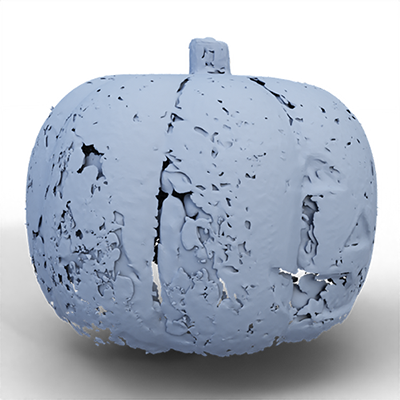}&
    \includegraphics[width=\imgleneight\textwidth]{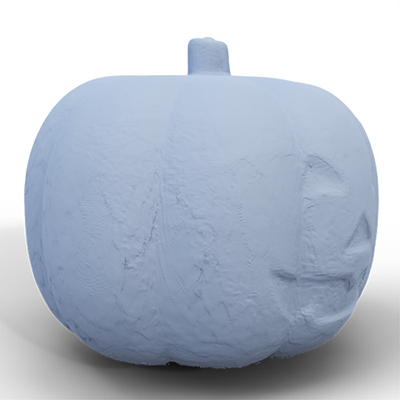}&
    \includegraphics[width=\imgleneight\textwidth]{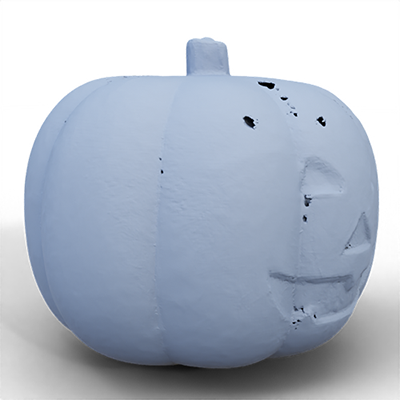}&
    \includegraphics[width=\imgleneight\textwidth]{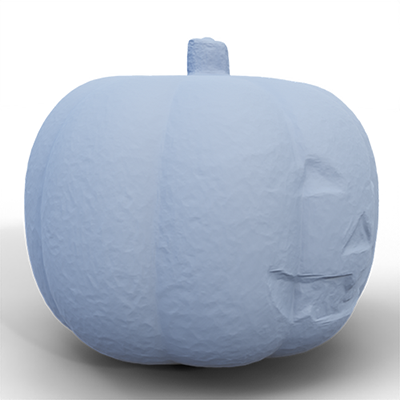}
    \\
    \includegraphics[width=\imgleneight\textwidth]{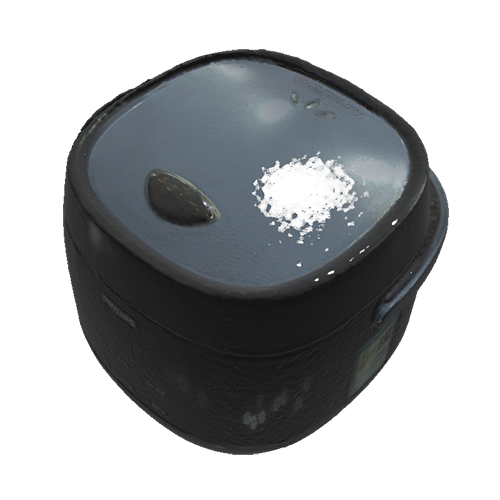}&
    \includegraphics[width=\imgleneight\textwidth]{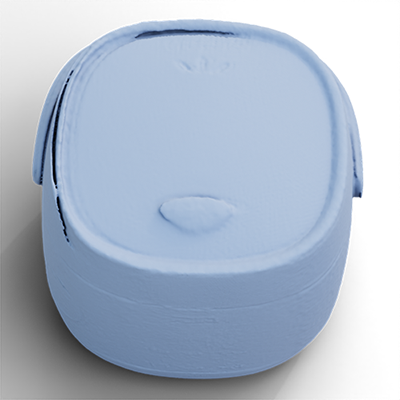}&
    \includegraphics[width=\imgleneight\textwidth]{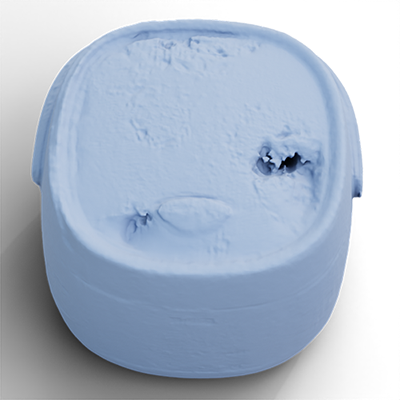}&
    \includegraphics[width=\imgleneight\textwidth]{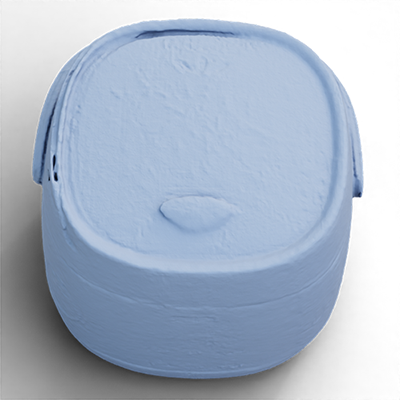}&
    \includegraphics[width=\imgleneight\textwidth]{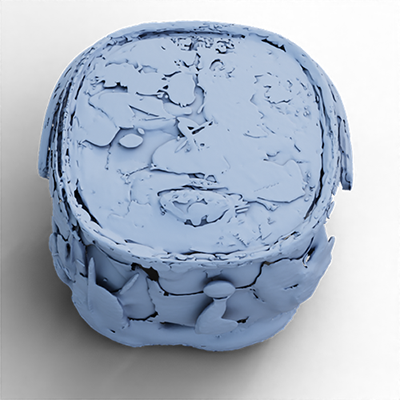}&
    \includegraphics[width=\imgleneight\textwidth]{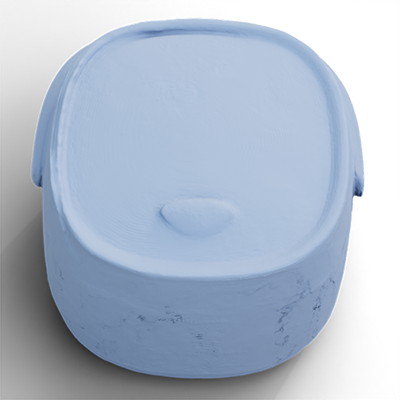}&
    \includegraphics[width=\imgleneight\textwidth]{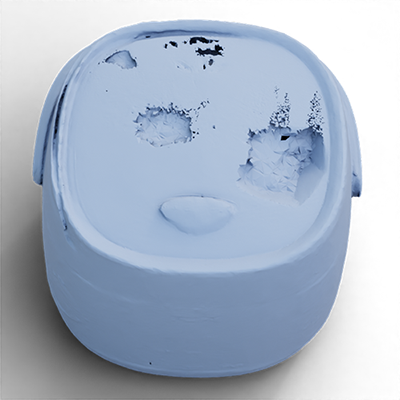}&
    \includegraphics[width=\imgleneight\textwidth]{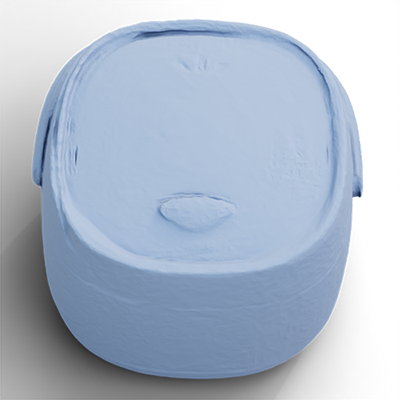}
    \\
    \includegraphics[width=\imgleneight\textwidth]{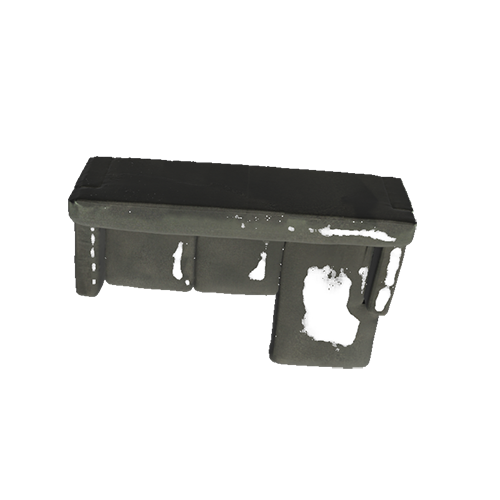}&
    \includegraphics[width=\imgleneight\textwidth]{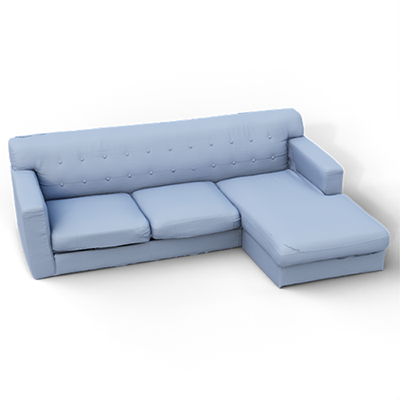}&
    \includegraphics[width=\imgleneight\textwidth]{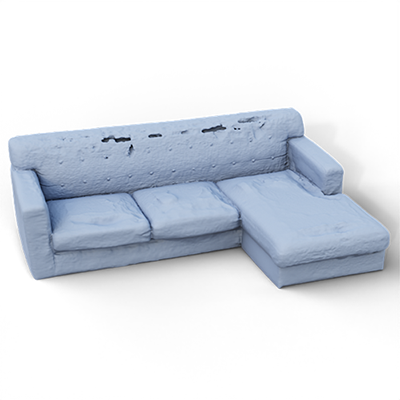}&
    \includegraphics[width=\imgleneight\textwidth]{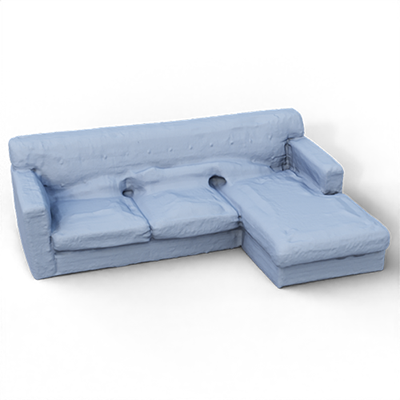}&
    \includegraphics[width=\imgleneight\textwidth]{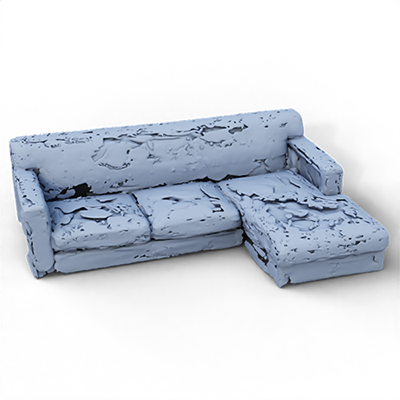}&
    \includegraphics[width=\imgleneight\textwidth]{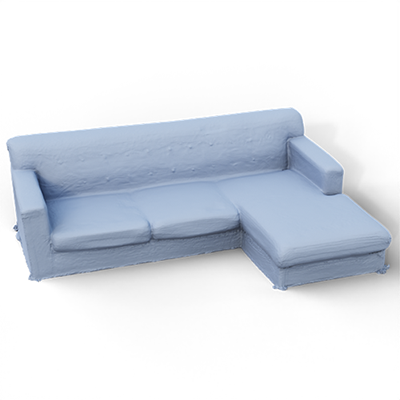}&
    \includegraphics[width=\imgleneight\textwidth]{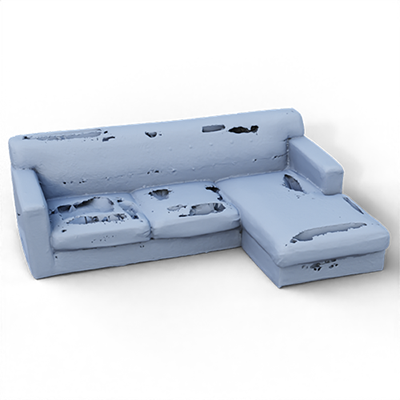}&
    \includegraphics[width=\imgleneight\textwidth]{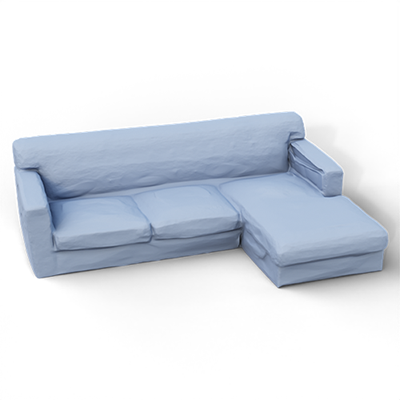}
    \\
    \includegraphics[width=\imgleneight\textwidth]{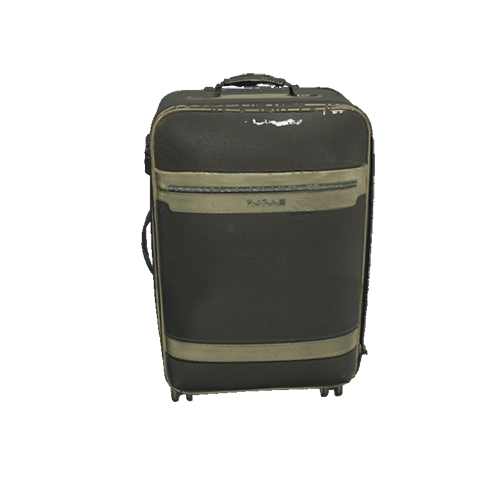}&
    \includegraphics[width=\imgleneight\textwidth]{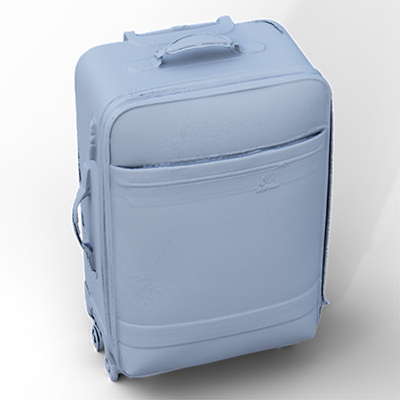}&
    \includegraphics[width=\imgleneight\textwidth]{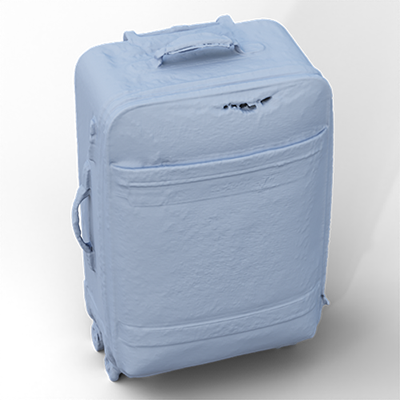}&
    \includegraphics[width=\imgleneight\textwidth]{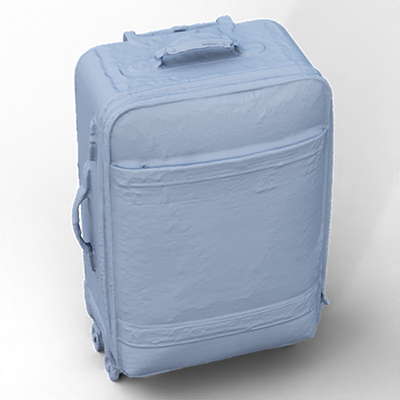}&
    \includegraphics[width=\imgleneight\textwidth]{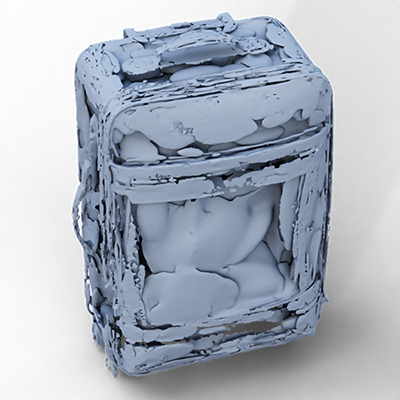}&
    \includegraphics[width=\imgleneight\textwidth]{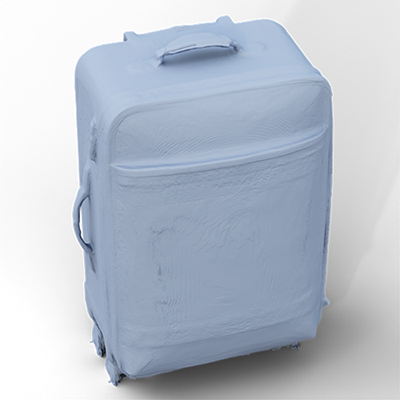}&
    \includegraphics[width=\imgleneight\textwidth]{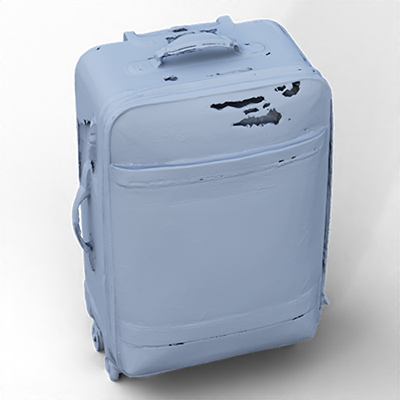}&
    \includegraphics[width=\imgleneight\textwidth]{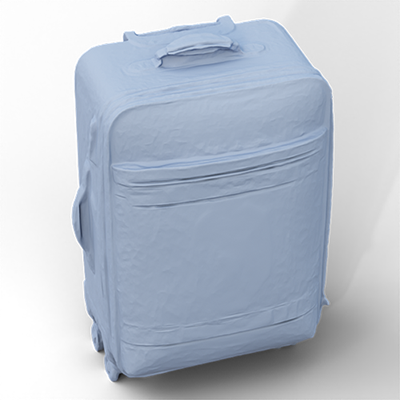}
    \\
    \includegraphics[width=\imgleneight\textwidth]{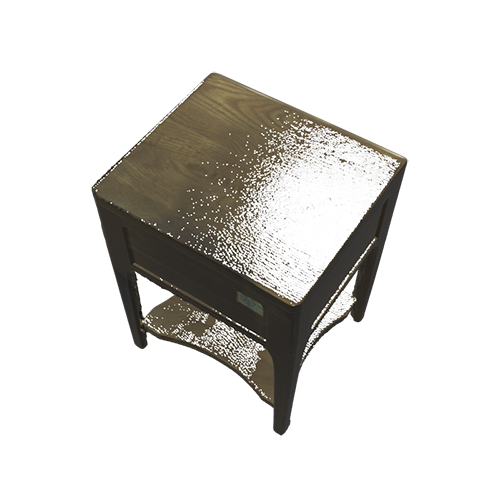}&
    \includegraphics[width=\imgleneight\textwidth]{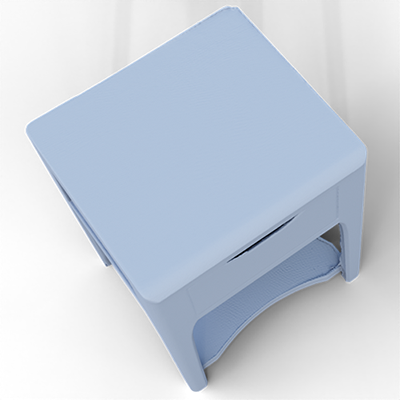}&
    \includegraphics[width=\imgleneight\textwidth]{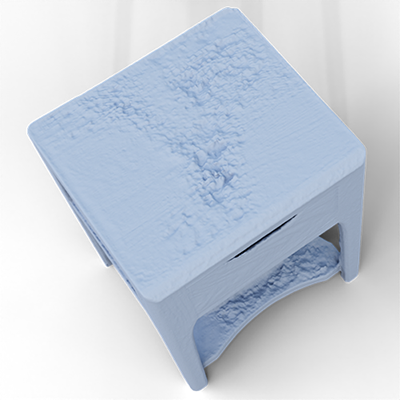}&
    \includegraphics[width=\imgleneight\textwidth]{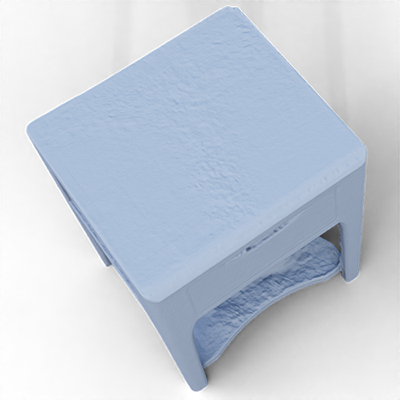}&
    \includegraphics[width=\imgleneight\textwidth]{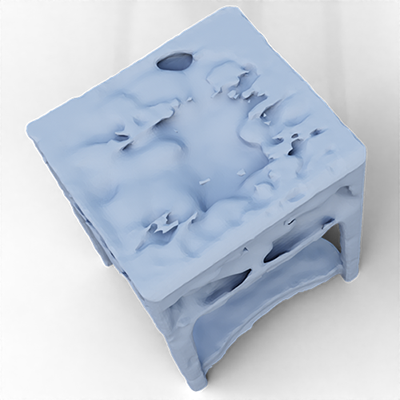}&
    \includegraphics[width=\imgleneight\textwidth]{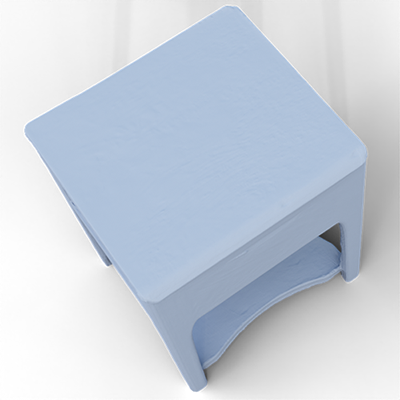}&
    \includegraphics[width=\imgleneight\textwidth]{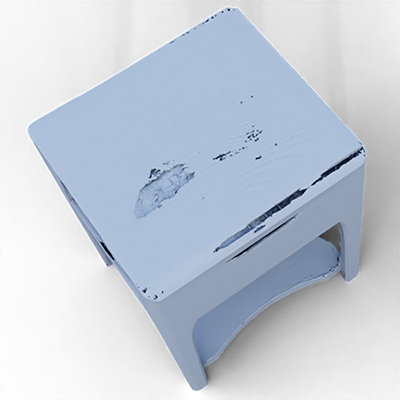}&
    \includegraphics[width=\imgleneight\textwidth]{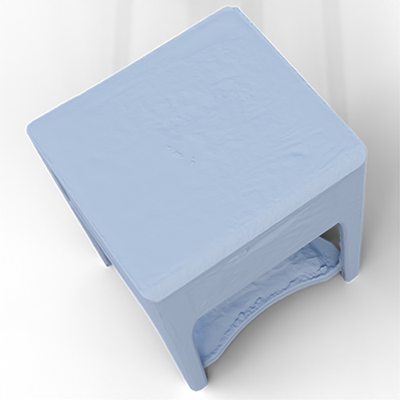}
    \\
    \includegraphics[width=\imgleneight\textwidth]{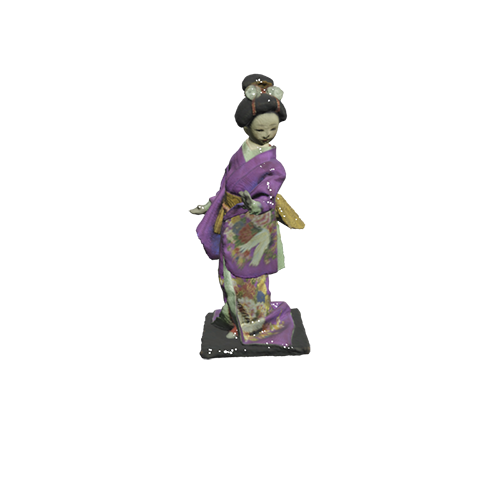}&
    \includegraphics[width=\imgleneight\textwidth]{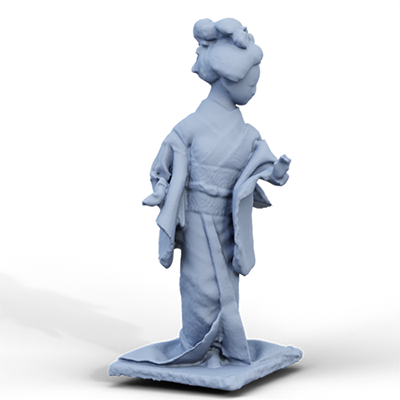}&
    \includegraphics[width=\imgleneight\textwidth]{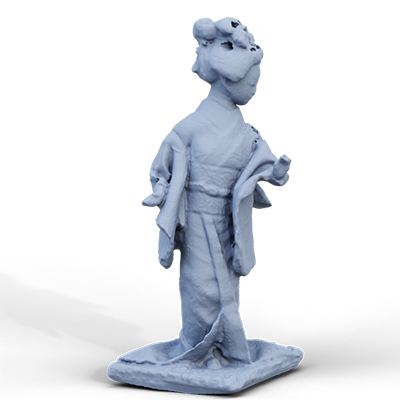}&
    \includegraphics[width=\imgleneight\textwidth]{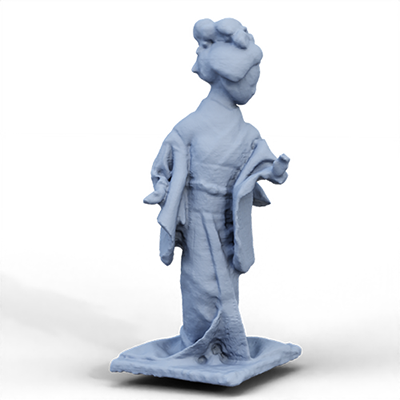}&
    \includegraphics[width=\imgleneight\textwidth]{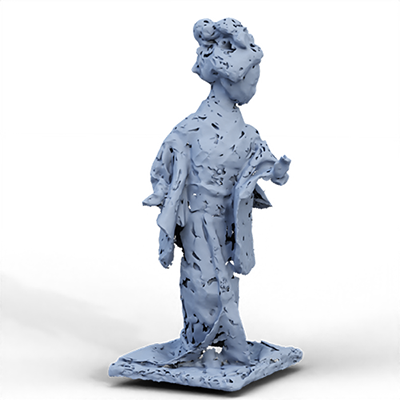}&
    \includegraphics[width=\imgleneight\textwidth]{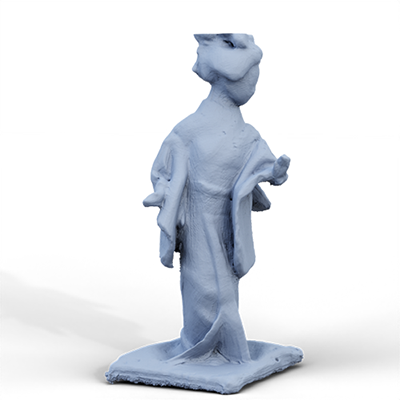}&
    \includegraphics[width=\imgleneight\textwidth]{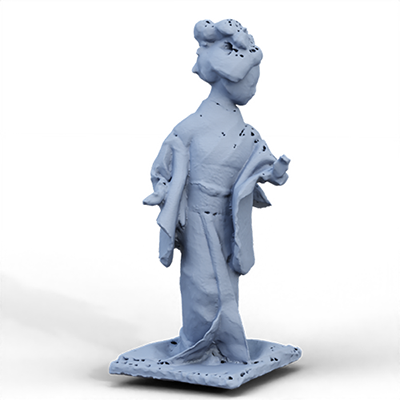}&
    \includegraphics[width=\imgleneight\textwidth]{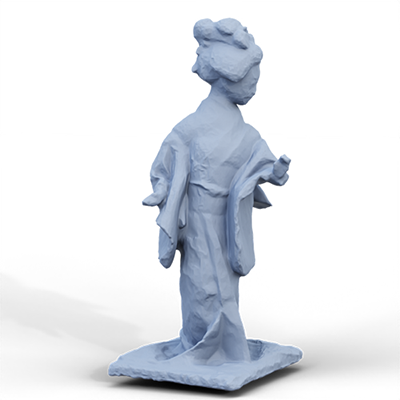}
    \\
    \includegraphics[width=\imgleneight\textwidth]{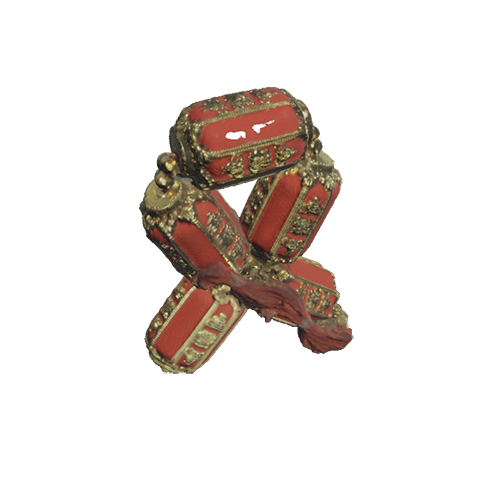}&
    \includegraphics[width=\imgleneight\textwidth]{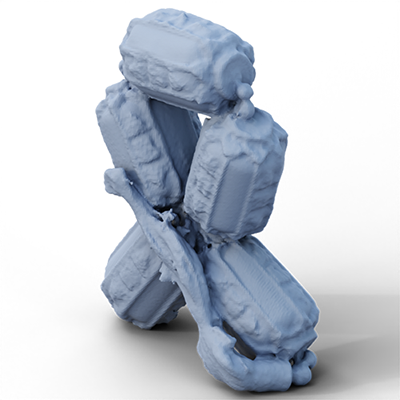}&
    \includegraphics[width=\imgleneight\textwidth]{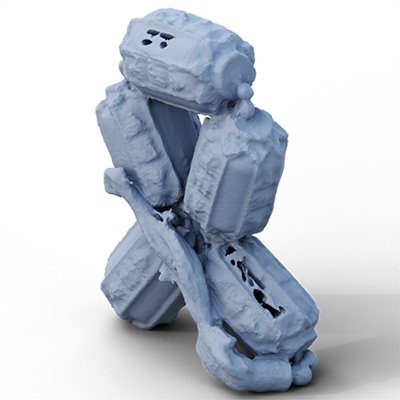}&
    \includegraphics[width=\imgleneight\textwidth]{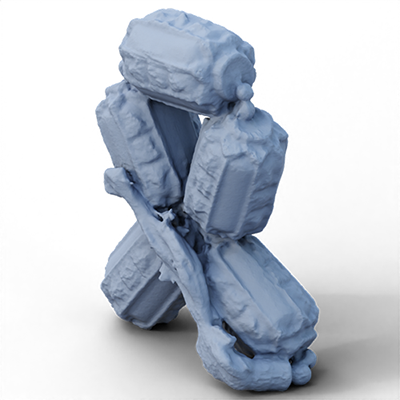}&
    \includegraphics[width=\imgleneight\textwidth]{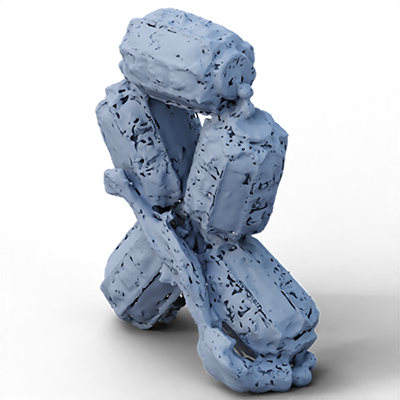}&
    \includegraphics[width=\imgleneight\textwidth]{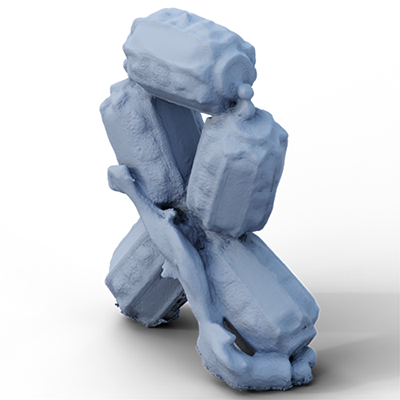}&
    \includegraphics[width=\imgleneight\textwidth]{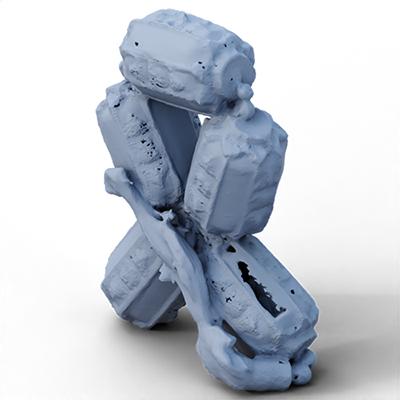}&
    \includegraphics[width=\imgleneight\textwidth]{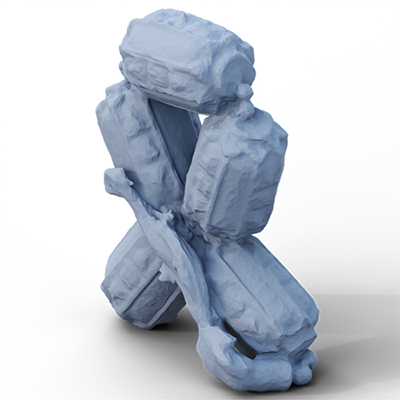}
    \\
    \includegraphics[width=\imgleneight\textwidth]{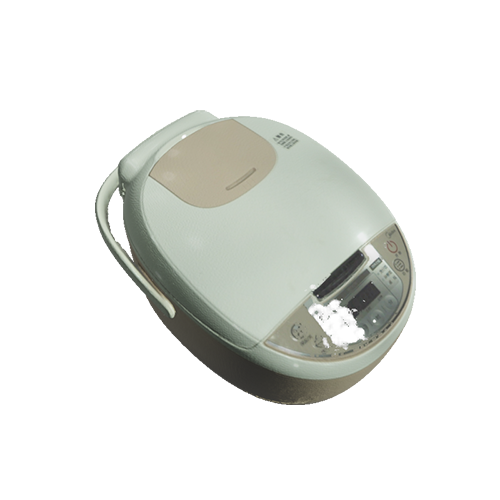}&
    \includegraphics[width=\imgleneight\textwidth]{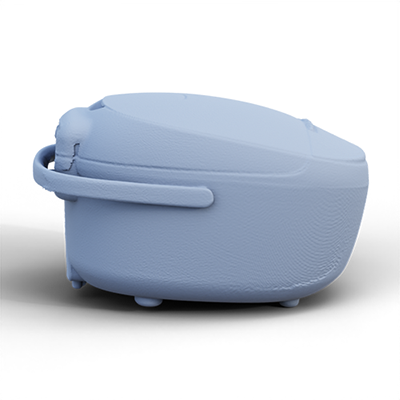}&
    \includegraphics[width=\imgleneight\textwidth]{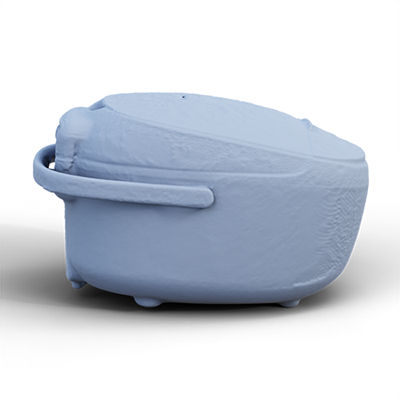}&
    \includegraphics[width=\imgleneight\textwidth]{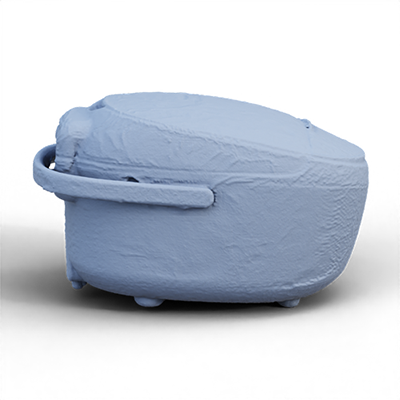}&
    \includegraphics[width=\imgleneight\textwidth]{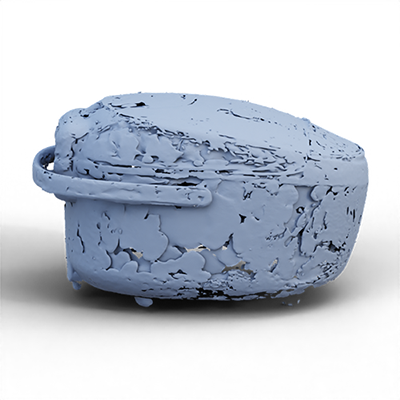}&
    \includegraphics[width=\imgleneight\textwidth]{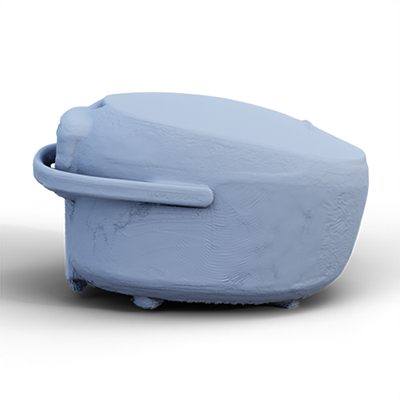}&
    \includegraphics[width=\imgleneight\textwidth]{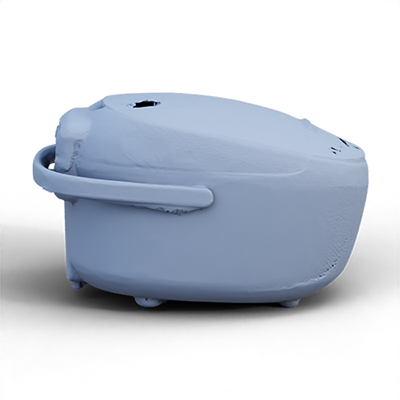}&
    \includegraphics[width=\imgleneight\textwidth]{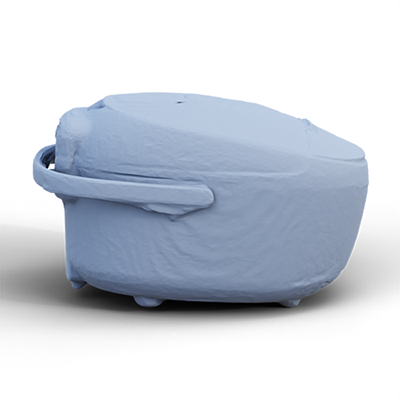}
    \\
    \includegraphics[width=\imgleneight\textwidth]{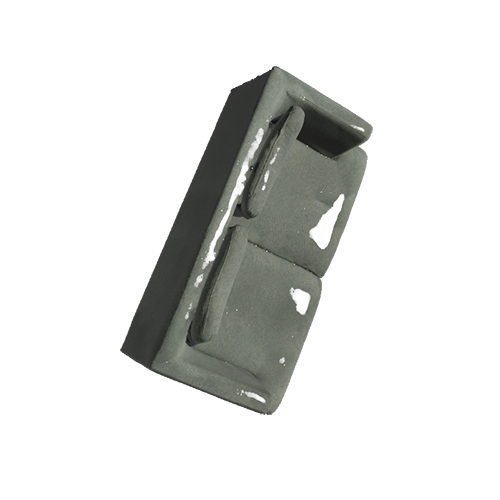}&
    \includegraphics[width=\imgleneight\textwidth]{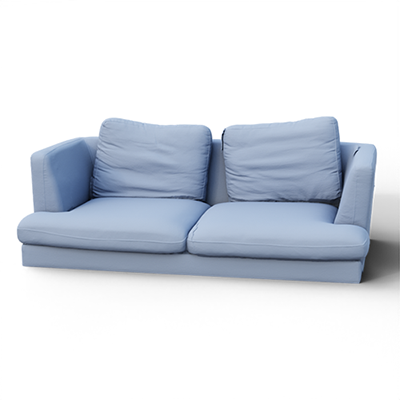}&
    \includegraphics[width=\imgleneight\textwidth]{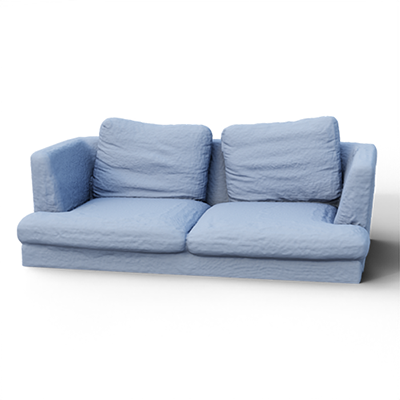}&
    \includegraphics[width=\imgleneight\textwidth]{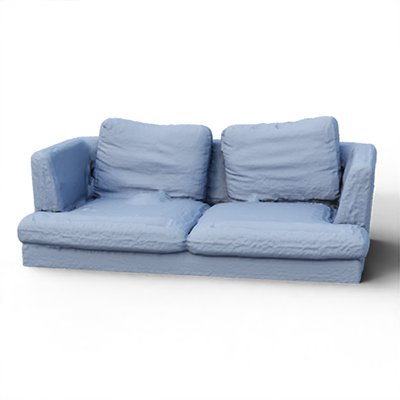}&
    \includegraphics[width=\imgleneight\textwidth]{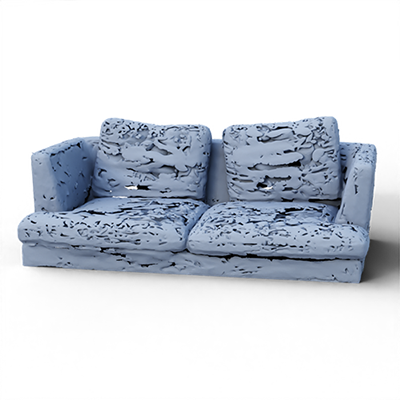}&
    \includegraphics[width=\imgleneight\textwidth]{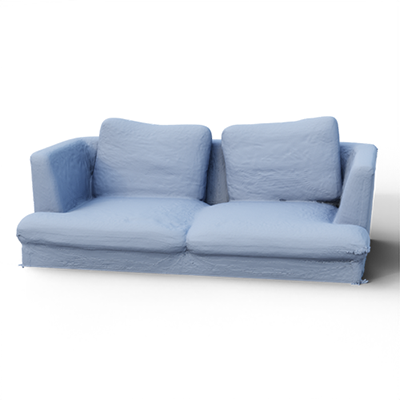}&
    \includegraphics[width=\imgleneight\textwidth]{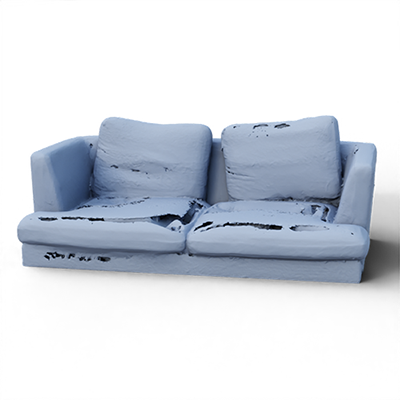}&
    \includegraphics[width=\imgleneight\textwidth]{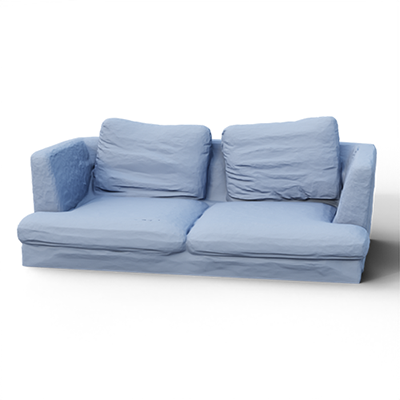}
    \\
     \includegraphics[width=\imgleneight\textwidth]{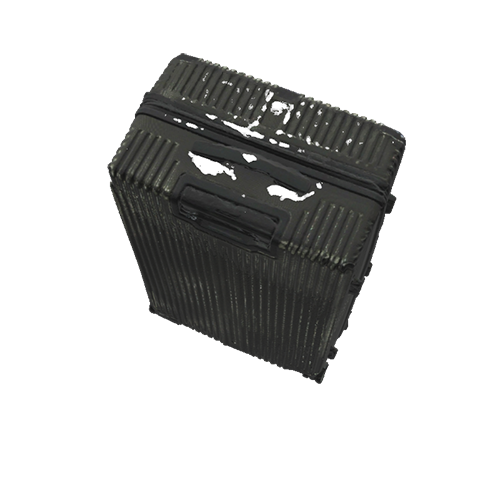}&
    \includegraphics[width=\imgleneight\textwidth]{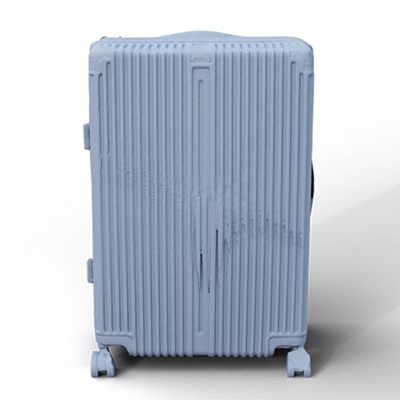}&
    \includegraphics[width=\imgleneight\textwidth]{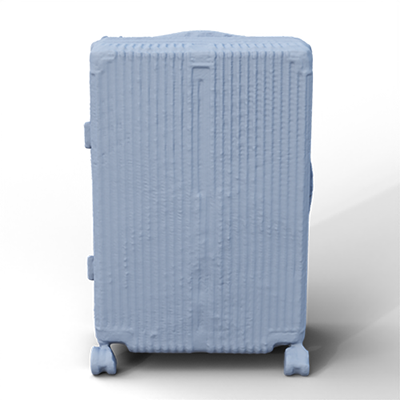}&
    \includegraphics[width=\imgleneight\textwidth]{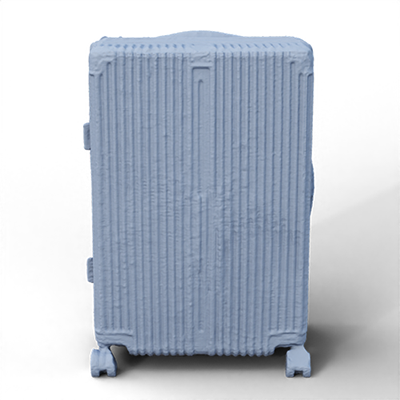}&
    \includegraphics[width=\imgleneight\textwidth]{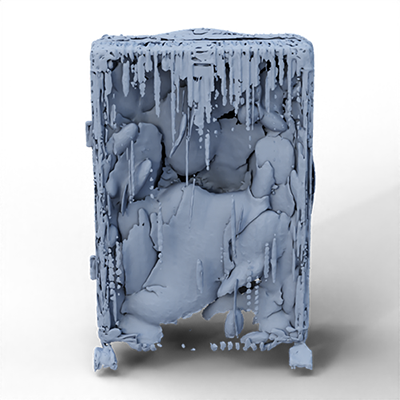}&
    \includegraphics[width=\imgleneight\textwidth]{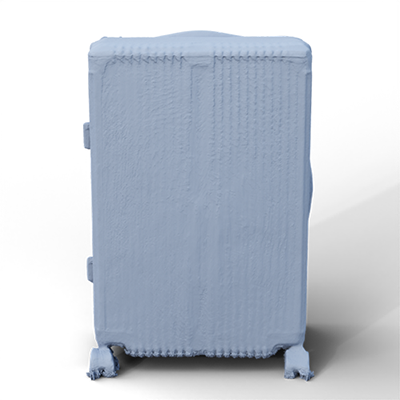}&
    \includegraphics[width=\imgleneight\textwidth]{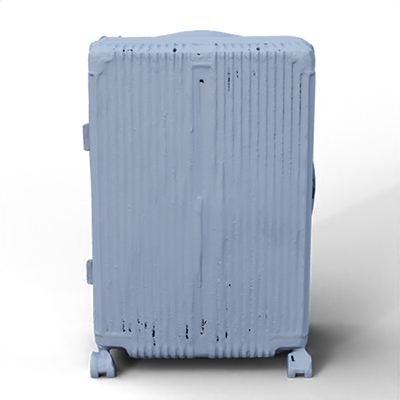}&
    \includegraphics[width=\imgleneight\textwidth]{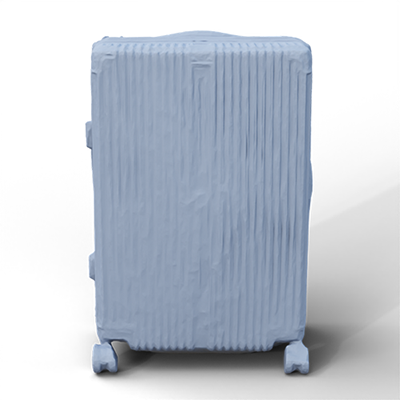}
    \\
    \includegraphics[width=\imgleneight\textwidth]{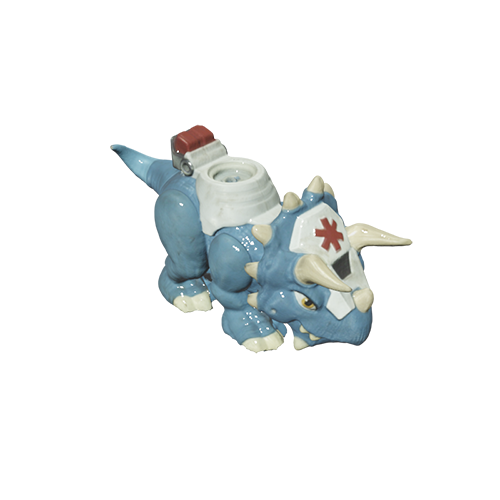}&
    \includegraphics[width=\imgleneight\textwidth]{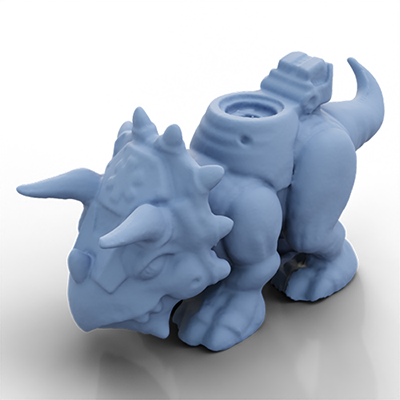}&
    \includegraphics[width=\imgleneight\textwidth]{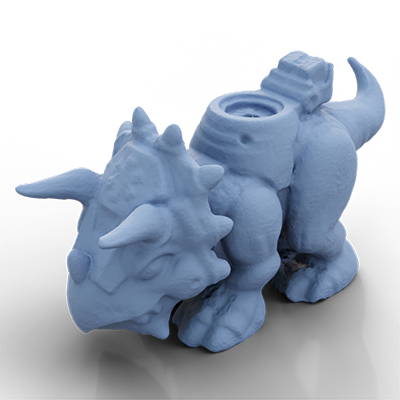}&
    \includegraphics[width=\imgleneight\textwidth]{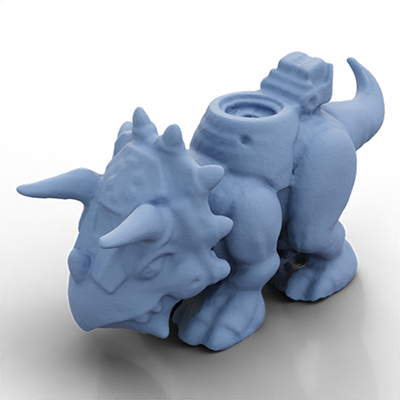}&
    \includegraphics[width=\imgleneight\textwidth]{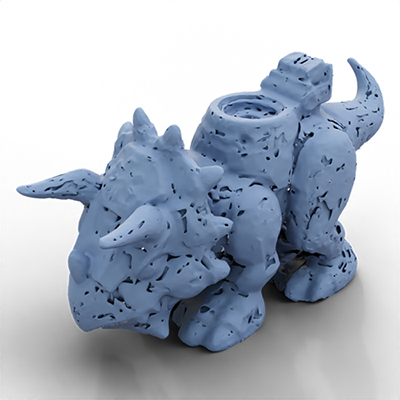}&
    \includegraphics[width=\imgleneight\textwidth]{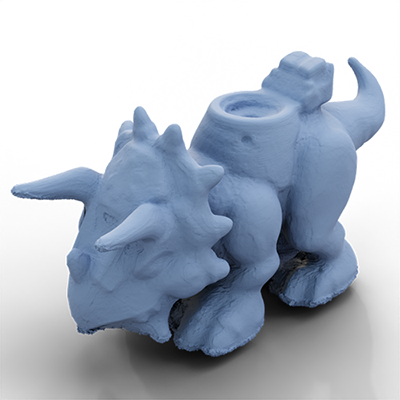}&
    \includegraphics[width=\imgleneight\textwidth]{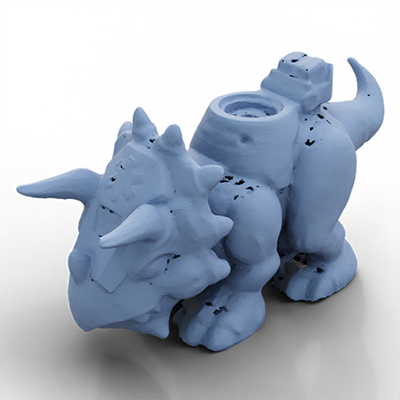}&
    \includegraphics[width=\imgleneight\textwidth]{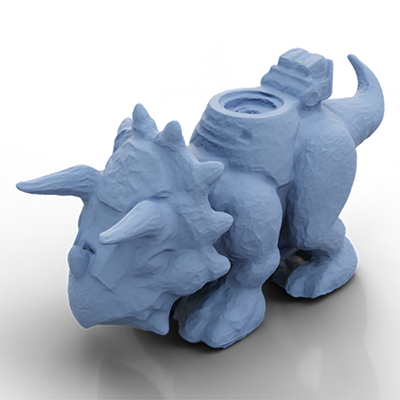}
    \\
    GT image & GT mesh & Voxurf & NeuS2 & SuGaR & 2DGS & GOF & Ours\\
    \end{tabular}
    }
    \end{footnotesize}
      \caption{More comparison of geometric reconstruction results on the OO3D-SL dataset.}
      \label{fig:OO3D-more}
      \vspace*{0.5in}
\end{figure*}

\begin{figure*}[t!] 
    \centering
    \setlength\tabcolsep{1pt}
    \begin{footnotesize}
    \begin{tabular}{cccccccc}

    \includegraphics[width=\imgleneight\textwidth]{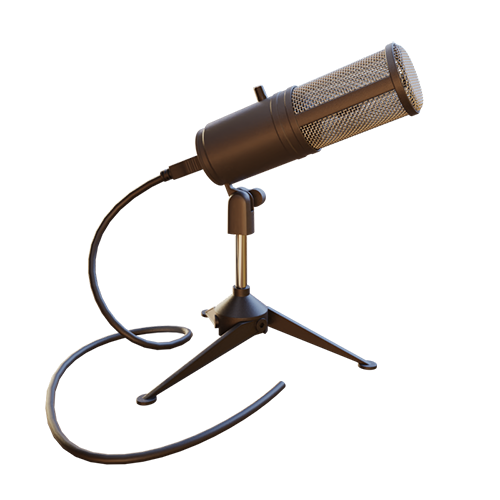}&
    \includegraphics[width=\imgleneight\textwidth]{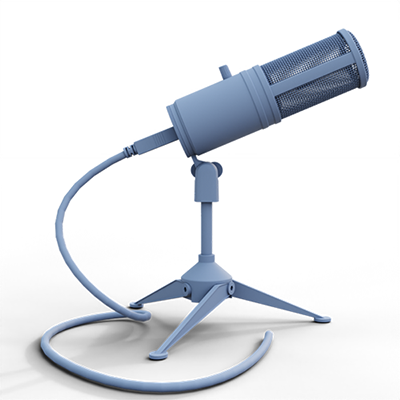}&
    \includegraphics[width=\imgleneight\textwidth]{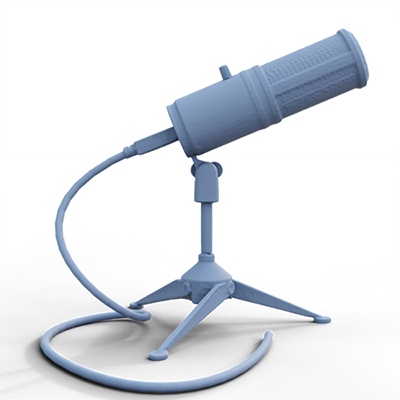}&
    \includegraphics[width=\imgleneight\textwidth]{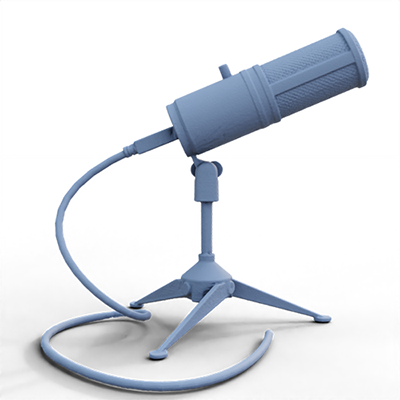}&
    \includegraphics[width=\imgleneight\textwidth]{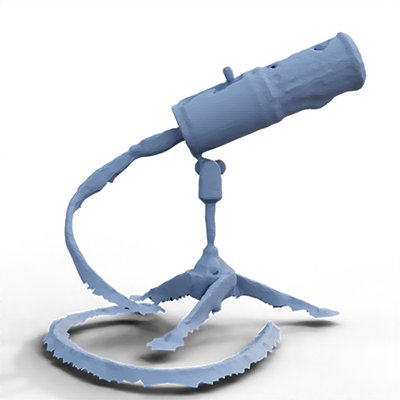}&
    \includegraphics[width=\imgleneight\textwidth]{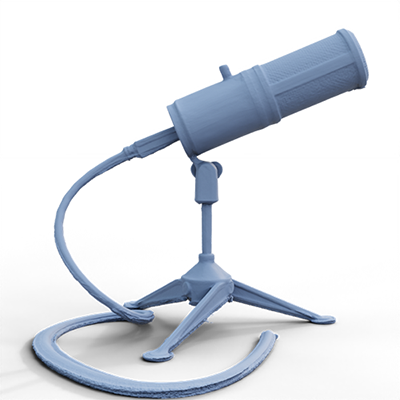}&
    \includegraphics[width=\imgleneight\textwidth]{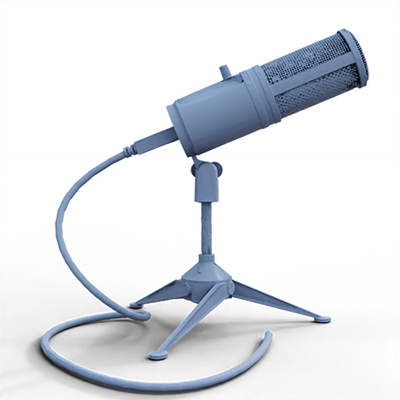}&
    \includegraphics[width=\imgleneight\textwidth]{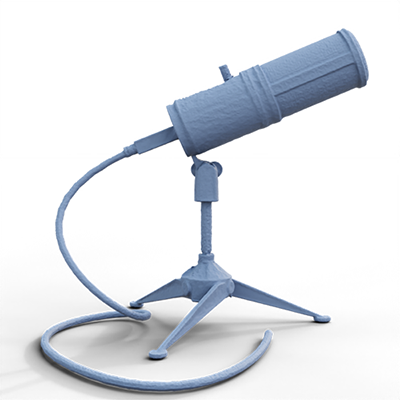}
    \\
    \includegraphics[width=\imgleneight\textwidth]{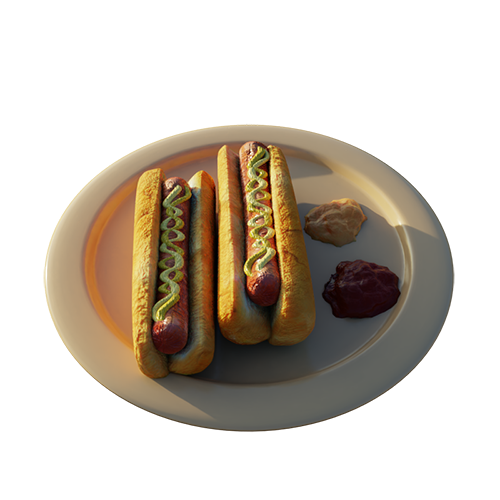}&
    \includegraphics[width=\imgleneight\textwidth]{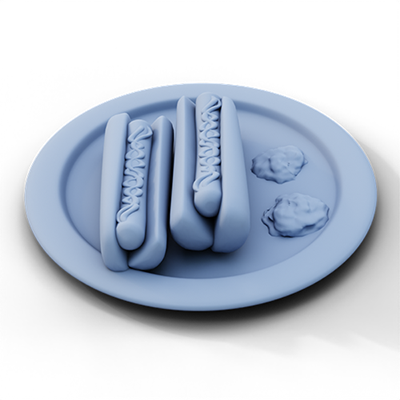}&
    \includegraphics[width=\imgleneight\textwidth]{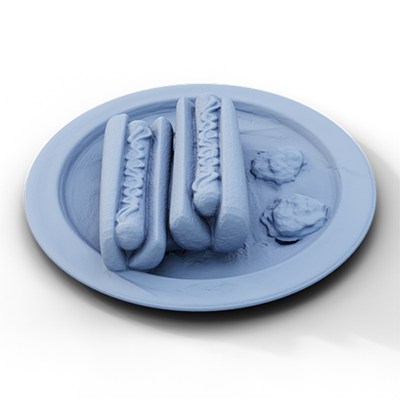}&
    \includegraphics[width=\imgleneight\textwidth]{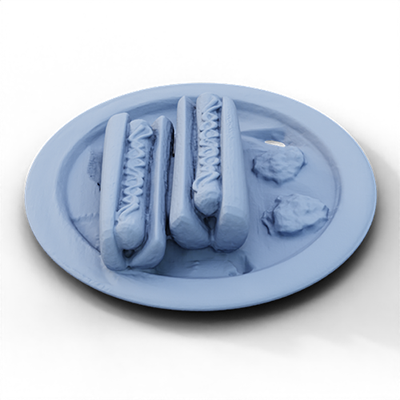}&
    \includegraphics[width=\imgleneight\textwidth]{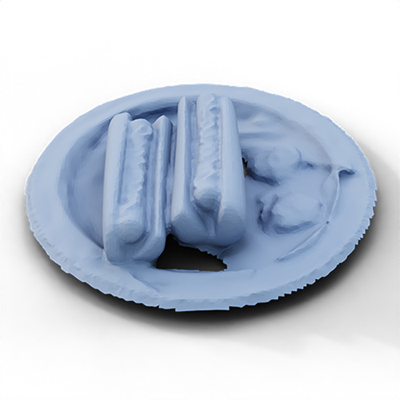}&
    \includegraphics[width=\imgleneight\textwidth]{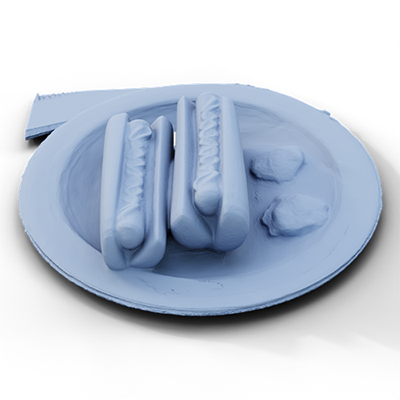}&
    \includegraphics[width=\imgleneight\textwidth]{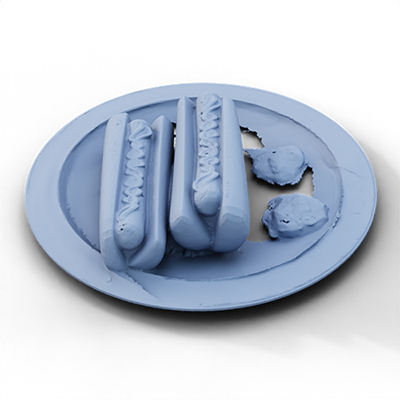}&
    \includegraphics[width=\imgleneight\textwidth]{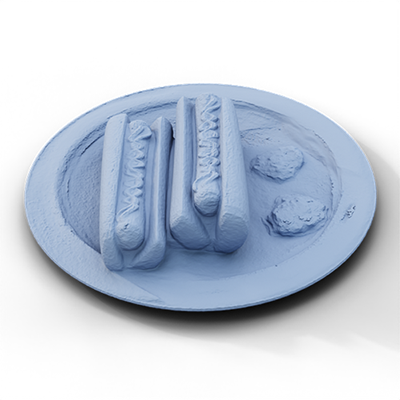}
    \\
    GT image & GT mesh & Voxurf & NeuS2 & SuGaR & 2DGS & GOF & Ours\\
    \end{tabular}
    
    \end{footnotesize}
      \caption{More comparison of geometric reconstruction results on the NeRF-Synthetic dataset.} \label{fig:OO3D-more-2}
      \vspace*{7in}
\end{figure*}

\end{document}